\documentclass[11pt,a4paper]{article}
\usepackage[hyperref]{acl2021}
\usepackage{times}
\usepackage{latexsym}

\usepackage{microtype} 

\aclfinalcopy 

\usepackage{color,xcolor}
\usepackage{graphicx}
\usepackage{algorithmic}
\usepackage{amssymb}
\usepackage{amsmath}
\usepackage{arydshln}
\usepackage{subfigure}
\usepackage{multirow}
\usepackage{amssymb}
\usepackage{pifont}

\newcommand{\m}[1]{\mathbf{#1}}

\newcommand{\tabincell}[2]{\begin{tabular}{@{}#1@{}}#2\end{tabular}}
\newcommand{\cmark}{\ding{51}}%
\newcommand{\xmark}{\ding{55}}%
\newcommand{\TODO}[1]{}

\title{BinaryBERT: Pushing the Limit of BERT Quantization}

\author{
Haoli Bai$^1$, Wei Zhang$^2$, Lu Hou$^2$, Lifeng Shang$^2$, \\
\textbf{Jing Jin$^3$, Xin Jiang$^2$, Qun Liu$^2$, Michael Lyu$^1$, Irwin King$^1$} \\
	$^1$ The Chinese University of Hong Kong	\\
	$^2$Huawei Noah's Ark Lab, $^3$Huawei Technologies Co., Ltd. \\
	{\{hlbai, lyu, king\}@cse.cuhk.edu.hk} \\
{\{zhangwei379, houlu3, shang.lifeng, jinjing12, jiang.xin, qun.liu\}@huawei.com}
}

\date{\footnote{Preprint. Working in progress.}}

\begin{document}
\maketitle

\begin{abstract}
The rapid development of large pre-trained language models has greatly increased the demand for model compression techniques, among which quantization is a popular solution. 
In this paper, we propose BinaryBERT, which pushes BERT quantization to the limit by weight binarization.
We find that a binary BERT is hard to be trained directly than a ternary counterpart due to its complex and irregular loss landscape. 
Therefore, we propose ternary weight splitting, 
which initializes BinaryBERT by equivalently splitting from a half-sized ternary network. 
The binary model thus inherits the good performance of the ternary one, and can be further enhanced by fine-tuning the new architecture after splitting.
Empirical results show that our BinaryBERT has only a slight performance drop compared with the full-precision model while being $24\times$ smaller, achieving the state-of-the-art compression results on the GLUE and SQuAD benchmarks.
\end{abstract}

\section{Introduction}
Recent pre-trained language models have achieved remarkable performance improvement in various natural language tasks~\citep{vaswani2017attention,devlin2019bert}. 
However, the improvement generally comes at the cost of increasing model size and computation, 
which limits the deployment of these huge pre-trained language models to edge devices.
Various methods have been recently proposed to compress these models, such as 
knowledge distillation~\citep{sanh2019distilbert,sun2019patient,jiao2020tinybert},
pruning~\citep{michel2019sixteen,fan2019reducing},
low-rank approximation~\citep{ma2019tensorized,lan2020albert}, 
weight-sharing~\citep{dehghani2019universal,lan2020albert,huang2021ghostbert},
dynamic networks with adaptive depth and/or width~\citep{hou2020dynabert,xin2020deebert,zhou2020bert}, and quantization~\citep{zafrir2019q8bert,shen2020qbert,fan2020training,zhang2020ternarybert}.

 \begin{figure}[t!]
 	\centering
 	\subfigure[\vspace{-1ex}MRPC.]{
 	    \includegraphics[width=0.21\textwidth]{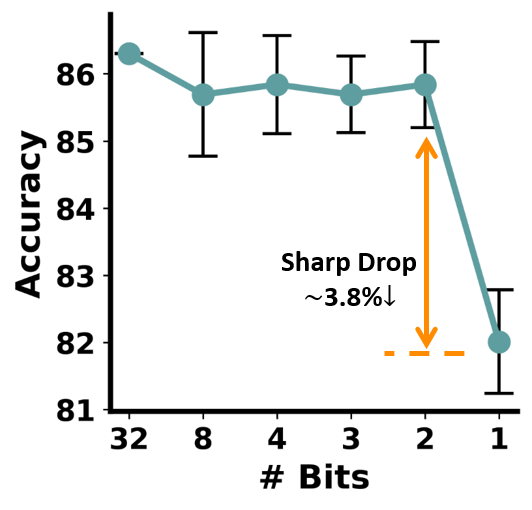}
 	}
 	\subfigure[\vspace{-1ex}MNLI-m.]{
 	    \includegraphics[width=0.225\textwidth]{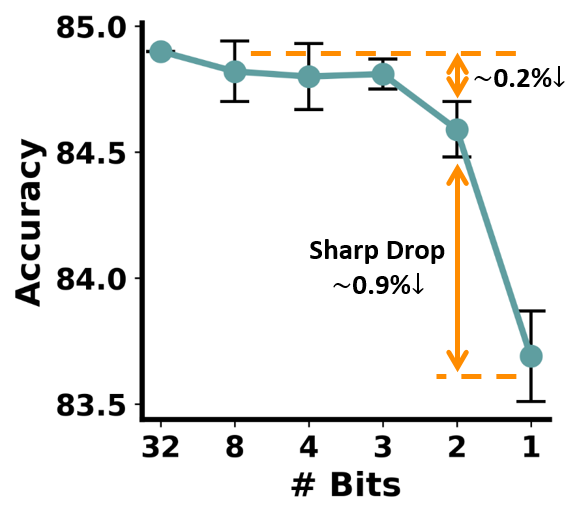}
 	}
 	\vspace{-0.15in}
 	\caption{Performance of quantized BERT with varying weight bit-widths and 8-bit activation. We report the mean results with standard deviations from 10 seeds on MRPC and 3 seeds on MNLI-m, respectively.}
 	\label{fig:acc_drop}
 \end{figure}

%


Among all these model compression approaches, quantization is a popular solution as it does not require designing a smaller model architecture.
Instead, it compresses the model by replacing each 32-bit floating-point parameter with a low-bit fixed-point representation. 
Existing attempts try to quantize pre-trained models~\citep{zafrir2019q8bert,shen2020qbert,fan2020training} to even as low as ternary values~(2-bit)  
with minor performance drop \citep{zhang2020ternarybert}. However, none of them achieves the binarization~(1-bit).
As the limit of quantization, weight binarization could bring at most $32\times$ reduction in model size
and replace most floating-point multiplications with additions.
Moreover, quantizing activations to 8-bit or 4-bit further replaces the floating-point addition with int8 and int4 addition, decreasing the energy burden and the area usage on chips~\cite{courbariaux2015binaryconnect}.


In this paper, we explore to binarize BERT parameters with quantized activations, pushing BERT quantization to the limit. 
We find that directly training a binary network is rather challenging.
According to Figure~\ref{fig:acc_drop},
there is a sharp performance drop when reducing weight bit-width from 2-bit to 1-bit,
compared to other bit configurations.
To explore the challenges of binarization, we analyze the loss landscapes of models under different precisions both qualitatively and quantitatively.
It is found that while the full-precision and ternary~(2-bit) models enjoy relatively flat and smooth loss surfaces, the binary model suffers from a rather steep and complex landscape, which poses great challenges to the optimization.


Motivated by the above empirical observations, we propose \textit{ternary weight splitting}, which takes the ternary model as a proxy to bridge the gap between the binary and full-precision models.
Specifically, ternary weight splitting equivalently converts both the quantized and latent full-precision weights in a well-trained ternary model to initialize BinaryBERT. 
Therefore, BinaryBERT retains the good performance of the ternary model, and can be further refined on the new architecture. 
While  neuron splitting is previously studied~\citep{chen2016net2net,wu2019splitting} for full-precision network, 
our ternary weight splitting is much more complex
due to the additional equivalence requirement of quantized weights.
Furthermore, the proposed BinaryBERT also supports \textit{adaptive splitting}. It can adaptively perform splitting on the most important ternary modules while leaving the rest as binary, based on efficiency constraints such as model size or floating-point operations (FLOPs). Therefore, our approach allows flexible sizes of binary models for various edge devices' demands.


Empirical results show that BinaryBERT split from a half-width ternary network is much better than
a directly-trained binary model with the original width.
On the GLUE and SQuAD benchmarks, our BinaryBERT has only a slight performance drop 
compared to the full-precision BERT-base model, while being $\mathbf{ 24\times}$ smaller. 
Moreover, BinaryBERT with the proposed importance-based adaptive splitting also outperforms other splitting criteria across a variety of model sizes.

\section{Difficulty in Training Binary BERT}
\vspace{-0.5ex}
\label{sec:difficulty}

\begin{figure*}[t]
\vspace{-0.2in}
    \subfigure[Full-precision Model.\label{fig:fp_curvature}]{
	    \includegraphics[width=0.23\textwidth]{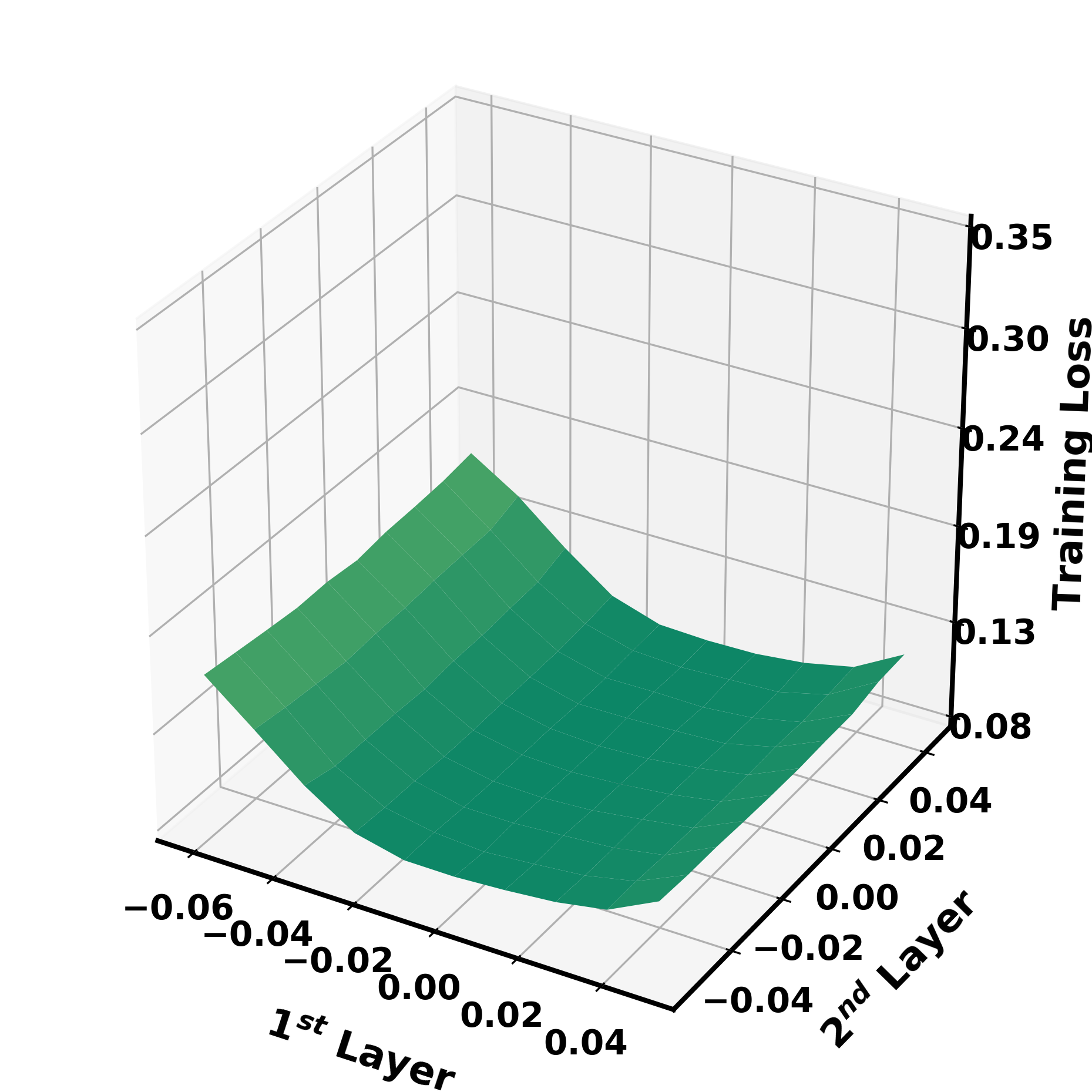}
	}
    \subfigure[Ternary Model.\label{fig:ternary_curvature}]{
	    \includegraphics[width=0.23\textwidth]{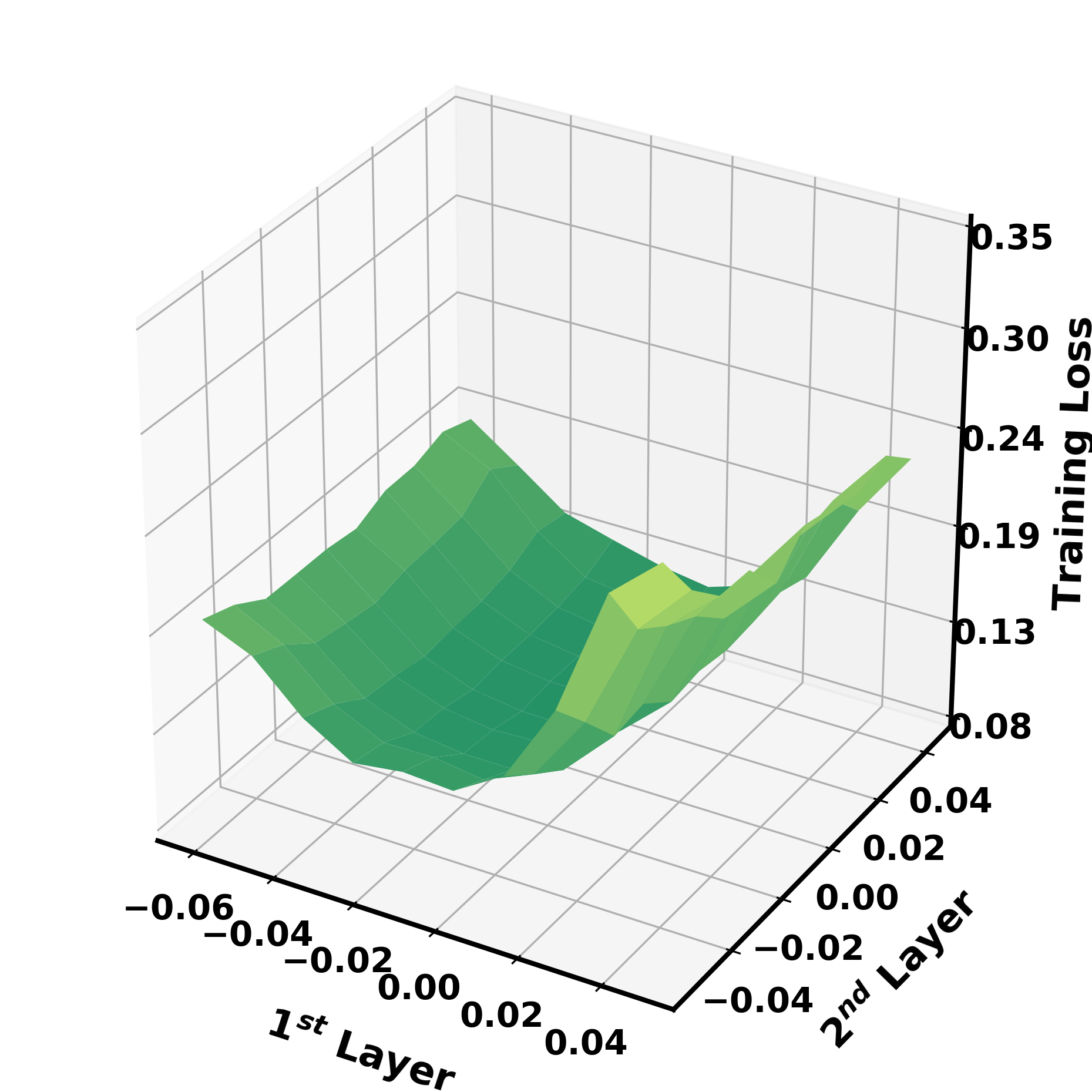}
	}
    \subfigure[Binary Model.\label{fig:binary_curvature}]{
	    \includegraphics[width=0.23\textwidth]{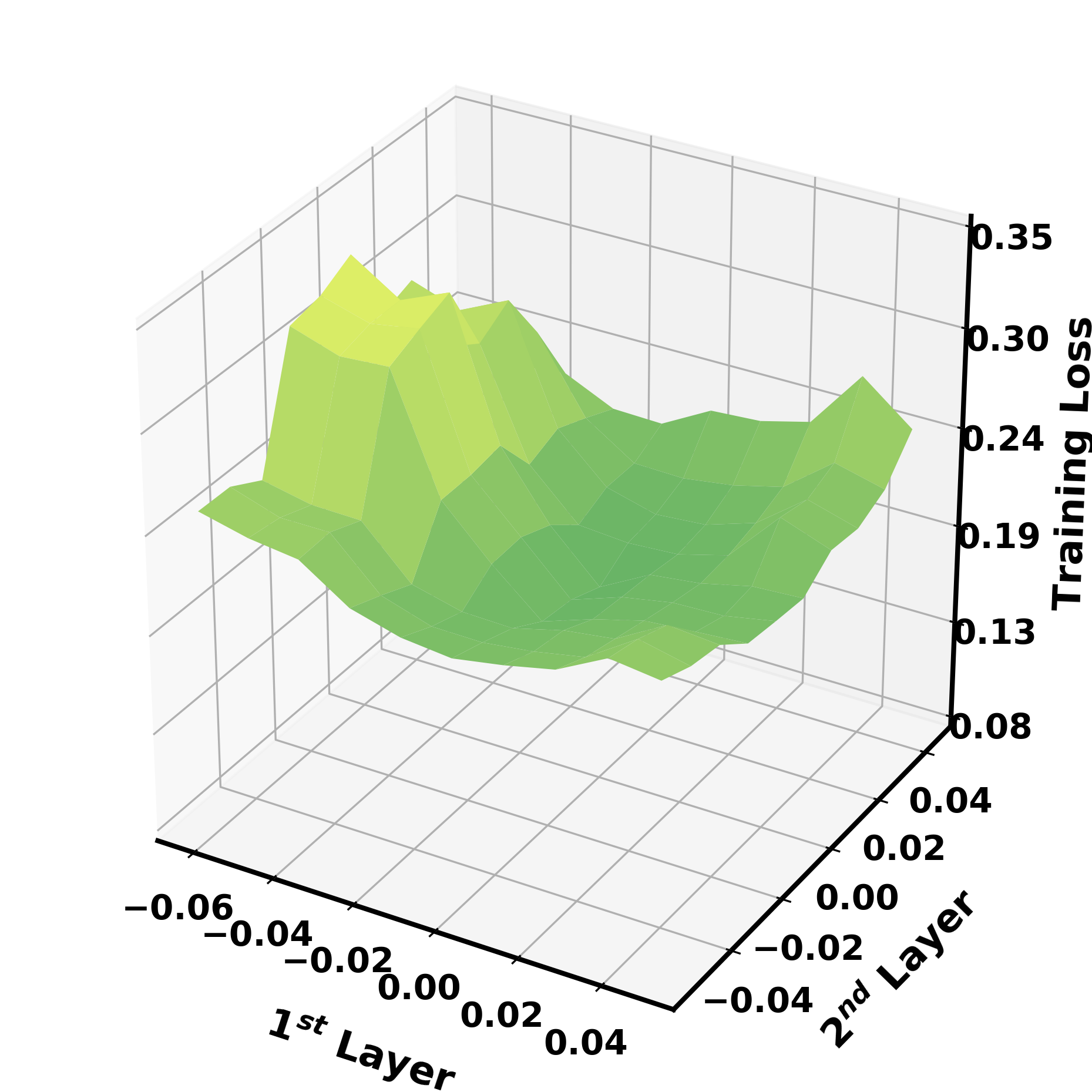}
	}
	\subfigure[All Together. \label{fig:all_curvature}] { 
		\includegraphics[width=0.23\textwidth]{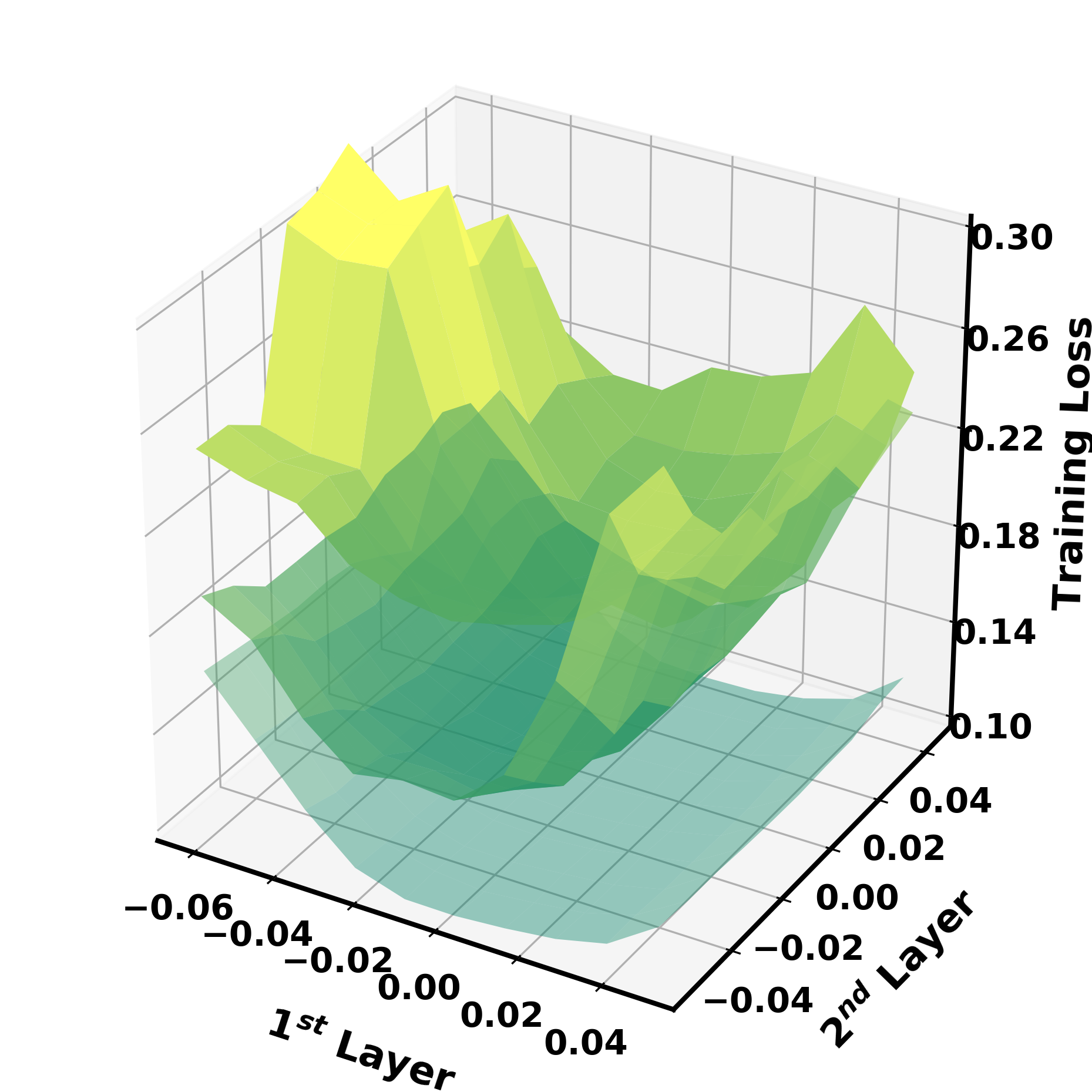}
	}
	\vspace{-0.15in}
	\caption{Loss landscapes visualization of the full-precision, ternary and binary models on MRPC. For (a), (b) and (c), we perturb the (latent) full-precision weights of the value layer in the $1^{\textrm{st}}$ and $2^{\textrm{nd}}$ Transformer layers, and compute their corresponding training loss. (d) shows the gap among the three surfaces by stacking them together.}
		\label{fig:loss_landscape}
\vspace{-0.1in}
\end{figure*}



\begin{figure*}[t]
	\subfigure[MHA-QK.]{
	    \includegraphics[width=0.2\textwidth]{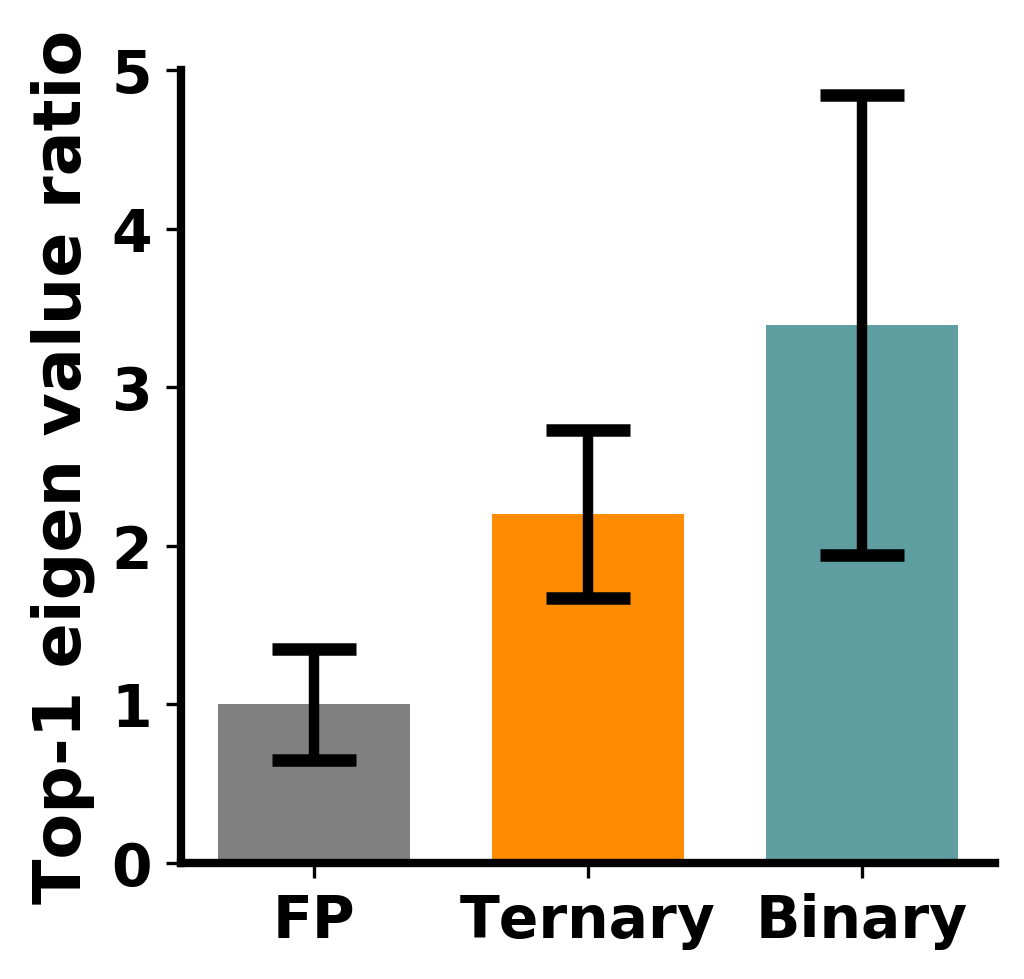}
	}
	\subfigure[MHA-V.]{
	    \includegraphics[width=0.18\textwidth]{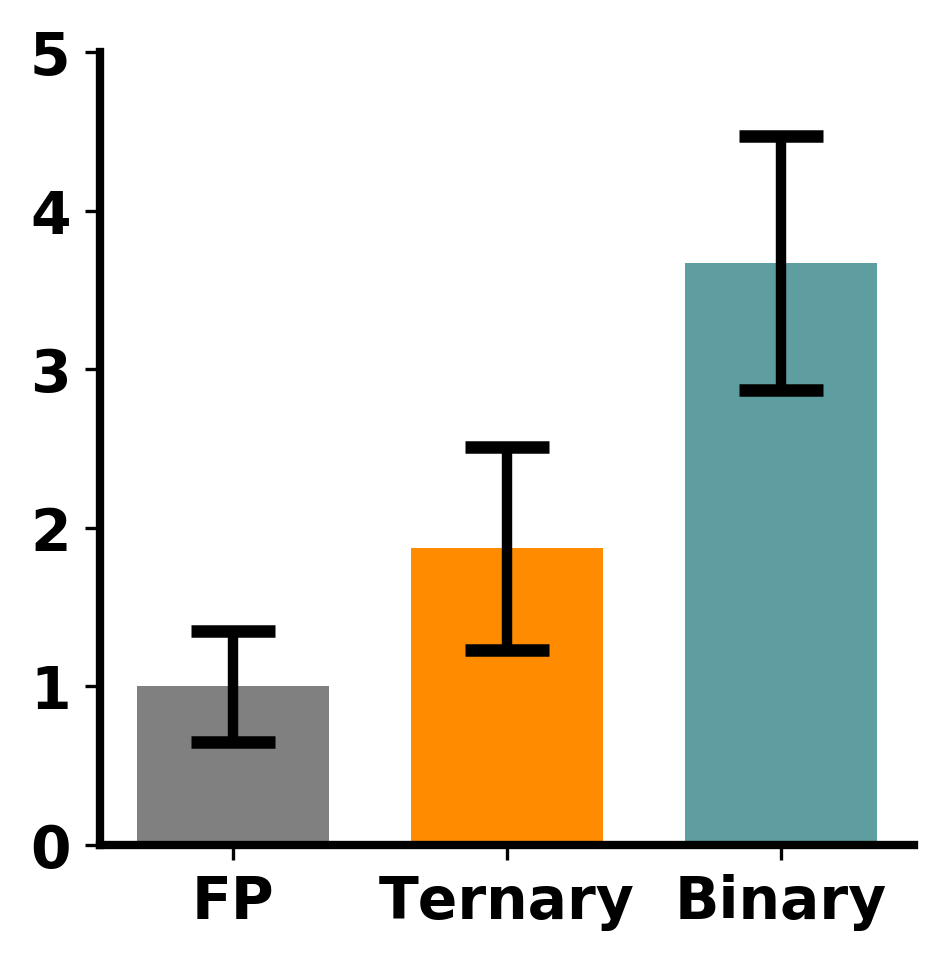}
	}
	\subfigure[MHA-O.]{
	    \includegraphics[width=0.18\textwidth]{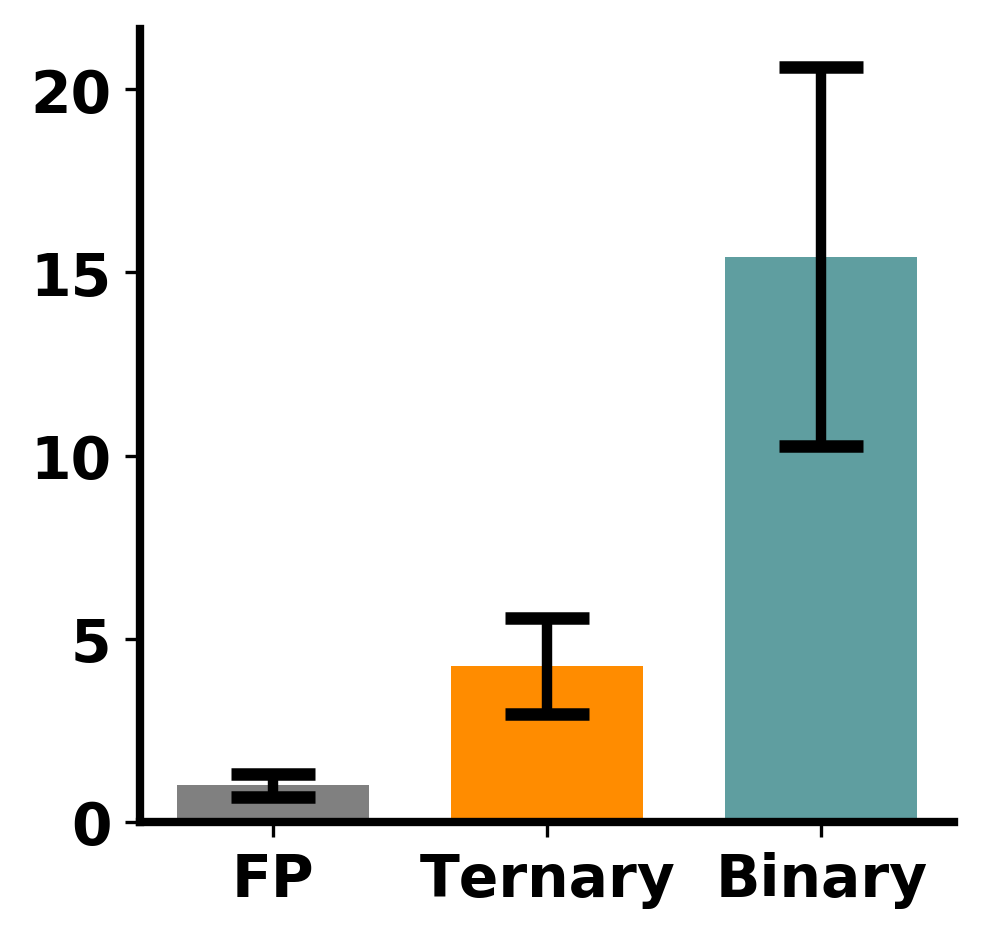}
	}
	\subfigure[FFN-Mid.]{
	    \includegraphics[width=0.18\textwidth]{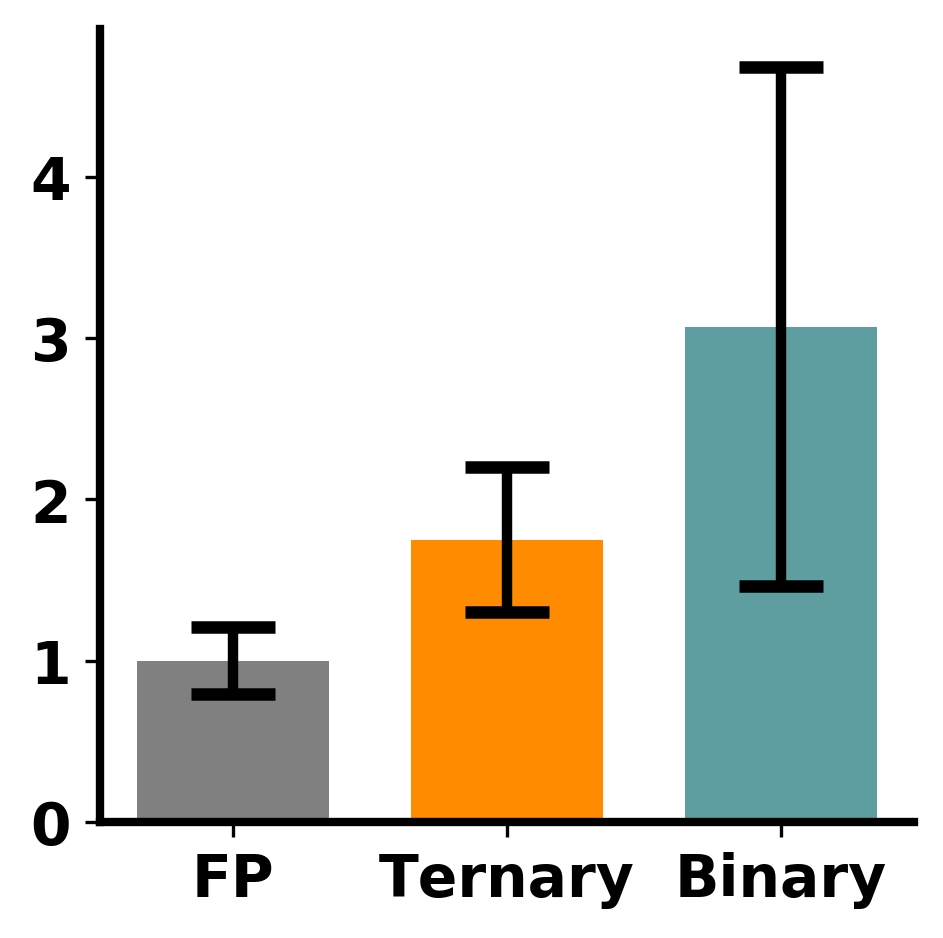}
	}
	\subfigure[FFN-Out.]{
	    \includegraphics[width=0.18\textwidth]{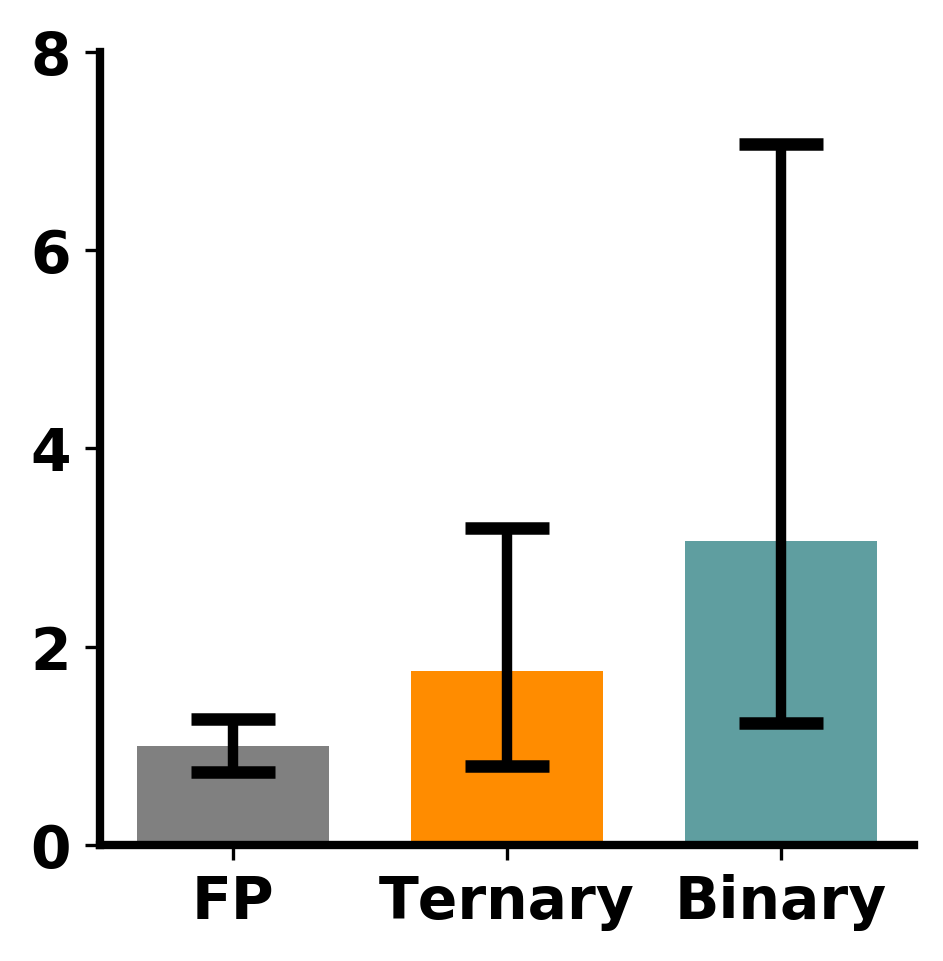}
	}
	\vspace{-0.15in}
    \caption{The top-1 eigenvalues of parameters at different Transformer parts of the full-precision (FP), ternary and binary BERT. 
    For easy comparison, we report the ratio of eigenvalue between the ternary/binary models and the full-precision model. The error bar is estimated of all Transformer layers over different data mini-batches. }
    \label{fig:top_eigen} 
\end{figure*}

In this section, we show that it is challenging to train a binary BERT with conventional binarization approaches directly. 
Before diving into details, we first 
review the necessary backgrounds. 

We follow the standard quantization-aware training procedure~\cite{zhou2016dorefa}. 
Specifically, given weight $\m w \in \mathbb{R}^n $~(a.k.a latent full-precision weights), each forward propagation quantizes it to $\hat{\m w} = \mathcal{Q} (\m w)$ 
by some quantization function $\mathcal{Q}(\cdot)$, and then computes the loss $\ell(\hat{\m w})$ at $\hat{\m w}$. 
During back propagation, we use $\nabla \ell(\hat{\m w})$ to update latent full-precision weights $\m w$ due to the non-differentiability of $\mathcal{Q}(\cdot)$, which is known as the straight-through estimator~\cite{courbariaux2015binaryconnect}.




Recent TernaryBERT~\citep{zhang2020ternarybert} follows Ternary-Weight-Network~(TWN)~\citep{li2016ternary} to quantize the elements in $\m w$ to three values $\{\pm \alpha, 0\}$. 
To avoid confusion, we use superscript $t$ and $b$ for the latent full-precision weights and quantized weights in ternary and binary models, respectively.
Specifically, TWN ternarizes each element $w_i^t$ in the ternary weight  $\m w^t$ as
\begin{equation}
\label{eq:soft_thresh}
\hat{w}_i^t \!=\!  \mathcal{Q} (w_i^t) \!=\!\left \{
\begin{array}{cc}
\!\!\!\alpha \cdot \textrm{sign}(w_i^t)  & \!\!\! |w_i^t| \geq \Delta \\
\!\!\!0  &\!\!\! | w_i^t| < \Delta
\end{array},\!\!\!\!
\right.
\end{equation}
where $\textrm{sign}(\cdot)$ is the sign function, $\Delta = \frac{0.7}{n} \|\m w^t \|_1$ and
$\alpha\!=\!\frac{1}{|\mathcal{I}|}\sum_{i\in \mathcal{I}} |w_i^t|$ 
 with
 $\mathcal{I} = \{i\ |\ \hat{w}^t_i \neq 0\}$.

\paragraph{Binarization.} 
Binarization is first proposed in \citep{courbariaux2015binaryconnect} and has been extensively studied in the academia~\citep{rastegari2016xnor,hubara2016binarized,liu2018bi}.
As a representative work, Binary-Weight-Network ~(BWN)~\citep{hubara2016binarized} binarizes $\m w^b$ element-wisely with a scaling parameter $\alpha$ as follows:
\begin{equation}
\hat{w}_i^b = \mathcal{Q} (w_i^b) = \alpha \cdot \textrm{sign}(w_i^b),\  \alpha = \frac{1}{n} \|\m w^b\|_1.
\end{equation}

Despite the appealing properties of network binarization, we show that it is non-trivial to obtain a binary BERT with these binarization  approaches.

\subsection{Sharp Performance Drop with Weight Binarization}
\label{sec:sharp_drop}
To study the performance drop of BERT quantization, we train the BERT model with full-precision, \{8,4,3,2,1\}-bit weight quantization and 8-bit activations on MRPC and MNLI-m from the GLUE benchmark~\citep{wang2018glue}~\footnote{We conduct more experiments on other GLUE datasets and with different settings in Appendix~\ref{sec:more_drop}, and find similar empirical results to MRPC and MNLI-m here.}.
We use loss-aware weight quantization (LAQ)~\citep{hou2018loss} for 8/4/3-bit weight quantization, TWN~\citep{li2016ternary} for weight ternarization and BWN~\citep{hubara2016binarized} for weight binarization. Meanwhile, we adopt 8-bit uniform quantization for activations. We follow the default experimental settings detailed in Section~\ref{sec:setup} and Appendix~\ref{sec:more_drop}.

From Figure~\ref{fig:acc_drop}, the performance drops mildly from 32-bit to as low as 2-bit, 
i.e., around $0.6\%\downarrow$ on MRPC and $0.2\%\downarrow$ on MNLI-m.
However, when reducing the bit-width to one, the performance drops sharply, i.e, $\sim3.8\%\downarrow$ and $\sim0.9\%\downarrow$ on the two tasks, respectively.
Therefore, weight binarization may severely harm the performance, which may explain why most current approaches stop at 2-bit weight quantization~\citep{shen2020qbert,zadeh2020gobo,zhang2020ternarybert}.
To further push weight quantization to the limit, a first step is to study the potential reasons behind the sharp drop from ternarization to binarization.


\subsection{Exploring the Quantized Loss Landscape}
\label{sec:challenges}

\paragraph{Visualization.}
To learn about the challenges behind the binarization, we first visually compare the loss landscapes of full-precision, ternary, and binary BERT models.
Following~\citep{nahshan2019loss}, we extract parameters $\m w_x, \m w_y$ from the value layers\footnote{We also extract parameters from other parts of the Transformer in Appendix~\ref{sec:more_visual_loss}, and the observations are similar.} of multi-head attention in the first two Transformer layers, and assign the following perturbations on parameters:
\vspace{-0.05in}
\begin{align}
    \tilde{\m w}_x = \m w_x + x\cdot \boldsymbol{1}_x, \quad \tilde{\m w}_y = \m w_y + y\cdot \boldsymbol{1}_y,
\end{align}
where $x\in \{\pm 0.2 \bar{w}_x, \pm 0.4 \bar{w}_x, ..., \pm 1.0\bar{w}_x\}$ are perturbation magnitudes based the absolute mean value $\bar{w}_x$ of $\m w_x$, and similar rules hold for $y$. $\boldsymbol{1}_x$ and $\boldsymbol{1}_y$ are vectors with all elements being 1.
For each pair of $(x, y)$, we evaluate the corresponding training loss and plot the surface in Figure~\ref{fig:loss_landscape}.

As can be seen, the full-precision model (Figure~\ref{fig:fp_curvature}) has the lowest overall training loss, and its loss landscape is flat and robust to the perturbation. 
For the ternary model (Figure~\ref{fig:ternary_curvature}), despite the surface tilts up with larger perturbations, 
it looks locally convex and is thus easy to optimize.
This may also explain why the BERT model can be ternarized without severe accuracy drop~\citep{zhang2020ternarybert}.
However, the loss landscape of the binary model (Figure~\ref{fig:binary_curvature}) turns out to be both higher and more complex. 
By stacking the three landscapes together (Figure~\ref{fig:all_curvature}), 
the loss surface of the binary BERT stands on the top with a clear margin with the other two. 
The steep curvature of loss surface reflects a higher sensitivity to binarization, which attributes to the training difficulty. 


\begin{figure*}[t!]
    \vspace{-0.2in}
	\centering
	\includegraphics[width=0.8\textwidth]{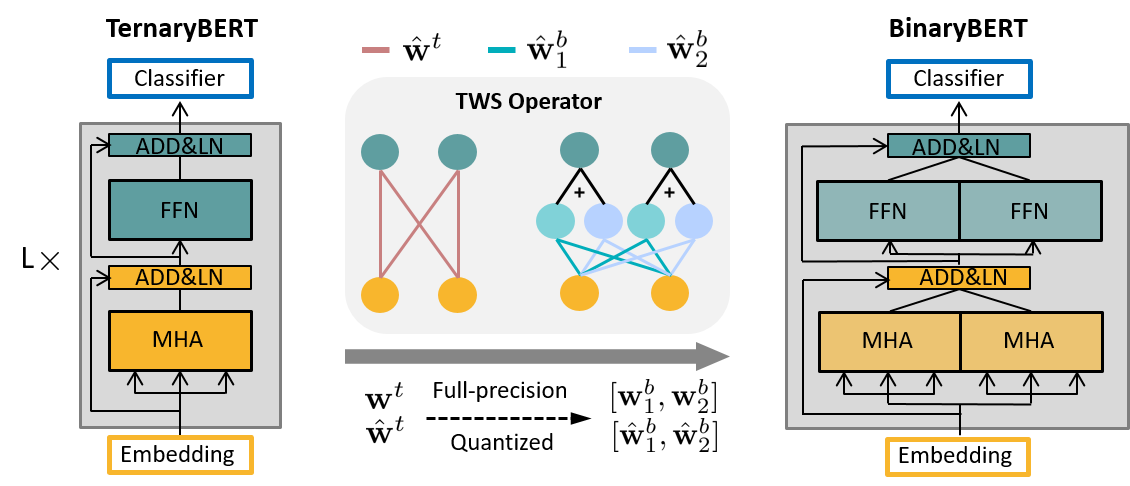}
		\vspace{-0.1in}
	\caption{The overall workflow of training BinaryBERT. We first train a half-sized ternary BERT model, and then apply ternary weight splitting operator~(Equations~\eqref{eq:solution_1} and~\eqref{eq:solution_2}) to obtain the latent full-precision and quantized weights as the initialization of the full-sized BinaryBERT. We then fine-tune BinaryBERT for further refinement.}
	\label{fig:framework}
	\vspace{-0.1in}
\end{figure*}

\paragraph{Steepness Measurement.}
To quantitatively measure the steepness of loss landscape, we start from a local minima $\m w$ and apply the second order approximation to the curvature. According to the Taylor's expansion, the loss increase induced by quantizing $\m w$ can be approximately upper bounded by
\vspace{-0.05in}
\begin{equation}
\label{eq:taylor}
\ell(\hat{\m w}) - \ell(\m w) \approx \boldsymbol{\epsilon}^{\top} \m H \boldsymbol{\epsilon} \leq \lambda_{\max} \|\boldsymbol{\epsilon}\|^2,
\end{equation}
where $\boldsymbol{\epsilon}=\m w - \hat{\m w}$ is the quantization noise, and $\lambda_{\max}$ is the largest eigenvalue of the Hessian $\m H$ at $\m w$. 
Note that the first-order term is skipped due to $\nabla \ell(\m w)=0$. Thus we take $\lambda_{\max}$ as a quantitative measurement for the steepness of the loss surface. 
Following~\citep{shen2020qbert} we adopt the power method to compute $\lambda_{\max}$. 
As it is computationally expensive to estimate $\m H$ for all $\m w$ in the network, we consider them separately as follows:
(1) the query/key layers~({MHA-QK}), 
(2) the value layer~({MHA-V}), 
(3) the output projection layer~({MHA-O}) in the multi-head attention, 
(4) the intermediate~ layer ({FFN-Mid}), and 
(5) the output layer~({FFN-Out}) in the feed-forward network.
Note that we group key and query layers as they are used together to calculate the attention scores.

From Figure~\ref{fig:top_eigen}, the top-1 eigenvalues of the binary model are higher both on expectation and standard deviation compared to the full-precision baseline and the ternary model. For instance, the top-1 eigenvalues of MHA-O in the binary model are $\sim 15\times$ larger than the full-precision counterpart.
Therefore, the quantization loss increases of full-precision and ternary model are tighter bounded than the binary model in Equation~\eqref{eq:taylor}. 
The highly complex and irregular landscape by binarization thus poses more challenges to the optimization.

\section{Proposed Method}

\subsection{Ternary Weight Splitting}
\label{sec:split}
\vspace{-0.5ex}

Given the challenging loss landscape of binary BERT, we propose \textit{ternary weight splitting}~(TWS) that exploits the flatness of ternary loss landscape as the optimization proxy of the binary model. 
As is shown in Figure~\ref{fig:framework}, we first train the half-sized ternary BERT to convergence, 
and then split both the latent full-precision weight $\m w^t$ and quantized $\hat{\m w}^t$ to their binary counterparts $\m w_1^b, \m w_2^b$ and $\hat{\m w}_1^b, \hat{\m w}_2^b$ via the \textit{TWS operator}. To inherit the performance of the ternary model after splitting, the TWS operator requires the splitting equivalency~(i.e., the same output given the same input):
\vspace{-0.5ex}
\begin{equation}
\begin{array}{ll}
\m w^t = \m w^b_1 + \m w^b_2, \quad 
\hat{\m w}^t = \hat{\m w}^b_1 + \hat{\m w}^b_2 
\end{array}.  \label{eq:split_obj}
\end{equation}
While solution to Equation~\eqref{eq:split_obj} is not unique, we constrain the latent full-precision weights after splitting 
$\m w^b_1, \m w^b_2$ to satisfy $\m w^t = \m w^b_1 + \m w^b_2$ as 
\begin{align}
\!\!\!\!& w^b_{1,i} = \left \{
\begin{array}{ll}
a\cdot w^t_i & \;\;\;\;\textrm{if}  \;\; \hat{w}^t_i \neq 0 \\
b + w^t_i & \;\;\;\; \textrm{if} \;\; \hat{w}^t_i = 0,  w^t_i\! > 0  \\
b  & \;\;\;\; \textrm{otherwise}
\end{array},
\right. \label{eq:solution_1} \\
\!\!\!\!& w^b_{2,i} = \left \{
\begin{array}{ll}
(1\!-\!a) w^t_i & \textrm{if} \;\; \hat{w}^t_i \neq 0 \\
-b  & \textrm{if}  \;\; \hat{w}^t_i = 0,  w^t_i \!> 0  \\
-b + w^t_i  & \textrm{otherwise}
\end{array},
\right. \label{eq:solution_2}
\end{align}
where $a$ and $b$ are the variables to solve.
By Equations \eqref{eq:solution_1} and \eqref{eq:solution_2} with
$\hat{\m w}^t = \hat{\m w}^b_1 + \hat{\m w}^b_2 $, we get
\begin{eqnarray}
\label{eq:solution}
a & = &\frac{\sum_{i\in\mathcal{I}} |w^t_i| + \sum_{j\in \mathcal{J}} |w^t_j| - \sum_{k\in \mathcal{K}}|w^t_k| }{2\sum_{i\in\mathcal{I}} |w^t_i|}, \nonumber \\
b & =& \frac{\frac{n}{|\mathcal{I}|}\sum_{i\in \mathcal{I}} |w^t_i| - \sum_{i=1}^n |w^t_i| }{2(|\mathcal{J}| + |\mathcal{K}|)},
\end{eqnarray}    
where we denote $\mathcal{I} = \{i\ |\ \hat{w}^t_i \neq 0\}$, $\mathcal{J} = \{j\ |\ \hat{w}^t_j = 0 \ \textrm{and}\ w^t_j > 0 \}$ and $\mathcal{K} = \{k\ |\ \hat{w}^t_k = 0 \ \textrm{and}\ w^t_k < 0 \}$. 
$|\cdot|$ denotes the cardinality of the set. 
Detailed derivation of Equation~\eqref{eq:solution} is in Appendix~\ref{sec:appendix:derive}.








\paragraph{Quantization Details.}
Following \citep{zhang2020ternarybert}, for each weight matrix in the Transformer layers, we use layer-wise ternarization (i.e., one scaling parameter
for all elements in the weight matrix).
For word embedding, we use row-wise ternarization (i.e., one scaling
parameter for each row in the embedding).
After splitting, each of the two split matrices has its own scaling factor. 

Aside from weight binarization, we simultaneously quantize activations 
before all matrix multiplications, which could accelerate inference on specialized hardwares~\citep{shen2020qbert,zafrir2019q8bert}.
Following~\citep{zafrir2019q8bert,zhang2020ternarybert}, we skip the quantization for all layer-normalization~(LN) layers, skip connections, and bias as their calculations are negligible compared to matrix multiplication.
The last classification layer is also not quantized to avoid a large accuracy drop.

\paragraph{Training with Knowledge Distillation.}
Knowledge distillation is shown to benefit BERT quantization~\cite{zhang2020ternarybert}.
Following~\citep{jiao2020tinybert,zhang2020ternarybert}, we first perform \textit{intermediate-layer distillation} from the full-precision teacher network's embedding $\m E$, layer-wise MHA output $\m M_l$ and FFN output $\m F_l$ to the quantized student counterpart $\hat{\m E}, \hat{\m M}_l, \hat{\m F}_l$ ($l=1,2,...L$). We aim to minimize their mean sqaured errors, i.e., $\ell_{emb} = \textrm{MSE}(\hat{\m E}, \m E)$, $\ell_{mha} = \sum_{l} \textrm{MSE}(\hat{\m M}_l, \m M_l)$, and $\ell_{ffn} = \sum_{l} \textrm{MSE}(\hat{\m F}_l, \m F_l)$. Thus the objective function is
\vspace{-0.5ex}
\begin{equation}
\ell_{int} = \ell_{emb} + \ell_{mha} + \ell_{ffn}.
\end{equation}
\vspace{-0.5ex}
We then conduct \textit{prediction-layer distillation} by minimizing the soft cross-entropy ($\textrm{SCE}$) between quantized student logits $\hat{\m y}$ and teacher logits $\m y$, i.e.,
\vspace{-0.5ex}
\begin{equation}
\ell_{pred} = \textrm{SCE}(\hat{\m y}, \m y).
\end{equation}

\paragraph{Further Fine-tuning.}
After splitting from the half-sized ternary model, the binary model inherits its performance on a new architecture with full width. 
However, the original minimum of the ternary model may not hold in this new loss landscape after splitting. Thus we further fine-tune with prediction-layer distillation to look for a better solution. We dub the resulting model as BinaryBERT.



\subsection{Adaptive Splitting}
\label{sec:adaptive_split}
Our proposed approach also supports \textit{adaptive splitting} that can flexibly adjust the width of BinaryBERT, based on the parameter sensitivity to binarization and resource constraints of edge devices.


Specifically, given the resource constraints $\mathcal{C}$ (e.g., model size and computational FLOPs), we first train a mixed-precision model adaptively~(with sensitive parts being ternary and the rest being binary), and then split ternary weights into binary ones.
Therefore, adaptive splitting finally enjoys consistent arithmetic precision~(1-bit) for all weight matrices, which is usually easier to deploy than the mixed-precision counterpart.




\paragraph{Formulation.}
Intuitively, we assign ternary values to weight matrices that are more sensitive to quantization.
The quantization sensitivity of the weight matrix is empirically measured by the performance gain of not quantizing it comparing to the fully-quantized counterpart (Details are in Appendix~\ref{sec:sensitivity}.).
We denote $\m u\in \mathbb{R}_{+}^Z$ as the sensitivity vector, where $Z$ is the total number of splittable weight matrices in all Transformer layers, the word embedding layer and the pooler layer.
The cost vector $\m c\in \mathbb{R}_{+}^Z$ stores the additional increase of parameter or FLOPs of each ternary weight matrix against a binary choice. 
The splitting assignment can be represented as a binary vector $\m s\in \{0, 1\}^Z$, 
where $s_z=1$ means to ternarize the $z$-th weight matrix, and vice versa.
The optimal assignment  $ \m s^*$ can thus be solved from the following combinatorial optimization problem:
\begin{eqnarray}
\label{eq:dp_problem}
    & \max_{\m s}&\hspace{1ex}  \m u^{\top} \m s \hspace{2ex} \\
    &\textrm{s.t.}&\hspace{1ex} \m c^{\top} \m s \leq \mathcal{C} - \mathcal{C}_0, \hspace{1ex} \m s \in\{0, 1\}^Z, \nonumber
\end{eqnarray}
where $\mathcal{C}_0$ is the baseline efficiency of the half-sized binary network.
Dynamic programming can be applied to solve Equation \eqref{eq:dp_problem} to avoid NP-hardness.
\section{Experiments}
\label{sec:exp}

\begin{table*}[t]
	\centering
	\vspace{-0.1in}
	\resizebox{0.98\textwidth}{!}{
		\begin{tabular}{r||cccc|c|cccccccc|c}
			\textbf{\#} & 
			\textbf{\tabincell{c}{Quant}} &
			\textbf{\tabincell{c}{\#Bits\\(W-E-A)}} &  \textbf{\tabincell{c}{Size\\(MB)}} & 
			\textbf{\tabincell{c}{FLOPs\\(G)}} &
			\textbf{DA} &
			\textbf{\tabincell{c}{MNLI\\-m/mm}} & \textbf{QQP} & \textbf{QNLI} & \textbf{SST-2} & \textbf{CoLA} & \textbf{STS-B} & \textbf{MRPC} & \textbf{RTE} &
			\textbf{Avg.}\\\hline\hline
			1 &  - & \textit{full-prec.} & 417.6 & 22.5 & - & 84.9/85.5 & 91.4 & 92.1 & 93.2 & 59.7 & 90.1 & 86.3 & 72.2 & 83.9 \\\hline
			2 & BWN & 1-1-8 & 13.4 & 3.1 & \xmark & 84.2/84.0 & 91.1 & 90.7 & 92.3 & 46.7 & 86.8 & 82.6 & 68.6 & 80.8 \\
			3 & TWS & 1-1-8 &  16.5 & 3.1 & \xmark & \textbf{84.2/84.7} & \textbf{91.2} & \textbf{91.5} & \textbf{92.6} & \textbf{53.4} & \textbf{88.6} & \textbf{85.5} & \textbf{72.2} & \textbf{82.7} \\\hline
			4 & BWN  &  1-1-4 & 13.4 & 1.5 & \xmark & 83.5/83.4 & 90.9 & 90.7 & \textbf{92.3} & 34.8 & 84.9 & 79.9 & \textbf{65.3} & 78.4 \\
			5 & TWS & 1-1-4 & 16.5 & 1.5 & \xmark & \textbf{83.9/84.2} & \textbf{91.2} & \textbf{90.9} & \textbf{92.3} & \textbf{44.4} & \textbf{87.2} & \textbf{83.3} & \textbf{65.3} & \textbf{79.9} \\\hline
			6 & BWN & 1-1-8 & 13.4 & 3.1 & \cmark & 84.2/84.0 & 91.1 & 91.2 & 92.7 & 54.2 & 88.2 & \textbf{86.8} & 70.0 & 82.5 \\
			7 & TWS & 1-1-8 &  16.5 & 3.1 & \cmark & \textbf{84.2/84.7} & \textbf{91.2} & \textbf{91.6} & \textbf{93.2} & \textbf{55.5} & \textbf{89.2} & 86.0 & \textbf{74.0} & \textbf{83.3} \\\hline
			8 & BWN  &  1-1-4 & 13.4 & 1.5 & \cmark & 83.5/83.4 & 90.9 & 91.2 & 92.5 & 51.9 & 87.7 & 85.5 & 70.4 & 81.9 \\
			9 & TWS & 1-1-4 & 16.5 & 1.5 & \cmark & \textbf{83.9/84.2} & \textbf{91.2} & \textbf{91.4} & \textbf{93.7} & \textbf{53.3} & \textbf{88.6} & \textbf{86.0} & \textbf{71.5} & \textbf{82.6} \\
	\end{tabular}}
	\vspace{-0.05in}
	\caption{Results on the GLUE development set. 
	``\#Bits (W-E-A)'' represents the bit number for weights of Transformer layers, word embedding, and activations. 
	``DA'' is short for data augmentation. 
	``Avg." denotes the average results of all tasks including MNLI-m and MNLI-mm. 
	The higher results in each block are bolded.}
	\label{table:dev_glue}
\end{table*}

\begin{table*}[t]
	\centering
	\resizebox{0.98\textwidth}{!}{
		\begin{tabular}{l||cccc|c|cccccccc|c}
			\textbf{\#} & 
			\textbf{\tabincell{c}{Quant}} &
			\textbf{\tabincell{c}{\#Bits\\(W-E-A)}} &  \textbf{\tabincell{c}{Size\\(MB)}} &
			\textbf{\tabincell{c}{FLOPs\\(G)}} &
			\textbf{DA} &
			\textbf{\tabincell{c}{MNLI\\-m/mm}} & \textbf{QQP} & \textbf{QNLI} & \textbf{SST-2} & \textbf{CoLA} & \textbf{STS-B} & \textbf{MRPC} & \textbf{RTE} & \textbf{Avg.} \\\hline\hline
			1 &  - & \textit{full-prec.} & 417.6 & 22.5 & - & 84.5/84.1 & 89.5 & 91.3 & 93.0 & 54.9 & 84.4 & 87.9 & 69.9  & 82.2 \\\hline
			2 & BWN & 1-1-8 & 13.4 & 3.1 & \xmark  & 83.3/83.4 & 88.9 & \textbf{90.1} & 92.3 & 38.1 & 81.2 & \textbf{86.1} & 63.1 & 78.5 \\
			3 & TWS & 1-1-8 & 16.5 & 3.1 & \xmark & \textbf{84.1/83.6} & \textbf{89.0} & 90.0 & \textbf{93.1} & \textbf{50.5} & \textbf{83.4} & 86.0 & \textbf{65.8} & \textbf{80.6} \\\hline
			4 & BWN & 1-1-4 & 13.4 & 1.5 & \xmark & 83.5/82.5 & \textbf{89.0} & \textbf{89.4} & 92.3 & 26.7 & 78.9 & 84.2 & 59.9 & 76.3 \\
			5 & TWS &  1-1-4 & 16.5 & 1.5 & \xmark& \textbf{83.6/82.9} & \textbf{89.0} & 89.3 & \textbf{93.1} & \textbf{37.4} & \textbf{82.5} & \textbf{85.9} & \textbf{62.7} & \textbf{78.5} \\\hline
			6 & BWN & 1-1-8 & 13.4 & 3.1 & \cmark  & 83.3/83.4 & 88.9 & \textbf{90.3} & 91.3 & 48.4 & \textbf{83.2} & \textbf{86.3} & 66.1 & 80.1 \\
			7 & TWS  & 1-1-8 & 16.5 & 3.1 & \cmark & \textbf{84.1/83.5} & \textbf{89.0} & 89.8 & \textbf{91.9} & \textbf{51.6} & 82.3 & 85.9 & \textbf{67.3} & \textbf{80.6} \\\hline
			8 & BWN & 1-1-4 & 13.4 & 1.5 & \cmark & 83.5/82.5 & \textbf{89.0} & \textbf{89.9} & 92.0 & 45.0 & 81.9 & 85.2 & 64.1 & 79.2 \\
			9 & TWS &  1-1-4 & 16.5 & 1.5 & \cmark & \textbf{83.6/82.9} & \textbf{89.0} & 89.7 & \textbf{93.1} & \textbf{47.9} & \textbf{82.9} & \textbf{86.6} & \textbf{65.8} & \textbf{80.2} \\
	\end{tabular}}
    \vspace{-0.05in}	
	\caption{Results on the GLUE test set scored using the GLUE evaluation server.}
	\label{table:test_glue}	
\end{table*}

In this section, we empirically verify our proposed approach 
on the GLUE~\citep{wang2018glue} and SQuAD~\citep{rajpurkar2016squad,rajpurkar2018know} benchmarks. 
We first introduce the experimental setup in Section~\ref{sec:setup}, and then present the main experimental results on both benchmarks in Section~\ref{sec:results}. We compare with other state-of-the-arts in Section~\ref{sec:compare_to_sota}, and finally provide more discussions on the proposed methods in Section~\ref{sec:discussion}.
Code is available at \url{https://github.com/huawei-noah/Pretrained-Language-Model/tree/master/BinaryBERT}.




\subsection{Experimental Setup}
\label{sec:setup}

\paragraph{Dataset and Metrics.}
The GLUE benchmark contains multiple natural language understanding tasks.
We follow \citet{devlin2019bert} to evaluate the performance on these tasks: 
Matthews correlation for CoLA, 
Spearman correlation for STS-B and 
accuracy for the rest tasks: RTE, MRPC, SST-2, QQP, MNLI-m~(matched) and MNLI-mm~(mismatched).
For machine reading comprehension on SQuAD, we report the EM~(exact match) and F1 score.

Aside from the task performance, we also report the model size (MB) and computational FLOPs at inference. 
For quantized operations, we follow~\cite{zhou2016dorefa,liu2018bi,li2020additive} to count the bit-wise operations, 
i.e., the multiplication between an $m$-bit number and an $n$-bit number approximately takes $mn / 64$ FLOPs for a CPU with the instruction size of 64 bits.


\paragraph{Implementation.}
We take DynaBERT~\cite{hou2020dynabert} sub-networks as backbones as they offer both half-sized and full-sized models
for easy comparison. 
We start from training a ternary model of width $0.5\times$ 
with the two-stage knowledge distillation introduced in Section~\ref{sec:split}.
Then we split it into a binary model with width $1.0\times$, 
and perform further fine-tuning with prediction-layer distillation.
Each training stage takes the same number of training epochs.
Following~\cite{jiao2020tinybert,hou2020dynabert,zhang2020ternarybert}, we adopt data augmentation with one training epoch in each stage on all GLUE tasks except for MNLI and QQP. Aside from this default setting, we also remove data augmentation and perform vanilla training with 6 epochs on these tasks.
On MNLI and QQP, we train 3 epochs for each stage. 

We verify our ternary weight splitting~(\textbf{TWS}) against vanilla binary training~(\textbf{BWN}), the latter of which doubles training epochs to match the overall training time in TWS for fair comparison.
More training details are provided in Appendix~\ref{sec:appendix_implementation}.


\TODO{
	We compare with two baselines: 
	(i) \textbf{TWN} with $0.5\times$ width, and 
	(ii) \textbf{BWN} with the original $1.0\times$ width. 
	For fair comparison, both (i) and (ii) are trained with the same setting to our ternary weight splitting~(\textbf{TWS}) 
	besides the quantization scheme.
	Finally, we also compare with a number of state-of-the-art approaches in Section~\ref{sec:compare_to_sota}.
}





\paragraph{Activation Quantization.}
While BinaryBERT focuses on weight binarization, we also explore activation quantization in our implementation, which is beneficial for reducing the computation burden on specialized hardwares~\cite{hubara2016binarized,zhou2016dorefa,zhang2020ternarybert}. 
Aside from 8-bit uniform quantization~\citep{zhang2020ternarybert,shen2020qbert} in past efforts, we further pioneer to study 4-bit activation quantization. 
We find that uniform quantization can hardly deal with outliers in the activation. Thus we use Learned Step-size Quantization~(LSQ)~\citep{esser2019learned} to directly learn the quantized values, which empirically achieves better quantization performance.





\subsection{Experimental Results}
\label{sec:results}

\subsubsection{Results on the GLUE Benchmark}
\label{sec:glue_results}



The main results on the development set are shown in Table~\ref{table:dev_glue}. 
For results without data augmentation~(row \#2-5), our ternary weight splitting method outperforms BWN with a clear margin
\footnote{Note that DynaBERT only squeezes width in the Transformer layers but not the word embedding layer, 
	thus the split binary model has a slightly larger size than BWN.}. 
For instance, on CoLA, ternary weight splitting achieves $6.7\%\uparrow$ and $9.6\%\uparrow$ with 8-bit and 4-bit activation quantization, respectively.
While data augmentation~(row 6-9) mostly improves each entry, our approach still overtakes BWN consistently. 
Furthermore, 4-bit activation quantization empirically benefits more from ternary weight splitting~(row 4-5 and 8-9) compared with 8-bit activations~(row 2-3 and 6-7), demonstrating the potential of our approach in extremely low bit quantized models.


In Table~\ref{table:test_glue}, we also provide the results on the test set of GLUE benchmark. 
Similar to the observation in Table~\ref{table:dev_glue}, our approach achieves consistent improvement on 
both 8-bit and 4-bit activation quantization compared with BWN.









\subsubsection{Results on SQuAD Benchmark}
\label{sec:squad_results}
The results on the development set of SQuAD v1.1 and v2.0 are shown in Table~\ref{tab:main_squad}. 
Our proposed ternary weight splitting again outperforms BWN
w.r.t both EM and F1 scores on both datasets. 
Similar to previous observations, 4-bit activation enjoys a larger gain in performance from the splitting approach. 
For instance, our approach improves the EM score of 4-bit activation by $1.8\%$ and $0.6\%$ on SQuAD v1.1 and v2.0, respectively, 
both of which are higher than those of 8-bit activation.

\begin{table}[t]
	\centering
	\resizebox{0.48\textwidth}{!}{
		\begin{tabular}{c|ccc|cc}
			\textbf{\tabincell{c}{Quant}} &
			\textbf{\tabincell{c}{\#Bits\\(W-E-A)}} &
			\textbf{\tabincell{c}{Size\\(MB)}} & \textbf{\tabincell{c}{FLOPs\\(G)}} &  \textbf{\tabincell{c}{SQuAD\\v1.1}} & \textbf{\tabincell{c}{SQuAD\\v2.0}} \\ \hline
			- & \textit{full-prec.} &  417.6  & 22.5 & 82.6/89.7 & 75.1/77.5 \\ \hline
			BWN & 1-1-8 & 13.4  & 3.1 & 79.2/86.9 & \textbf{73.6/76.6}\\
			TWS & 1-1-8 & 16.5 & 3.1 & \textbf{80.8/88.3} & 73.6/76.5 \\\hline
			BWN & 1-1-4 & 13.4 & 1.5 & 77.5/85.8 & 71.9/75.1 \\
			TWS & 1-1-4 & 16.5 & 1.5 & \textbf{79.3/87.2} & \textbf{72.5/75.4} \\
		\end{tabular}
	}  
	\vspace{-0.1in}
	\caption{Development set results (EM/F1) on SQuAD. }
	\label{tab:main_squad}
\end{table}

\begin{figure}[t]
	\centering
	\subfigure{
		\includegraphics[width=0.4\textwidth]{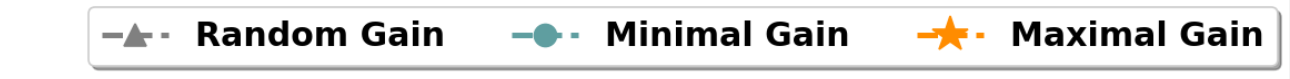}}
	\vspace{-0.1in}
	\\
	\addtocounter{subfigure}{-1}
	\subfigure[8-bit Activation.\label{fig:8bit}]{
		\includegraphics[width=0.235\textwidth]{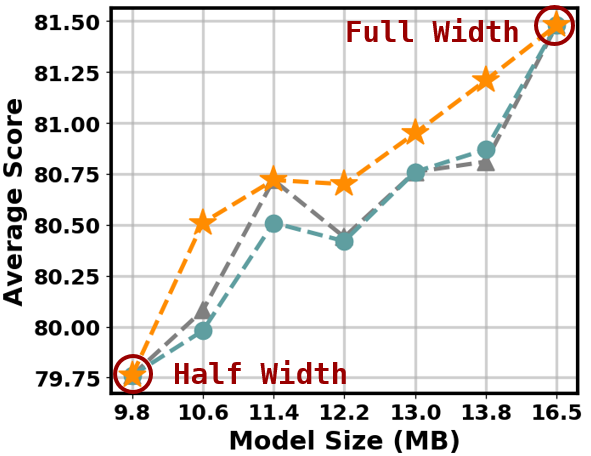}}
	\subfigure[4-bit Activation.\label{fig:4bit}]{
		\includegraphics[width=0.223\textwidth]{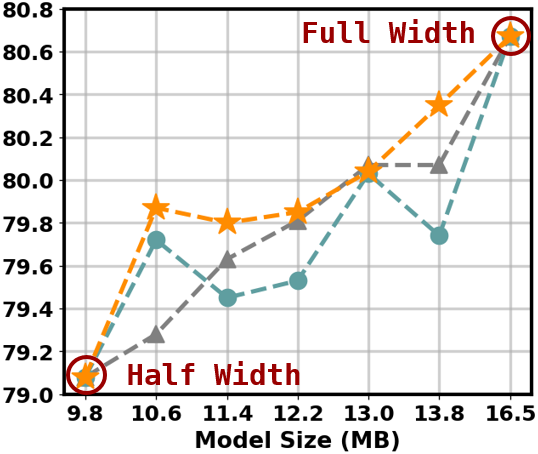}}
	\vspace{-0.15in}
	\caption{The average performance over six GLUE tasks of adaptive splitting strategies. 
	}
	\vspace{-0.1in}
	\label{fig:adaptive}
\end{figure}

\begin{table}[t]
	\centering
	\resizebox{0.48\textwidth}{!}{
		\begin{tabular}{l|ccccc}
			\textbf{Method} & \textbf{\tabincell{c}{\#Bits\\(W-E-A)}} & \textbf{\tabincell{c}{Size\\(MB)}} &
			\textbf{\tabincell{c}{Ratio\\($\downarrow$)}} &  \textbf{\tabincell{c}{SQuAD\\v1.1}} & \textbf{\tabincell{c}{MNLI\\-m}} \\ \hline
			BERT-base &   \textit{full-prec.}   &   418  & 1.0  & 80.8/88.5  & 84.6   \\
			DistilBERT      &   \textit{full-prec.}   &   250  &  1.7 &  79.1/86.9 & 81.6   \\
			LayerDrop-6L    &   \textit{full-prec.}   &   328  & 1.3  & - & 82.9 \\
			LayerDrop-3L    &   \textit{full-prec.}   &   224  & 1.9 & - & 78.6 \\ 
			TinyBERT-6L &   \textit{full-prec.}   &   55  &  7.6  & 79.7/87.5 &  82.8   \\
			ALBERT-E128     &   \textit{full-prec.}   &   45 & 9.3 & 82.3/89.3 & 81.6  \\
			ALBERT-E768     &   \textit{full-prec.}   &   120  & 3.5 & 81.5/88.6  & 82.0   \\\hline
			Quant-Noise    &   PQ & 38 & 11.0 & - & 83.6   \\
			Q-BERT & 2/4-8-8 &  53 & 7.9 &  79.9/87.5  & 83.5 \\
			Q-BERT & 2/3-8-8 &  46 & 9.1 &  79.3/87.0  & 81.8 \\
			Q-BERT & 2-8-8 & 28  & 15.0   &  69.7/79.6 &  76.6  \\
			GOBO   & 3-4-32 &   43 & 9.7 & -  &  83.7   \\
			GOBO   & 2-2-32 &   28 & 15.0 & -  &  71.0   \\
			TernaryBERT     & 2-2-8  &   28  & 15.0  & 79.9/87.4 & 83.5  \\\hline
			\textbf{BinaryBERT} & \textbf{1-1-8} & \textbf{17} & \textbf{24.6} & \textbf{80.8/88.3} & \textbf{84.2} \\
			\textbf{BinaryBERT} & \textbf{1-1-4} & \textbf{17} & \textbf{24.6} & \textbf{79.3/87.2} & \textbf{83.9} \\
		\end{tabular}
	}  
	\vspace{-0.1in}
	\caption{Comparison with other state-of-the-art methods on development set of SQuAD v1.1 and MNLI-m.}
	\vspace{-0.1in}
	\label{tab:sota}
\end{table}

\subsubsection{Adaptive Splitting}
The adaptive splitting in Section~\ref{sec:adaptive_split} supports the conversion of mixed ternary and binary precisions for more-fine-grained configurations. 
To verify its advantages, we name our approach as Maximal Gain according to Equation~\eqref{eq:dp_problem}, and compare it with two baseline strategies
i) Random Gain that randomly selects weight matrices to split; and 
ii) Minimal Gain that splits the least important modules
according to sensitivity.
We report the average score over six tasks~(QNLI, SST-2, CoLA, STS-B, MRPC and RTE) in Figure~\ref{fig:adaptive}. The end-points of 9.8MB and 16.5MB are the half-sized and full-sized BinaryBERT, respectively.
As can be seen, adaptive splitting generally outperforms the other two baselines under varying model size, indicating the effectiveness of maximizing the gain in adaptive splitting.
In Appendix~\ref{sec:detailed_adaptive}, we provide detailed performance on the six tasks, 
together with the architecture visualization of adaptive splitting.





\subsection{Comparison with State-of-the-arts}
\label{sec:compare_to_sota}
Now we compare our proposed approach with a variety of state-of-the-art counterparts, 
including Q-BERT~\citep{shen2020qbert}, GOBO~\cite{zadeh2020gobo}, Quant-Noise~\citep{fan2020training} and TernaryBERT~\citep{zhang2020ternarybert}. 
Aside from quantization, we also compare with other general compression approaches such as 
DistillBERT~\citep{sanh2019distilbert}, LayerDrop~\citep{fan2019reducing}, TinyBERT~\citep{jiao2020tinybert}, and ALBERT~\citep{lan2020albert}. 
The results are taken from the original papers, respectively.
From Table~\ref{tab:sota}, our proposed BinaryBERT has the smallest model size with the best performance among all quantization approaches. Compared with the full-precision model, our BinaryBERT retains competitive performance with a significant reduction of model size and computation. 
For example, we achieve more than $\mathbf{24\times}$ compression ratio compared with BERT-base, with only $0.4\%\downarrow$ and $0.0\%/0.2\%\downarrow$ drop on MNLI-m on SQuAD v1.1, respectively. 

\subsection{Discussion}
\label{sec:discussion}
\subsubsection{Further Improvement after Splitting}
\label{sec:further_improve}
We now demonstrate the performance gain by refining the binary model on the new architecture.
We evaluate the performance gain after splitting from a half-width ternary model (TWN$_{0.5\times}$) to the full-sized model (TWN$_{1.0\times}$) on the development set of SQuAD v1.1, MNLI-m, QNLI and MRPC. The results are shown in Table~\ref{tab:further-finetune}. As can be seen, further fine-tuning brings consistent improvement on both 8-bit and 4-bit activation. 


\begin{table}[t]
	\centering
	\resizebox{0.50\textwidth}{!}{
		\begin{tabular}{l|c|cccc}
			\textbf{\tabincell{c}{Quant}} &
			\textbf{\tabincell{c}{\#Bits\\(W-E-A)}} &
			\textbf{\tabincell{c}{SQuAD\\v1.1}} & 
			\textbf{\tabincell{c}{MNLI\\-m}} & 
			\textbf{\tabincell{c}{QNLI}} & 
			\textbf{\tabincell{c}{MRPC}} \\\hline 
			TWN$_{0.5\times}$ & 2-2-8 & 80.3/87.9 & 84.1 & 91.3 & 85.7 \\
			TWS$_{1.0\times}$ & 1-1-8 & \textbf{80.8/88.3} & \textbf{84.2} & \textbf{91.6} & \textbf{86.0} \\\hline
			TWN$_{0.5\times}$ & 2-2-4 & 78.0/86.4 & 83.7 & 90.9 & 85.5 \\
			TWS$_{1.0\times}$ & 1-1-4 & \textbf{79.3/87.2} & \textbf{83.9} & \textbf{91.4} & \textbf{86.0}
		\end{tabular}
	}
	\vspace{-0.05in}
	\caption{The performance gain by fine-tuning the binary model after splitting. $0.5\times$ and $1.0\times$ denote the half-sized and full-sized models, respectively.}
	\vspace{-0.15in}
	\label{tab:further-finetune}
\end{table}

\begin{figure}[t]
	\subfigure[8-bit Activation. \label{fig:loss_a8}]{
		\includegraphics[width=0.23\textwidth]{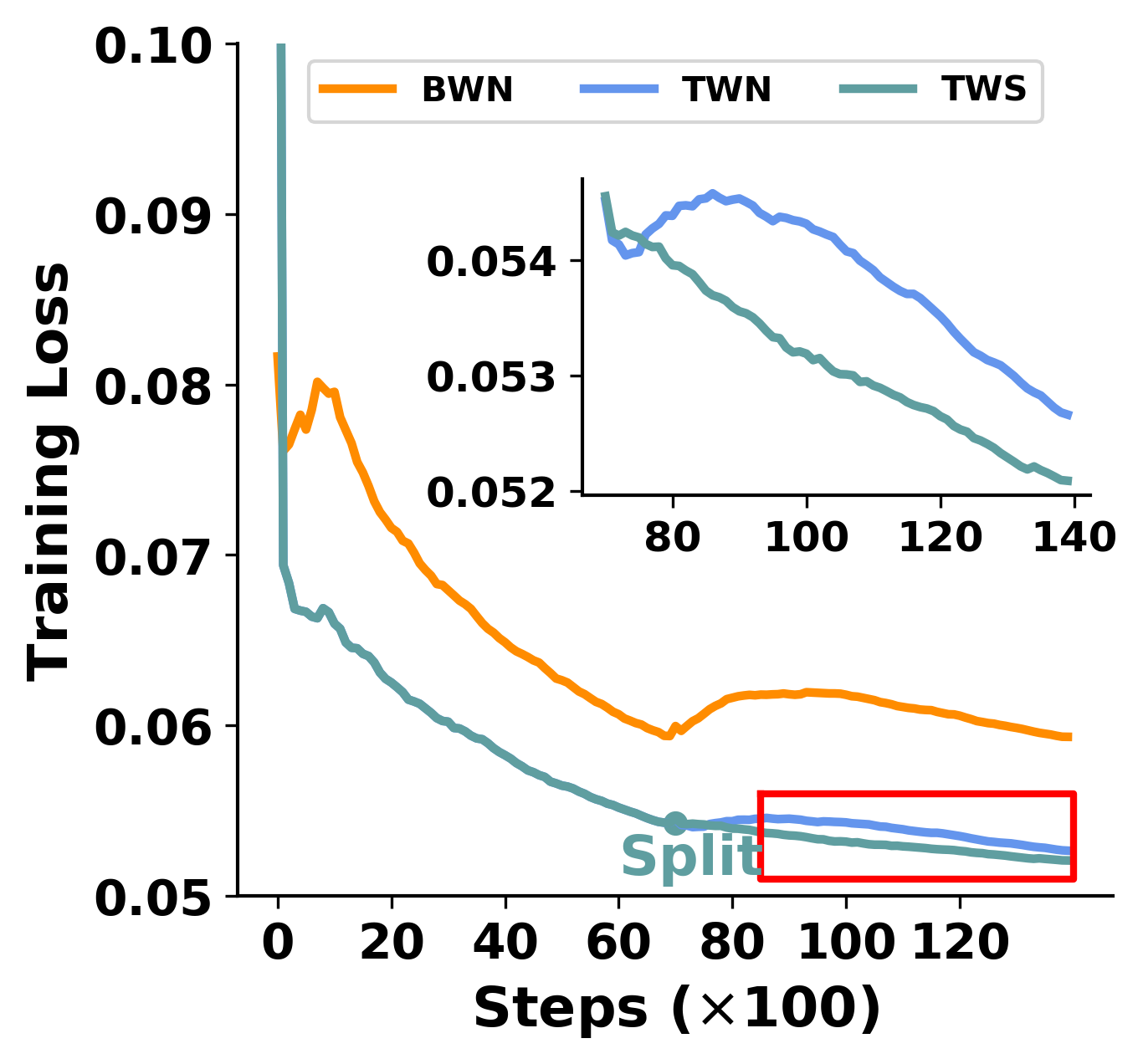}}
	\subfigure[4-bit Activation. \label{fig:loss_a4}]{
		\includegraphics[width=0.23\textwidth]{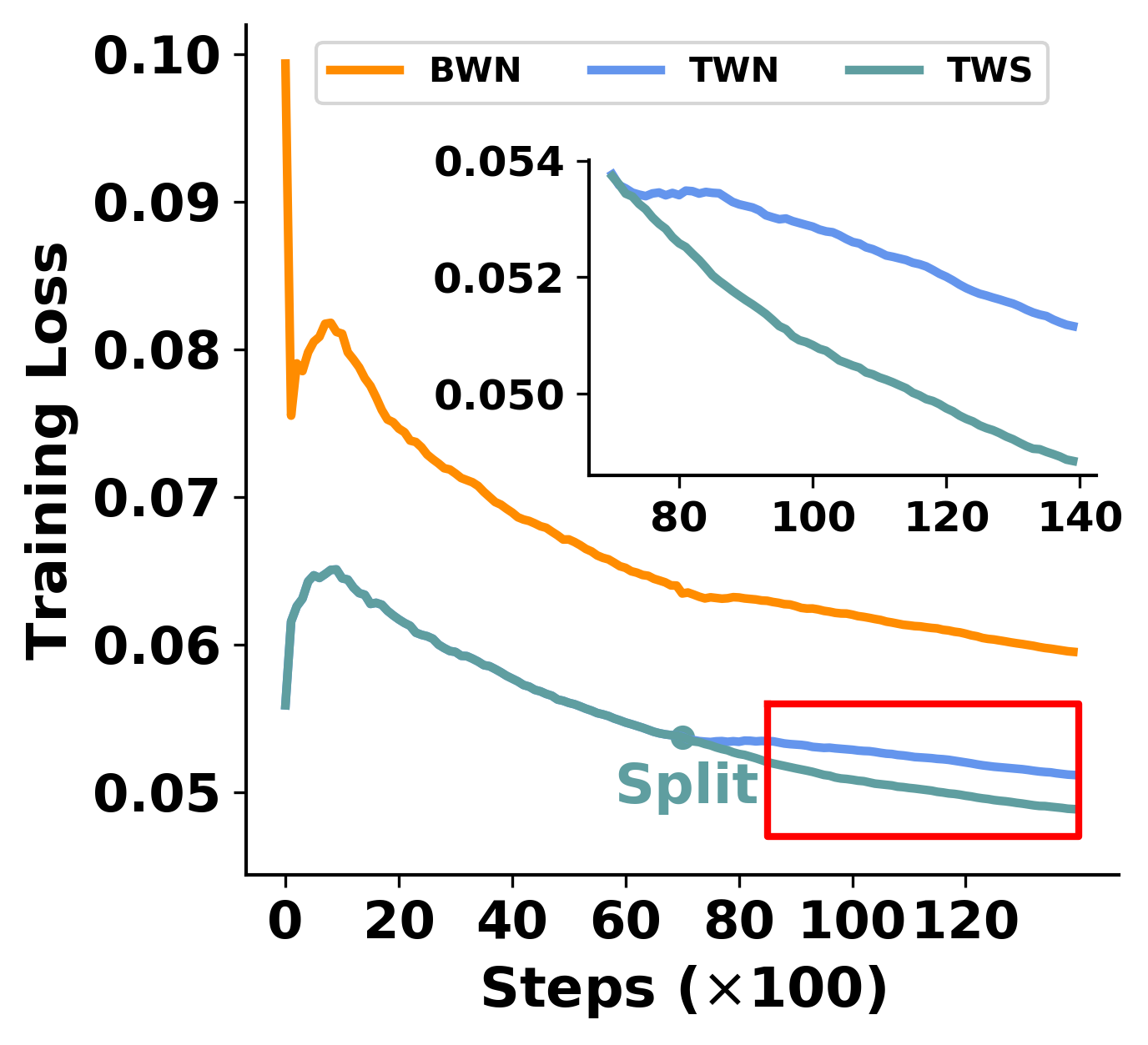}}
	\subfigure[8-bit Activation. \label{fig:traj_a8}]{
		\includegraphics[width=0.23\textwidth]{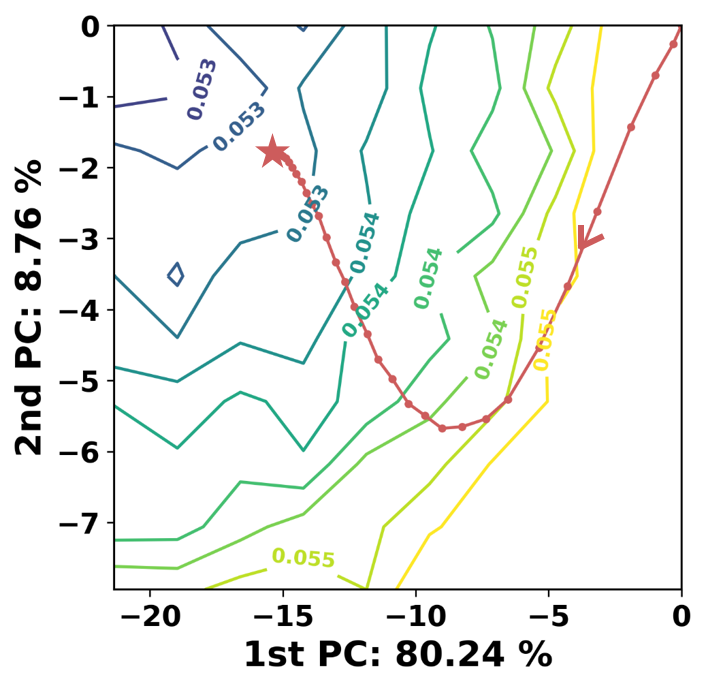}}
	\subfigure[4-bit Activation. \label{fig:traj_a4}]{
		\includegraphics[width=0.23\textwidth]{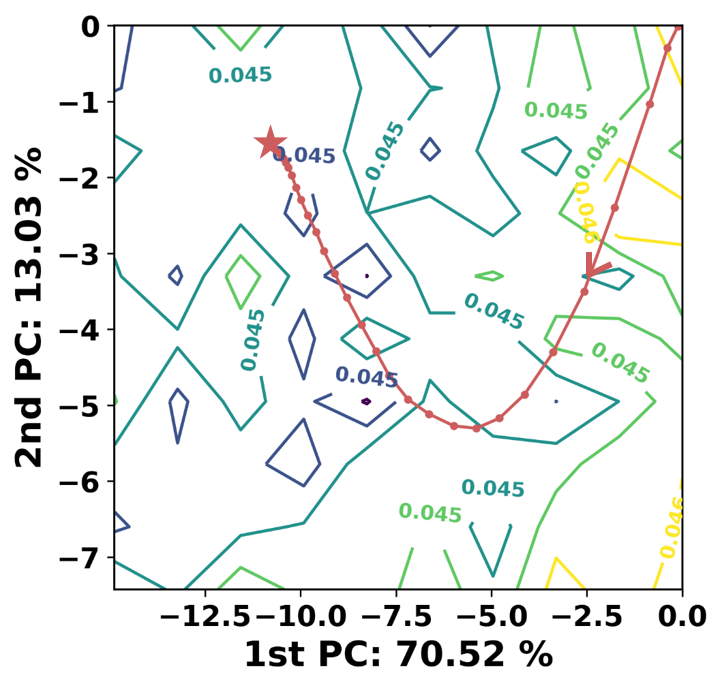}}	
	\vspace{-2ex}
	\caption{(a) and (b) show the training  curves on MRPC under different activation bits. 
	The red box is enlarged in the sub-figure. 
	(c) and (d) visualize the fine-tuning trajectories after splitting, on the 2-D loss contour of BinaryBERT.}
	\label{fig:tr_loss_curve}
	\vspace{-0.2in}
\end{figure}

\begin{table}[t]
	\centering
	\resizebox{0.47\textwidth}{!}{
		\begin{tabular}{l|c|ccccc}
			\textbf{\tabincell{c}{Quant}} &
			\textbf{\tabincell{c}{\#Bits\\(W-E-A)}} &
			\textbf{\tabincell{c}{SQuAD\\v1.1}} & 
			\textbf{\tabincell{c}{MNLI\\-m}} & \textbf{\tabincell{c}{QNLI}} & \textbf{\tabincell{c}{SST-2}} \\\hline 
			BWN & 1-1-8 & 79.2/86.9 & \textbf{84.2} & 91.2 & 92.7 \\
			LAB & 1-1-8 & 79.0/87.0 & 83.6 & 91.5 & 92.8 \\
			BiReal & 1-1-8  & 79.4/87.1 & 83.9 & 91.4 & 92.5 \\
			BWN$\dagger$ & 1-1-8  & 79.4/87.3 & \textbf{84.2} & 91.3 & 92.8  \\
			BWN$\ddagger$ & 1-1-8  & 79.6/87.2 & 83.5 & 91.2 & 92.9 \\
			TWS & 1-1-8 & \textbf{80.8/88.3} & \textbf{84.2} & \textbf{91.6} & \textbf{93.2} \\\hline
			BWN & 1-1-4 & 77.5/85.8 & 83.5 & 91.2 & 92.5 \\
			LAB & 1-1-4 & 76.7/85.5 & 83.3 & 91.3 & 92.9 \\
			BiReal  & 1-1-4 & 76.9/85.4 & 83.4 & 91.0 & 92.8 \\
			BWN$\dagger$ & 1-1-4  & 78.2/86.2 & 83.6 & 91.3 & 92.9 \\
			BWN$\ddagger$ & 1-1-4  & 78.3/86.5 &  83.1 & 90.9 & 92.9  \\
			TWS & 1-1-4 & \textbf{79.3/87.2} &  \textbf{83.9} & \textbf{91.4} & \textbf{93.7}
		\end{tabular}
	}
	\vspace{-0.05in}  
	\caption{Comparison with other binarization methods.}
	\label{tab:more_binarization}
\end{table}

%

\paragraph{Training Curves.}
Furthermore, we plot the training loss curves of BWN, TWN and our TWS on MRPC with data augmentation in Figures~\ref{fig:loss_a8} and ~\ref{fig:loss_a4}. 
Since TWS cannot inherit the previous optimizer due to the architecture change, we reset the optimizer and learning rate scheduler of BWN, TWN and TWS for a fair comparison, despite the slight increase of loss after splitting. 
We find that our TWS attains much lower training loss than BWN, and also surpasses TWN, verifying the advantages of fine-tuning on the wider architecture. 

\paragraph{Optimization Trajectory.}
We also follow~\cite{li2018visualizing,hao2019visualizing} to visualize the optimization trajectory after splitting in Figures~\ref{fig:traj_a8} and ~\ref{fig:traj_a4}. We calculate the first two principal components of parameters in the final BinaryBERT, which are the basis for the 2-D plane. The loss contour is thus obtained by evaluating each grid point in the plane. It is found that the binary models are heading towards the optimal solution for both 8/4-bit activation quantization on the loss contour.


\subsubsection{Exploring More Binarization Methods}
We now study if there are any improved binarization variants that can directly bring better performance.
Aside from BWN, we compare with LAB~\cite{hou2017loss} and BiReal~\cite{liu2018bi}. Meanwhile, we compare with gradual quantization, i.e., BWN training based on a ternary model, denoted as BWN$\dagger$. Furthermore, we also try the same scaling factor of BWN with TWN to make the precision change smooth, dubbed as BWN$\ddagger$. 
From Table~\ref{tab:more_binarization}, we find that our TWS still outperforms various binarization approaches in most cases, suggesting the superiority of splitting in finding better minima than direct binary training.

\section{Related Work}

Network quantization has been a popular topic with vast literature in efficient deep learning.
Below we give a brief overview for three research strands: network binarization, mixed-precision quantization and neuron splitting, all of which are related to our proposed approach.

\subsection{Network Binarization}
Network binarization achieves remarkable size reduction and is widely explored in computer vision. 
Existing binarization approaches can be categorized into  quantization error minimization~\citep{rastegari2016xnor,hou2017loss,zhang2018lq}, improving training objectives~\citep{martinez2020training,bai2020few} and reduction of gradient mismatch~\citep{bai2018proxquant,liu2018bi,liu2020reactnet}.
Despite the empirical success of these approaches in computer vision, there is little exploration of binarization in natural language processing tasks. Previous works on BERT quantization~\citep{zafrir2019q8bert,shen2020qbert,zhang2020ternarybert} push down the bit-width to as low as two, but none of them achieves binarization. 
On the other hand, our work serves as the first attempt to binarize the pre-trained language models.

\subsection{Mixed-precision Quantization}
Given the observation that neural network layers exhibit different sensitivity to quantization~\citep{dong2019hawq,wang2019haq}, mixed-precision quantization re-allocate layer-wise quantization bit-width for higher compression ratio. 
Inspired by neural architecture search~\citep{liu2019darts,wang2020revisiting}, common approaches of mixed-precision quantization are primarily based on differentiable search~\citep{wu2018mixed,li2020efficient}, reinforcement learning~\citep{wu2018pocketflow,wang2019haq}, or simply loss curvatures~\cite{dong2019hawq,shen2020qbert}.
While mixed-precision quantized models usually demonstrate better performance than traditional methods under the same compression ratio, 
they are also harder to deploy~\citep{habi2020hmq}.
On the contrary, BinaryBERT with adaptive splitting enjoy both the good performance from the mixed precision of ternary and binary values, and the easy deployment given the consistent arithmetic precision.

There are also works on binary neural architecture search~\citep{kim2020learning,bulat2020bats} which have a similar purpose to mixed-precision quantization. 
Nonetheless, such methods are usually time-consuming to train and are prohibitive for
large pre-trained language models.


\subsection{Neuron Splitting}
Neuron splitting is originally proposed to accelerate the network training, by progressively increasing the width of a network~\citep{chen2016net2net,wu2019splitting}. The split network equivalently inherits the knowledge from the antecessors and is trained for further improvement.
Recently, neuron splitting is also studied in  quantization~\citep{zhao2019improving,kim2019binaryduo}.
By splitting neurons with large magnitudes, the full-precision outliers are removed and thus the quantization error can be effectively reduced~\cite{zhao2019improving}.
\citet{kim2019binaryduo} apply neuron splitting to decompose ternary activation into two  binary activations based on bias shifting of the batch normalization layer.
However, such a method cannot be applied in BERT as there is no batch normalization layer. Besides, weight splitting is much more complex due to the equivalence constraint on both the quantized and latent full-precision weights.


\section{Conclusion}

In this paper, we propose BinaryBERT, pushing BERT quantization to the limit. 
As a result of the steep and complex loss landscape, we find directly training a BinaryBERT is hard with a large performance drop. We thus propose a ternary weight splitting that splits a trained ternary BERT to initialize BinaryBERT, followed by fine-tuning for further refinement. Our approach also supports adaptive splitting that can tailor the size of BinaryBERT based on the edge device constraints.
Empirical results show that our approach significantly outperforms vanilla binary training, achieving state-of-the-art performance on BERT compression.
\section*{Acknowledgement}
This work was partially supported by the National Key Research and Development Program of China (No. 2018AAA0100204), 
and Research Grants Council of the Hong Kong Special Administrative Region, China (No. CUHK 14210717 of the General Research Fund). We sincerely thank all anonymous reviewers for their insightful suggestions. 

\clearpage
\bibliography{acl2020}
\bibliographystyle{acl_natbib}

\clearpage
\appendix
\section{Derivation of Equation \eqref{eq:solution}}
\label{sec:appendix:derive}
In this section, we show the derivations to obtain $a$ and $b$.  
Recall the BWN quantizer introduced in Section~\ref{sec:difficulty}, we have 
\[
\hat{w}_{1,i}^b  = \alpha_1 \textrm{sign}(w_{1,i}^b),
\]
where 
\[
\alpha_1 = \frac{1}{n} \big[\sum_{i\in \mathcal{I}} |a w_i^t| + \sum_{i\in \mathcal{J}} | w^t_j + b| + \sum_{i\in \mathcal{K}} |b| \big].
\]
Similarly, 
\[
\hat{w}_{2,i}^b  = \alpha_2 \textrm{sign}(w_{2,i}^b),
\]
where 
\[
\alpha_2\! =\! \frac{1}{n} \big[\sum_{i\in \mathcal{I}} |(1-a) w_i^t| + \sum_{j\in \mathcal{J}} |-b| + \sum_{k\in \mathcal{K}} |w^t_k - b| \big].
\]
According to $\hat{\m w}^t = \hat{\m w}^b_1 + \hat{\m w}^b_2$, for those $\hat{w}_i^t = \hat{w}_{1,i}^b + \hat{w}_{2,i}^b = 0$, we have
\begin{align}
& \frac{1}{n} \big[\sum_{i\in \mathcal{I}} |a w_i^t| + \sum_{j\in \mathcal{J}} |w^t_j + b| + \sum_{k\in \mathcal{K}} |b| \big] \nonumber \\
& = \frac{1}{n} \big[\sum_{i\in \mathcal{I}} |(1\!-\!a) w_i^t| \!+\! \sum_{j\in \mathcal{J}} |-b|\! + \!\sum_{k\in \mathcal{K}} |w^t_k \!-\! b| \big]. \nonumber
\end{align}
By assuming $0<a<1$ and $b>0$, this can be further simplified to 
\begin{equation}
a \sum_{i\in \mathcal{I}} |w_i^t| + \sum_{j\in \mathcal{J}} |w^t_j| = (1-a)\sum_{i\in \mathcal{I}} |w_i^t| + \sum_{k\in \mathcal{K}} |w^t_k|, \nonumber
\end{equation}
which gives the solution of $a$ as 
\begin{equation*}
    a = \frac{\sum_{i\in\mathcal{I}} |w^t_i| + \sum_{j\in \mathcal{J}} |w^t_j| - \sum_{k\in \mathcal{K}}|w^t_k| }{2\sum_{i\in\mathcal{I}} |w^t_i|}.
\end{equation*}
We empirically find the solution satisifies $0<a<1$. For $\hat{w}_i^t 
\neq 0$, from $\hat{w}_i^t = \hat{w}_{1,i}^b + \hat{w}_{2,i}^b$, we have
\begin{align}
    & \frac{1}{|\mathcal{I}|} \sum_{i \in \mathcal{I}} |w_i^t| = \alpha_1 + \alpha_2 \nonumber \\
    & = \frac{1}{n} \big[\sum_{i\in \mathcal{I}} |a w_i^t| + \sum_{j\in \mathcal{J}} |w^t_j + b| + \sum_{k\in \mathcal{K}} |b| \big] \nonumber\\
    & + \frac{1}{n} \big[\sum_{i\in \mathcal{I}} |(1\!-\!a) w_i^t| + \sum_{j\in \mathcal{J}} |-b| \!+ \!\sum_{k\in \mathcal{K}} |w^t_k\! -\! b| \big] \nonumber\\
    & = \frac{1}{n} \big[ \sum_{i\in \mathcal{I}} |w_i^t| + \sum_{j\in \mathcal{J}} |w^t_j| +  \sum_{k\in\mathcal{K}} |w^t_k| \nonumber \\ 
    & + 2\sum_{j\in \mathcal{J}}|b| + 2\sum_{k\in \mathcal{K}}|b| \big] \nonumber \\
    & = \frac{1}{n} \big[ \sum_{i=1}^n |w_i^t| + 2(|\mathcal{J}|+|\mathcal{K}|)\cdot b \big]. \nonumber
\end{align}
Thus the solution for $b$ is
\begin{equation*}
    b = \frac{\frac{n}{|\mathcal{I}|} \sum_{i\in \mathcal{I}} | w^t_i| - \sum_{i=1}^n |w_i^t|  }{2 (|\mathcal{J}|+|\mathcal{K}|) },
\end{equation*}
which satisfies $b>0$.

\section{Implementation Details}
\label{sec:appendix_implementation}

\subsection{Detailed Procedure of Adaptive Splitting}
\label{sec:sensitivity}
As mentioned in Section~\ref{sec:adaptive_split}, the adaptive splitting requires to first estimate the quantization sensitivity vector $\m u$. 
We study the sensitivity in two aspects: the Transformer parts, and the Transformer layers.
For Transformer parts, we follow the weight categorization in Section~\ref{sec:challenges}: MHA-Q/K, MHA-V, MHA-O, FFN-Mid and FFN-Out.
For each of them, we compare the performance gap between quantizing and not quantizing that part (e.g., MHA-V), while leavging the rest parts all quantized (e.g., MHA-Q/K, MHA-O, FFN-Mid and FFN-Out).
Similarly, for each Transformer layer, we quantize all layers but leave the layer under investigation un-quantized, and calculate the performance gain compared with the fully qauntized baseline.
The performance gain of both Transformer parts and layers are shown in Figure~\ref{fig:sensitivity}.
As can be seen, for Transformer parts, the FFN-Mid and MHA-Q/K rank in the first and second place. 
In terms of Transformer layers, shallower layers are more sensitive to quantization than the deeper ones.

However, the absolute performance gain may not reflect the quantization sensitivity directly, since Transformer parts have different number of parameters. Therefore, we divide the performance gain by the number of parameters in that part or layer to obtain the parameter-wise performance gain. We are thus able to measure the quantization sensitivity of the $i$th Transformer part in the $j$th Transformer layer by summing their parameter-wise performance gain together.
We also apply the same procedure to word embedding and pooler layer to otain their sensitivity scores.

We are now able to solve Equation~\eqref{eq:dp_problem} by dynamic programming. The combinatorial optimization can be viewed as a knapsack problem, where the constraint $\mathcal{C}-\mathcal{C}_0$ is the volume of the knapsack, and the sensitivity scores $\m u$ are the item values.


\subsection{Hyper-parameter Settings}
\label{sec:hyper}
We first perform the two-stage knowledge distillation, i.e., intermediate-layer distillation (Int. Dstil.) and prediction-layer distillation (Pred. Dstil.) on the ternary model, and then perform ternary weight splitting followed by fine-tuning (Split Ft.) with only prediction-layer distillation after the splitting. 
The initial learning rate is set as $5\times 10^{-5}$ for the intermediate-layer distillation, and $2\times 10^{-5}$ for the prediction-layer distillation, both of which linearly decay to 0 at the end of training.
We conduct experiments on GLUE tasks both without and with data augmentation~(DA) except for MNLI and QQP due to their limited performance gain. The running epochs for MNLI and QQP are set to 3, and 6 for the rest tasks if without DA and 1 otherwise.
For the rest hyper-parameters, we follow the default setting in~\cite{devlin2019bert}. 
The detailed hyper-parameters are summarized in Table~\ref{tbl:hyperparam}.

\begin{table}[t]
	\centering
	\resizebox{0.5\textwidth}{!}{
		\begin{tabular}{l|c|c|c}
		 \hline
         & \multicolumn{3}{c}{BinaryBERT} \\ \cline{2-4}
         &  \tabincell{c}{Int. Dstil. \\(Ternary)} &  \tabincell{c}{Pred. Dstil. \\(Ternary)} & \tabincell{c}{Split Ft.\\(Binary)}  \\ \hline
         Batch Size     & 32  & 32  & 32 \\
         Sequence Length  & 128 & 128 & 128 \\
         Learning rate~(LR)   & 5e-5 &  2e-5  &   2e-5   \\
         LR Decay    & Linear  &   Linear & Linear  \\
         Warmup portion    &  0.1  & 0.1  & 0.1  \\
         Weight Decay    &  1e-2 & 1e-2 & 1e-2  \\
         Gradient Clipping &  1  & 1  & 1  \\
         Dropout &     0.1   &  0.1 &  0.1 \\
         \tabincell{l}{Epochs w/o DA\\\quad -other dataserts}  &  6 & 6 & 6  \\ 
         \tabincell{l}{Epochs w DA\\\quad -other dataserts}  &  1 & 1 & 1  \\
         \tabincell{l}{Epochs w/o DA\\\quad -MNLI, QQP}   &   3  & 3 & 3  \\\hline
		\end{tabular}
	}
	\caption{Hyper-parameters for training BinaryBERT on the GLUE benchmark at different stages.}
	\label{tbl:hyperparam}
\vspace{-0.15in}
\end{table}

\section{More Empirical Results}
\label{sec:appendix:more_results}

\begin{figure}[t]
    \centering
    \subfigure[Transformer Parts.]{
	    \includegraphics[width=0.48\textwidth]{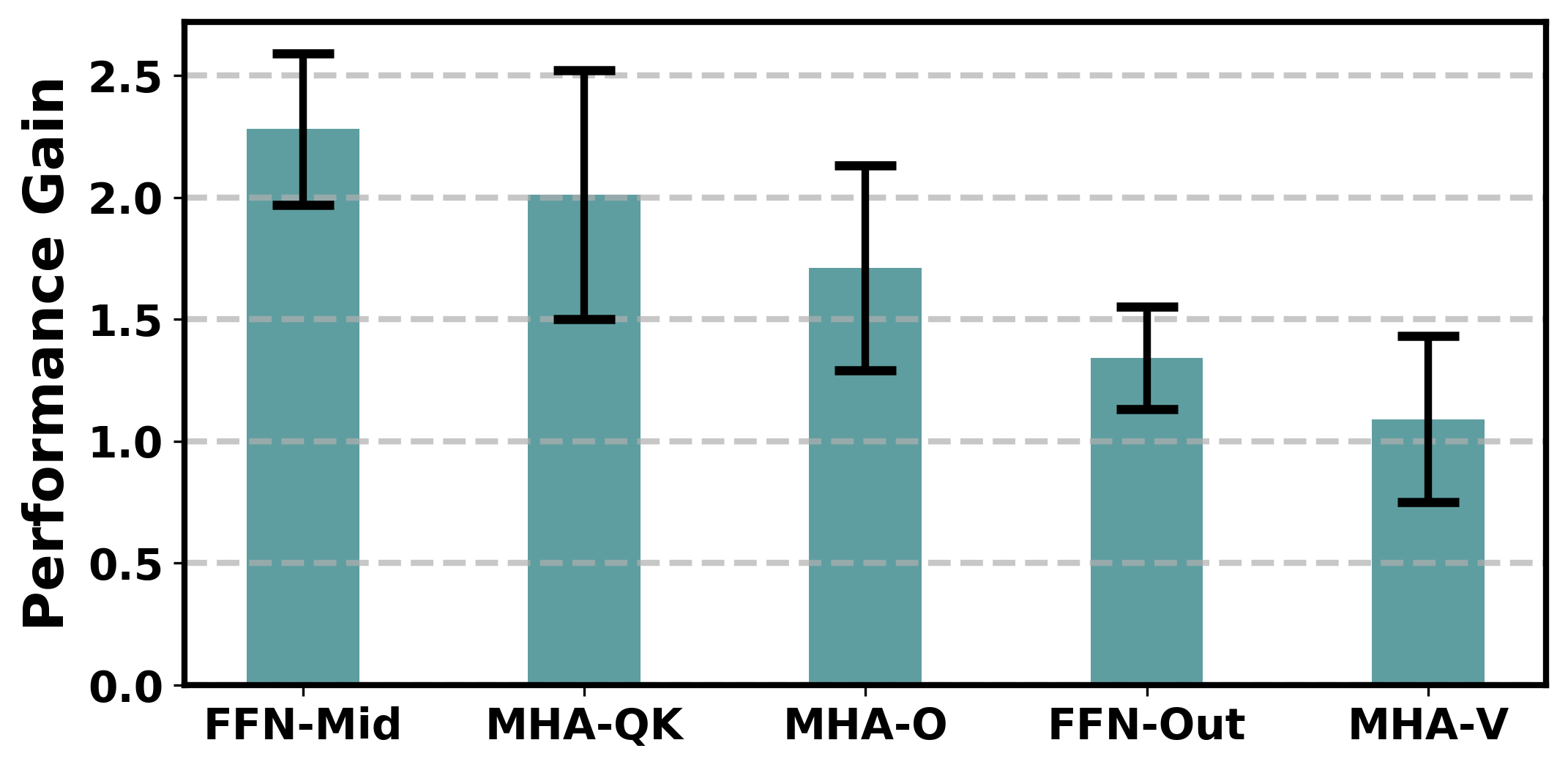}
	    \label{fig:part_sensitivity}
	}
    \subfigure[Transformer Layers.]{
	    \includegraphics[width=0.48\textwidth]{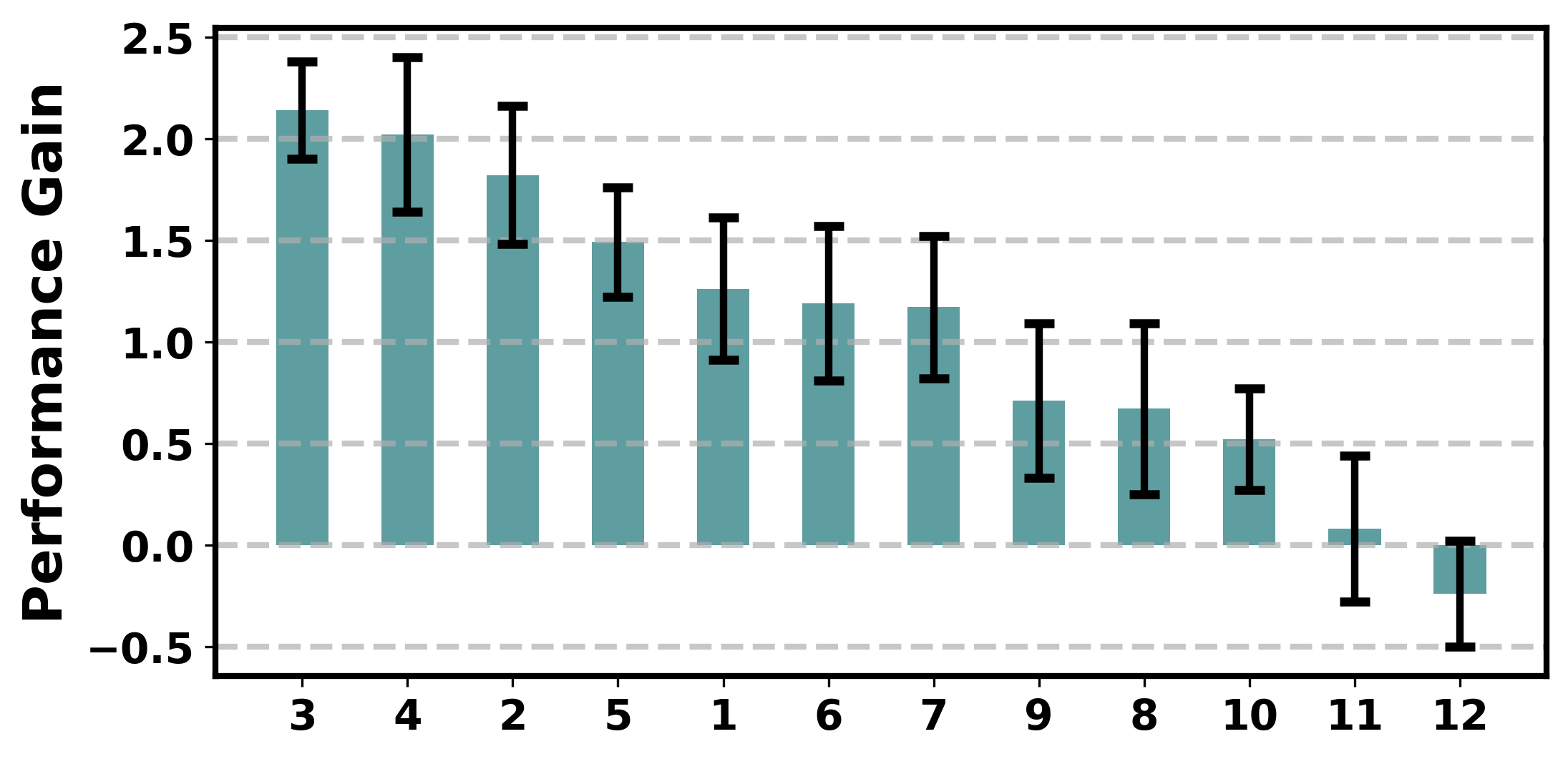}
	    \label{fig:layer_sensitivity}
	}
	\vspace{-0.15in}
	\caption{The performance gain of different Transformer parts and layers in descending order. 
	All numbers are averaged by 10 random runs with standard deviations reported.}
	\label{fig:sensitivity}
	\vspace{-0.15in}
\end{figure}

\subsection{Performance Drop by Binarization}
\label{sec:more_drop}

\begin{figure*}[t]
\vspace{-0.2in}
    \subfigure[MNLI-m.]{
	    \includegraphics[width=0.15\textwidth]{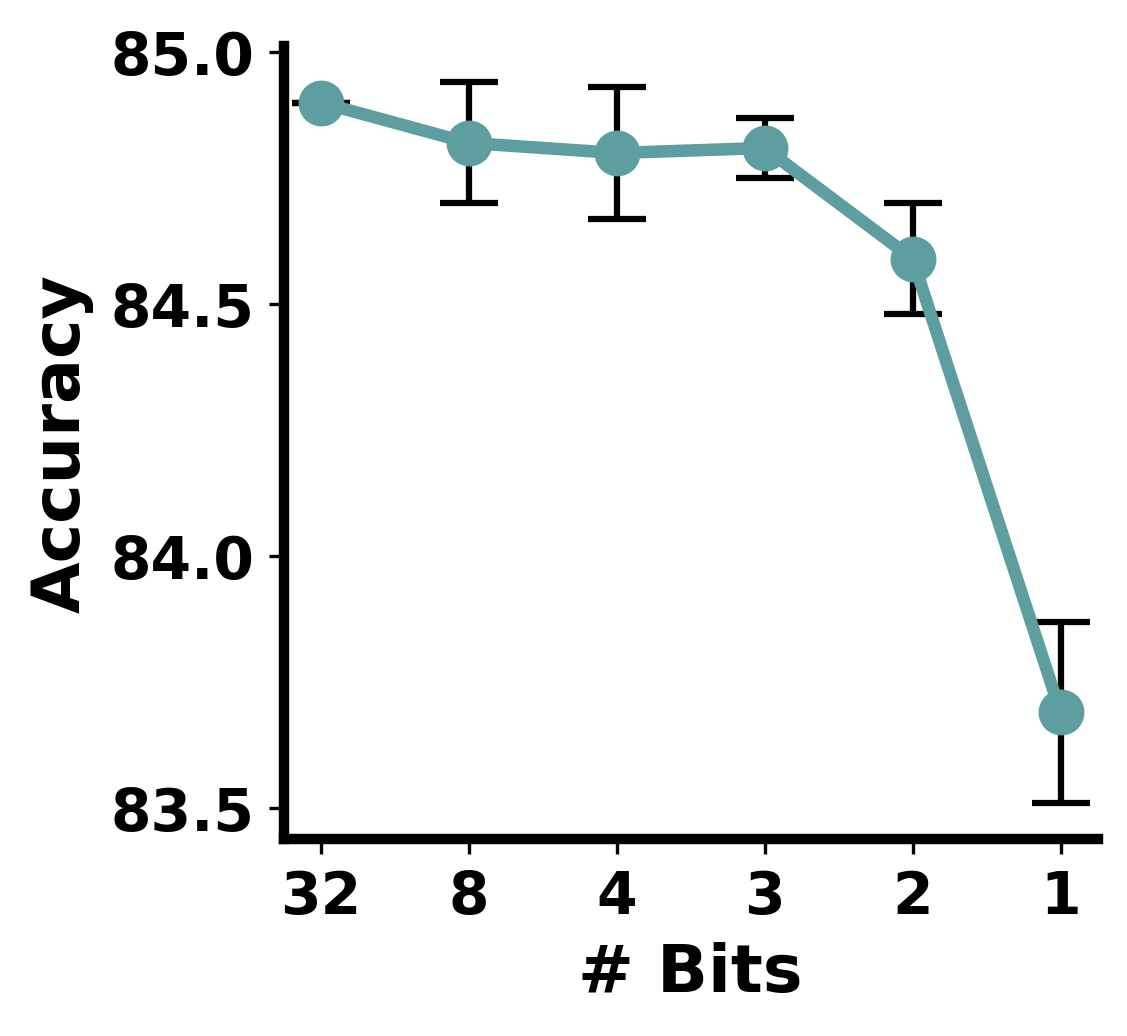}
	}
    \subfigure[SST-2.]{
	    \includegraphics[width=0.15\textwidth]{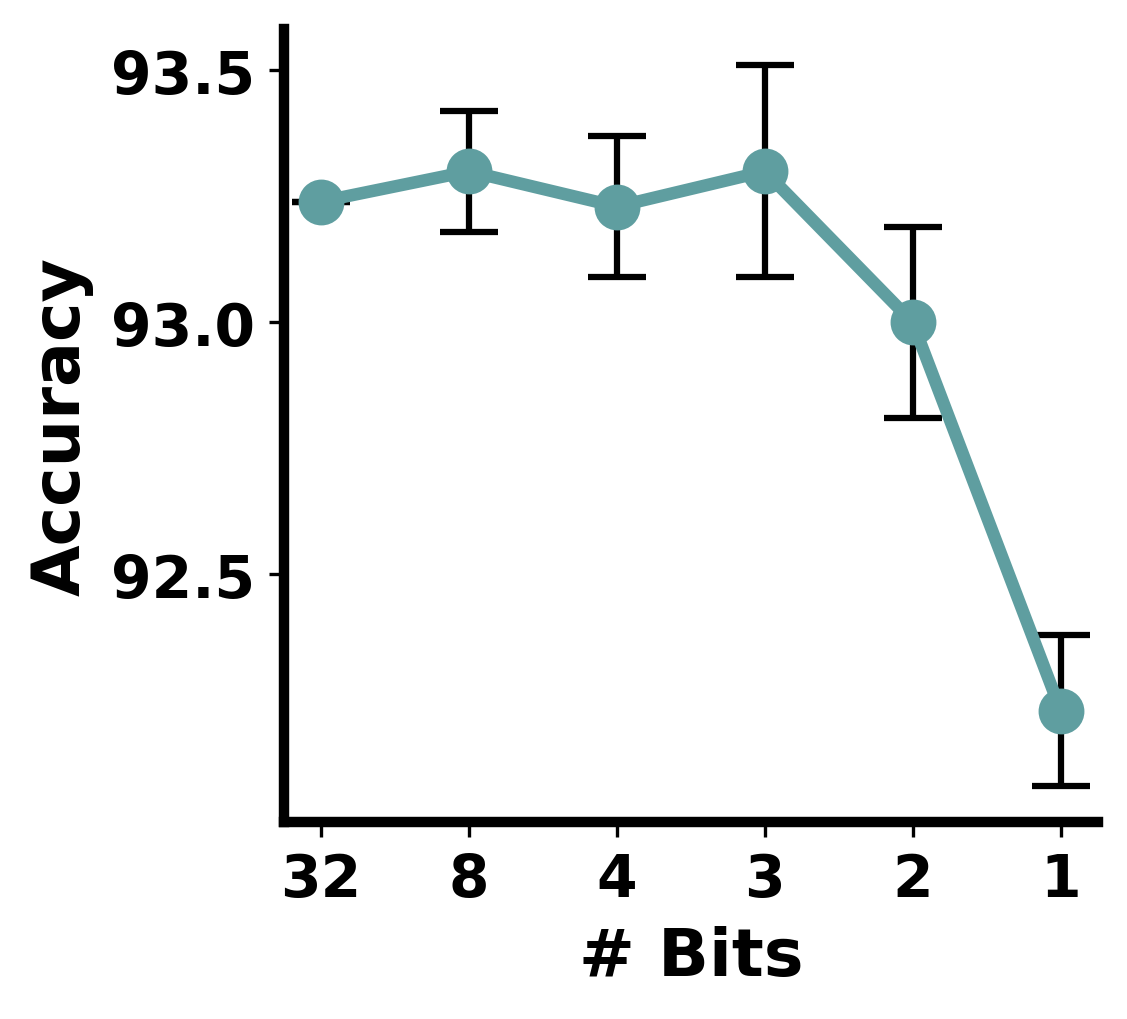}
	}
    \subfigure[CoLA.]{
	    \includegraphics[width=0.15\textwidth]{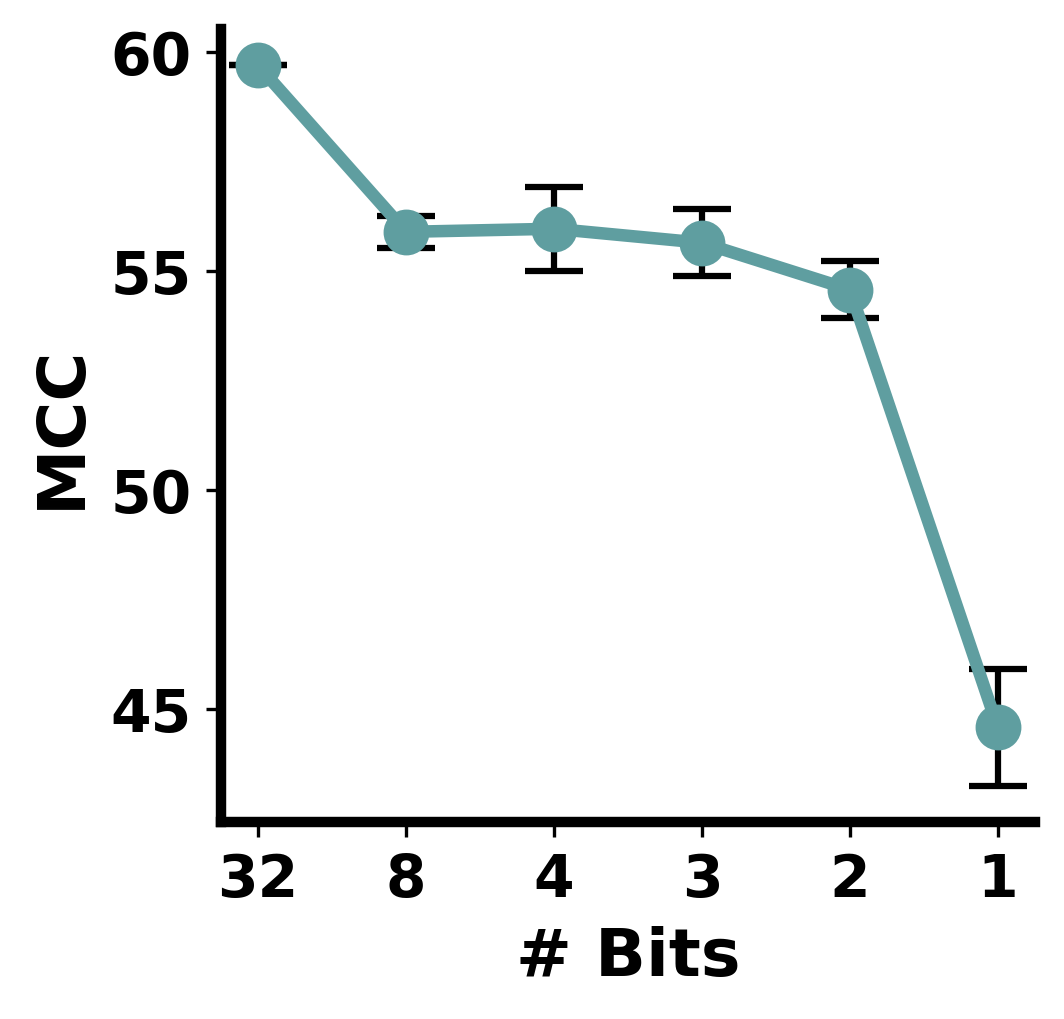}
	}
	\subfigure[STS-B.] { 
		\includegraphics[width=0.15\textwidth]{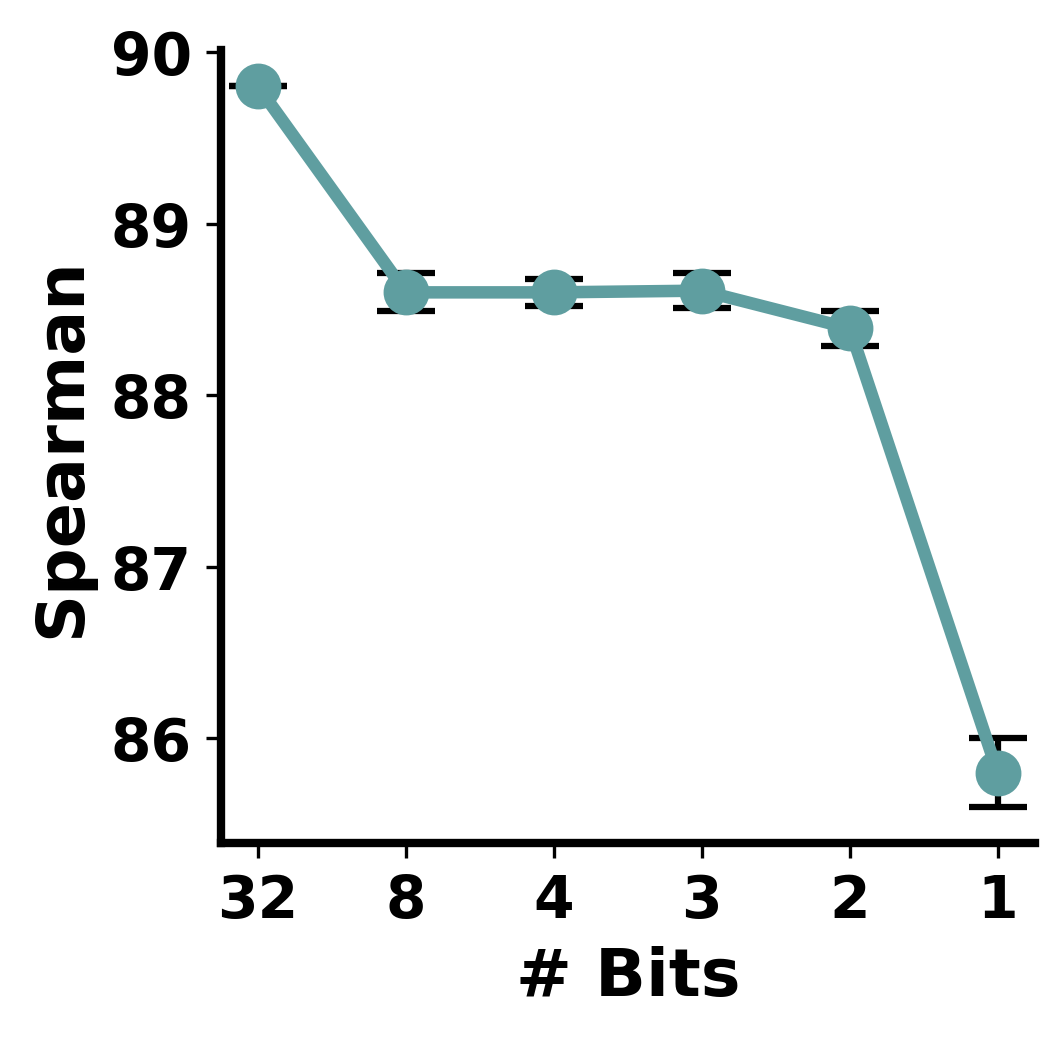}
	}
	\subfigure[MRPC.] { 
		\includegraphics[width=0.15\textwidth]{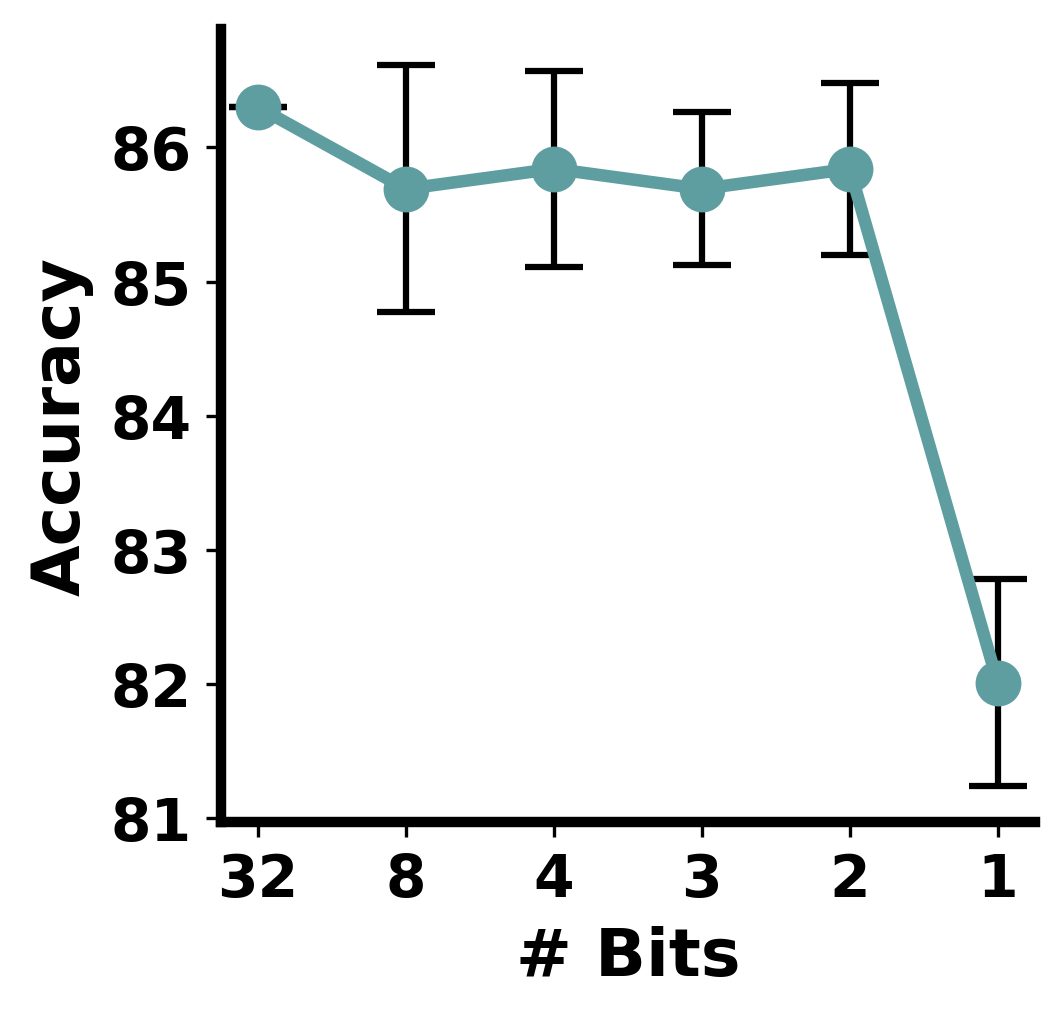}
	}
	\subfigure[RTE.] { 
		\includegraphics[width=0.15\textwidth]{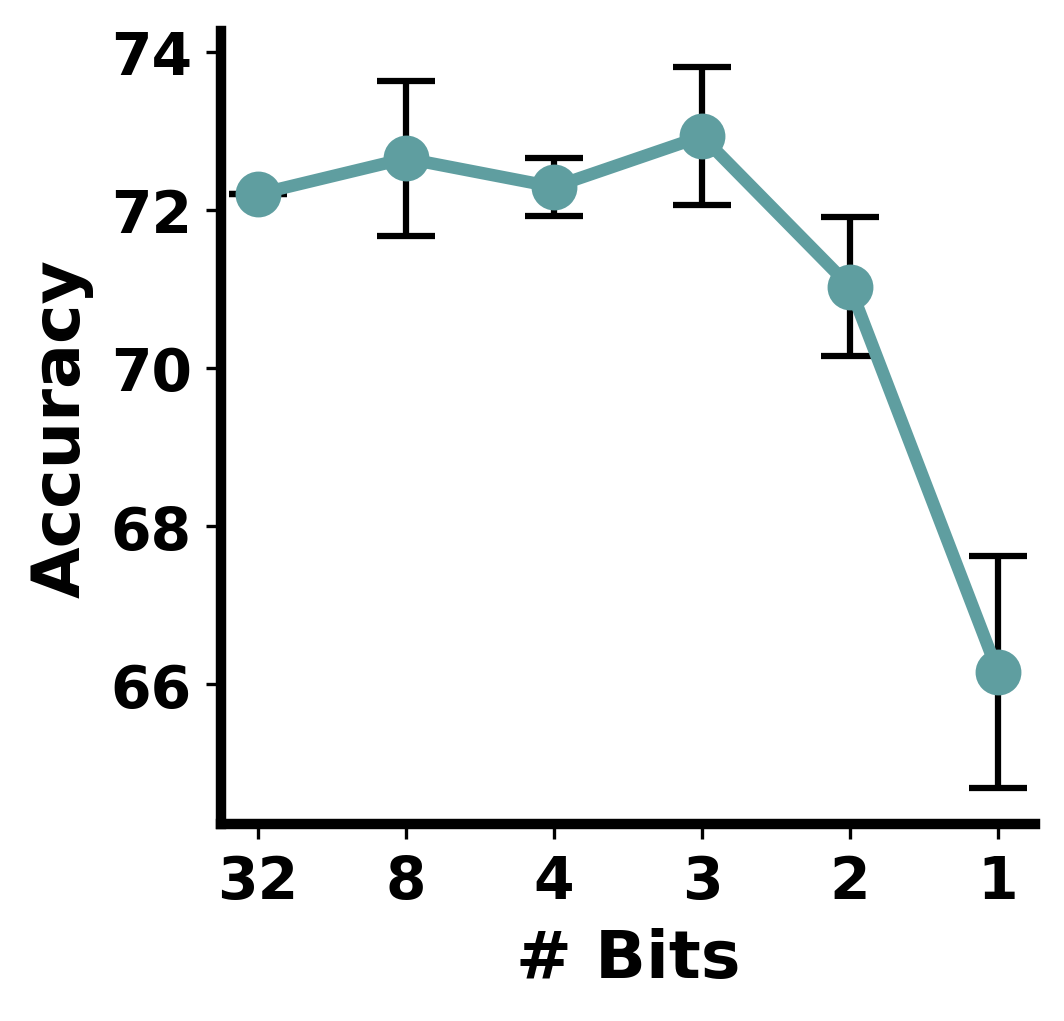}
	}	
	\vspace{-0.15in}
	\caption{Performance of quantized BERT with different weight bits and 8-bit activation on the GLUE Benchmarks. The results are obtained from 10 random seeds except for MNLI with 3 seeds. }
	\label{fig:acc_drop_a8}
\end{figure*}

\begin{figure*}[t]
    \subfigure[MNLI-m.]{
	    \includegraphics[width=0.15\textwidth]{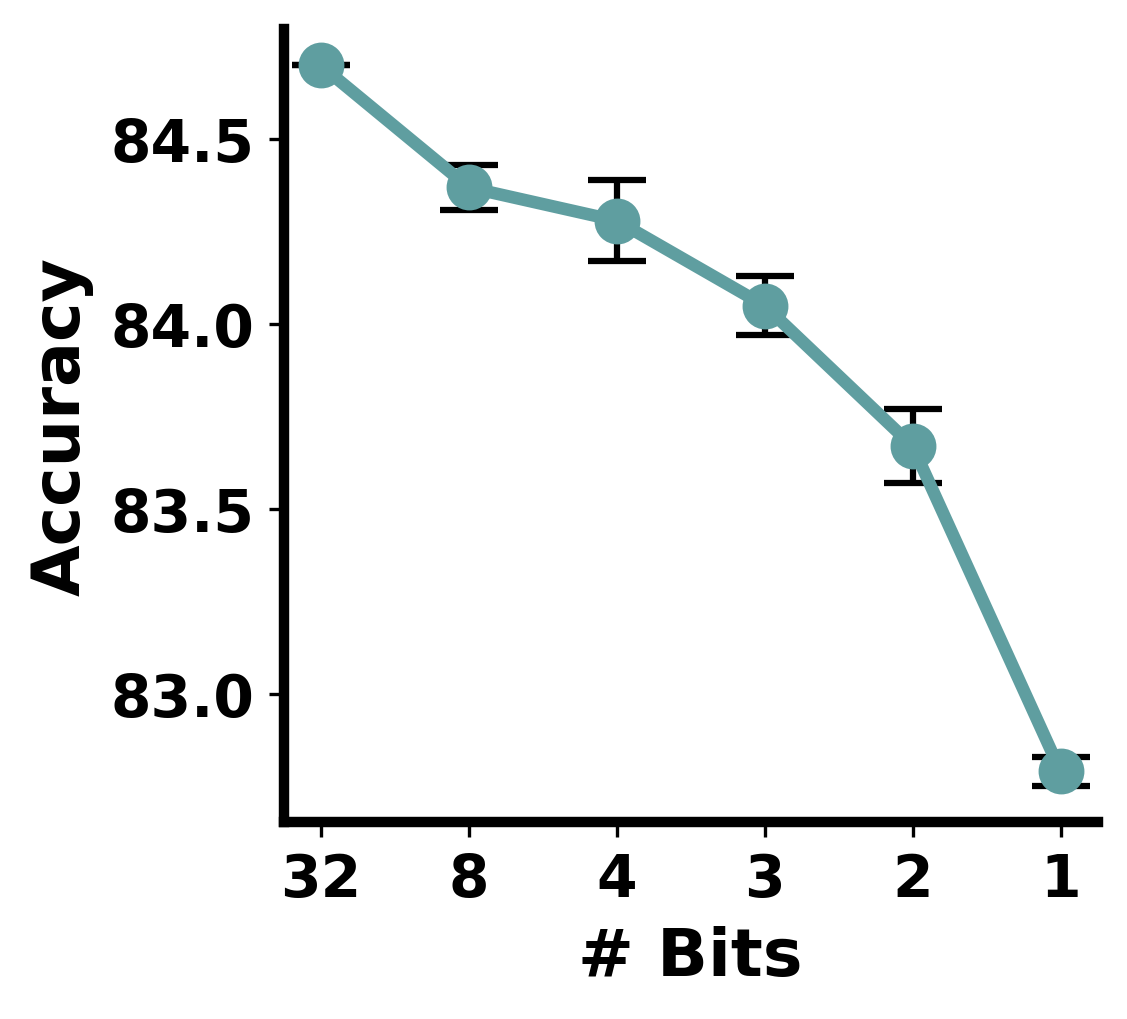}
	}
    \subfigure[SST-2.]{
	    \includegraphics[width=0.15\textwidth]{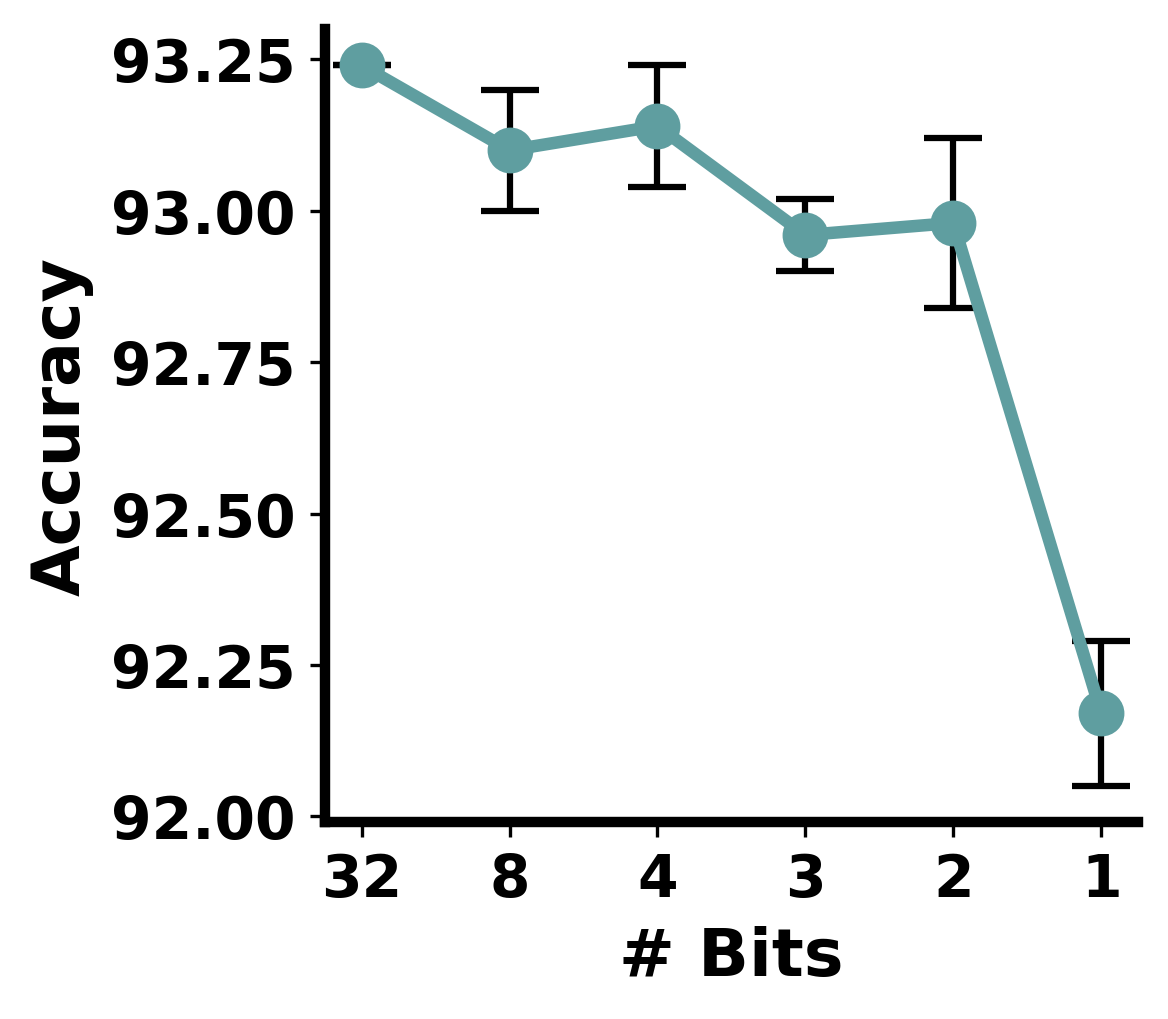}
	}
    \subfigure[CoLA.]{
	    \includegraphics[width=0.15\textwidth]{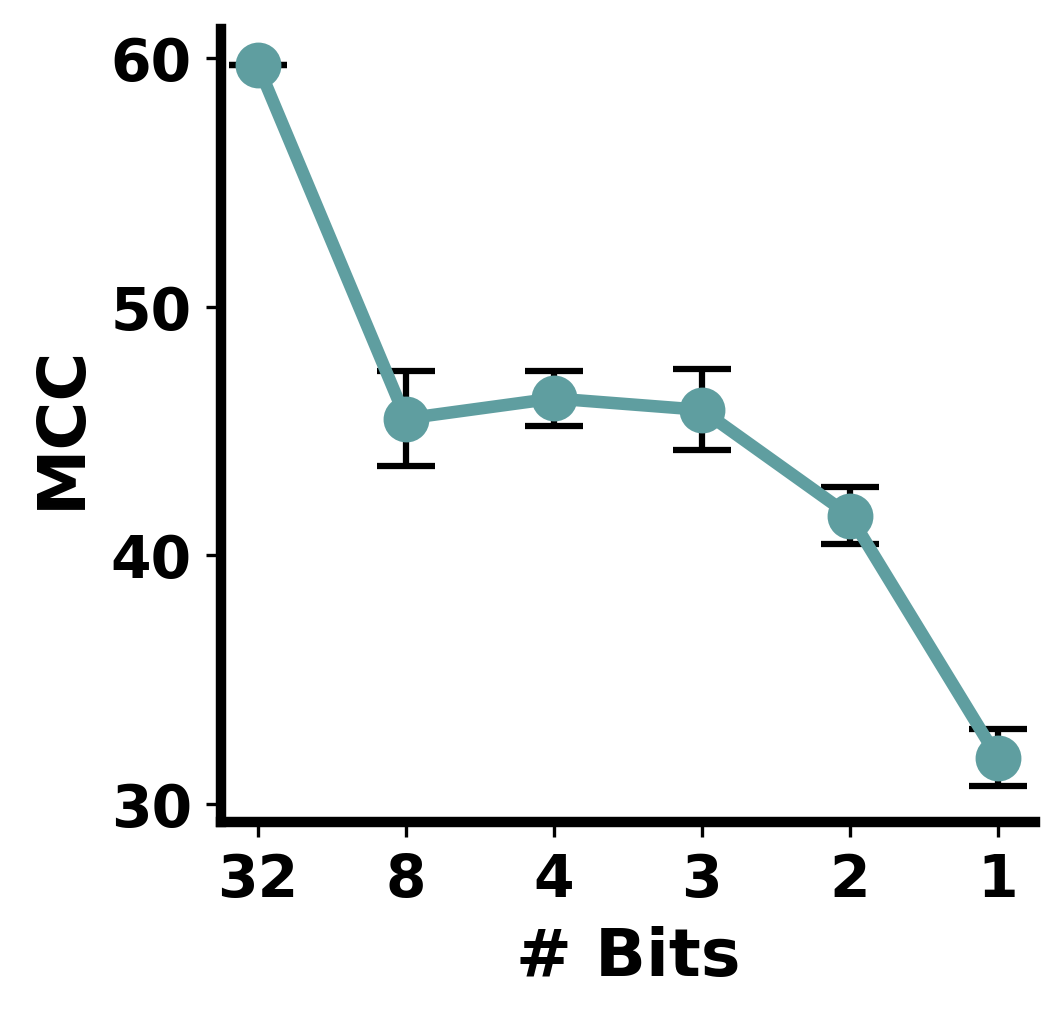}
	}
	\subfigure[STS-B.] { 
		\includegraphics[width=0.15\textwidth]{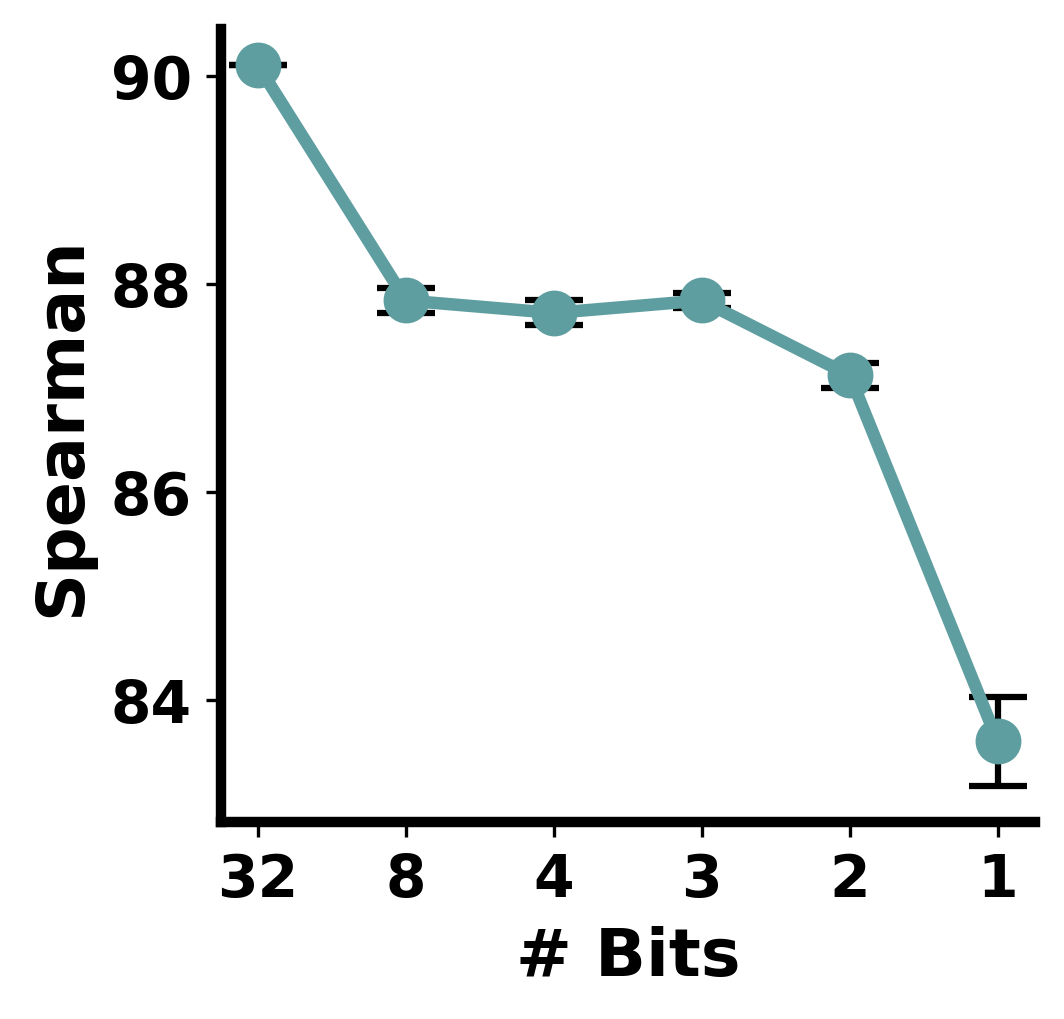}
	}
	\subfigure[MRPC.] { 
		\includegraphics[width=0.15\textwidth]{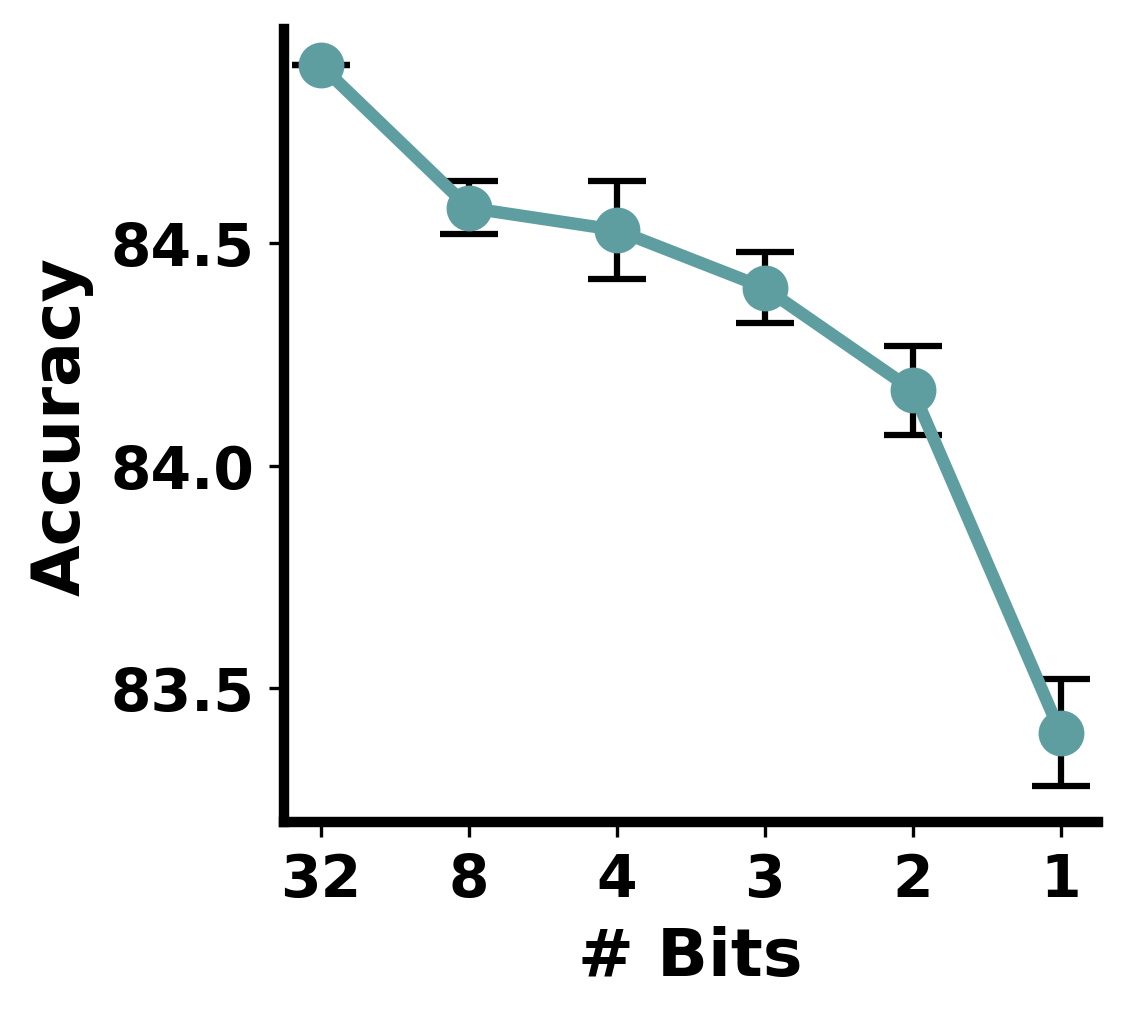}
	}
	\subfigure[RTE.] { 
		\includegraphics[width=0.15\textwidth]{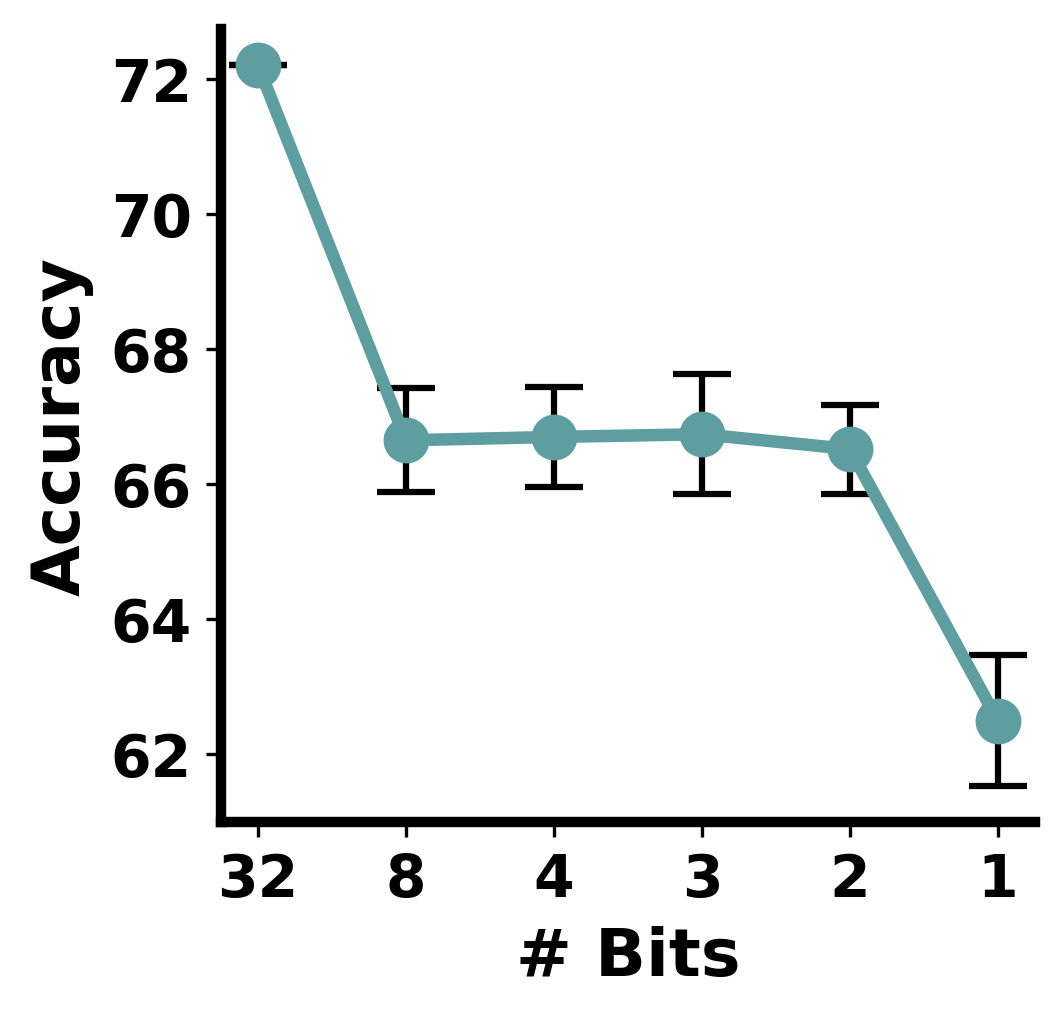}
	}	
	\vspace{-0.15in}
	\caption{Performance of quantized BERT with different weight bits and 4-bit activation on the GLUE Benchmarks. The results are obtained from 10 random seeds except for MNLI with 3 seeds.}
	\label{fig:acc_drop_a4}
\vspace{-0.1in}
\end{figure*}

Here we provide more empirical results on the sharp drop in performance as a result of binarization. We run multi-bit quantization on the BERT model over representative tasks of the GLUE benchmark, and activations are quantized in both 8-bit and 4-bit. We run 10 independent experiments for each task except for MNLI with 3 runs. We follow the same procedure  in Section~\ref{sec:sharp_drop}, and the default experimental setup in Appendix~\ref{sec:hyper} without data augmentation and splitting. The results are shown in Figures~\ref{fig:acc_drop_a8} and~\ref{fig:acc_drop_a4} respectively. It can be found that while the performance drops slowly from full-precision to ternarization, there is a consistent sharp drop by binarization in each tasks and on both 8-bit and 4-bit activation quantization. This is similar to the findings in Figure~\ref{fig:acc_drop}.

\subsection{More Visualizations of Loss Landscape}
\label{sec:more_visual_loss}
\begin{figure*}[h]
\vspace{-0.2in}
    \subfigure[Full-precision Model.]{
	    \includegraphics[width=0.22\textwidth]{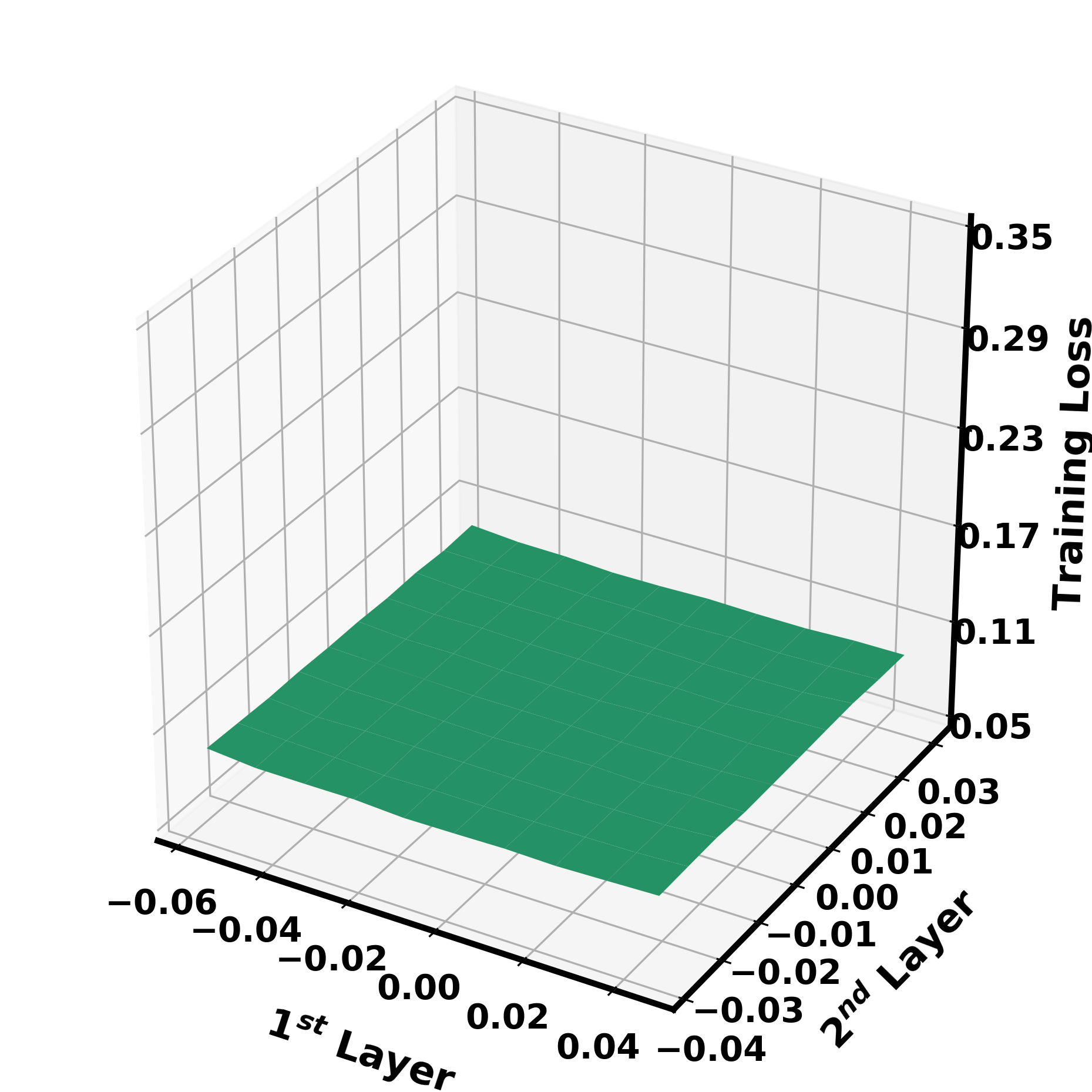}
	}
    \subfigure[Ternary Model.]{
	    \includegraphics[width=0.22\textwidth]{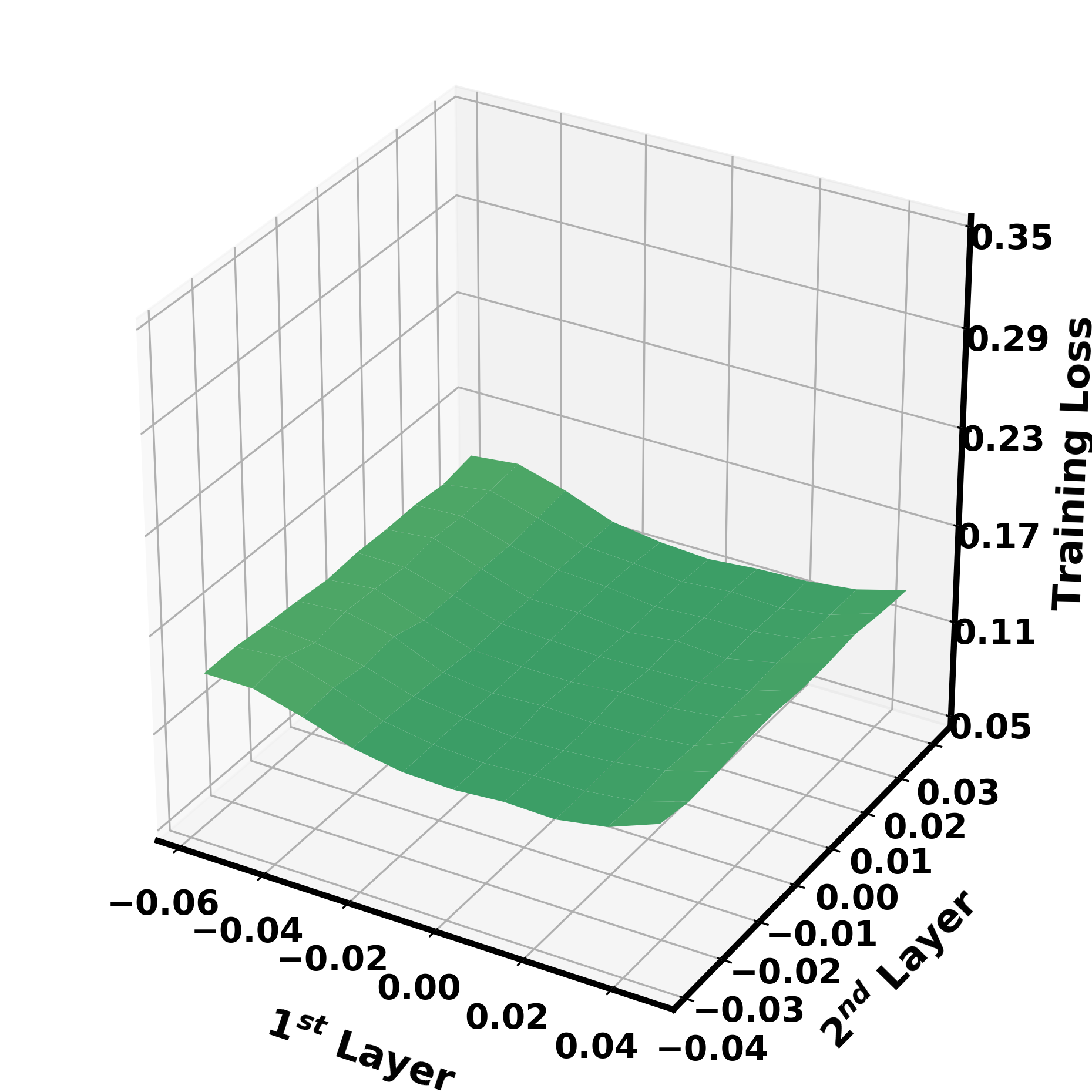}
	}
    \subfigure[Binary Model.]{
	    \includegraphics[width=0.22\textwidth]{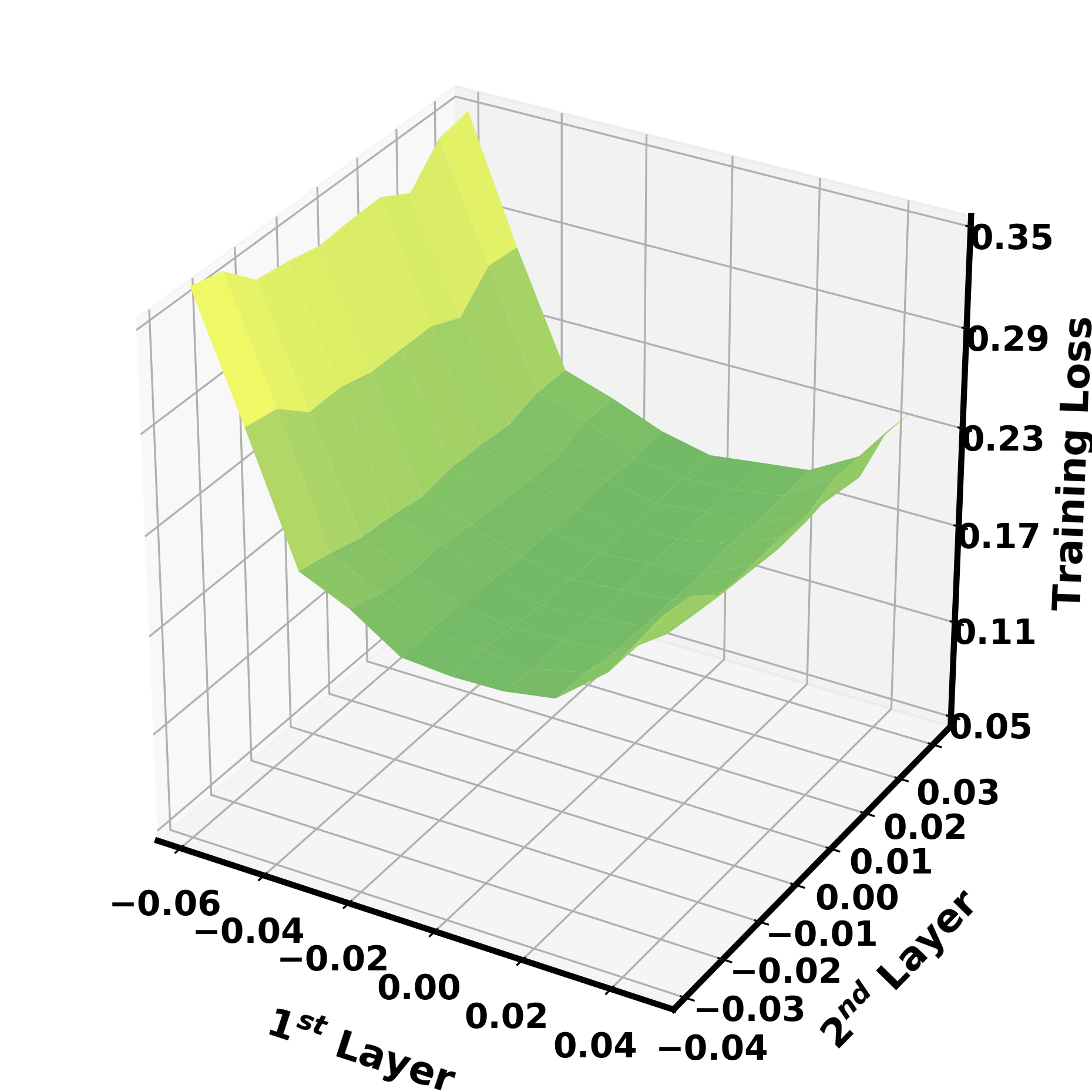}
	}
	\subfigure[All Together.] { 
		\includegraphics[width=0.22\textwidth]{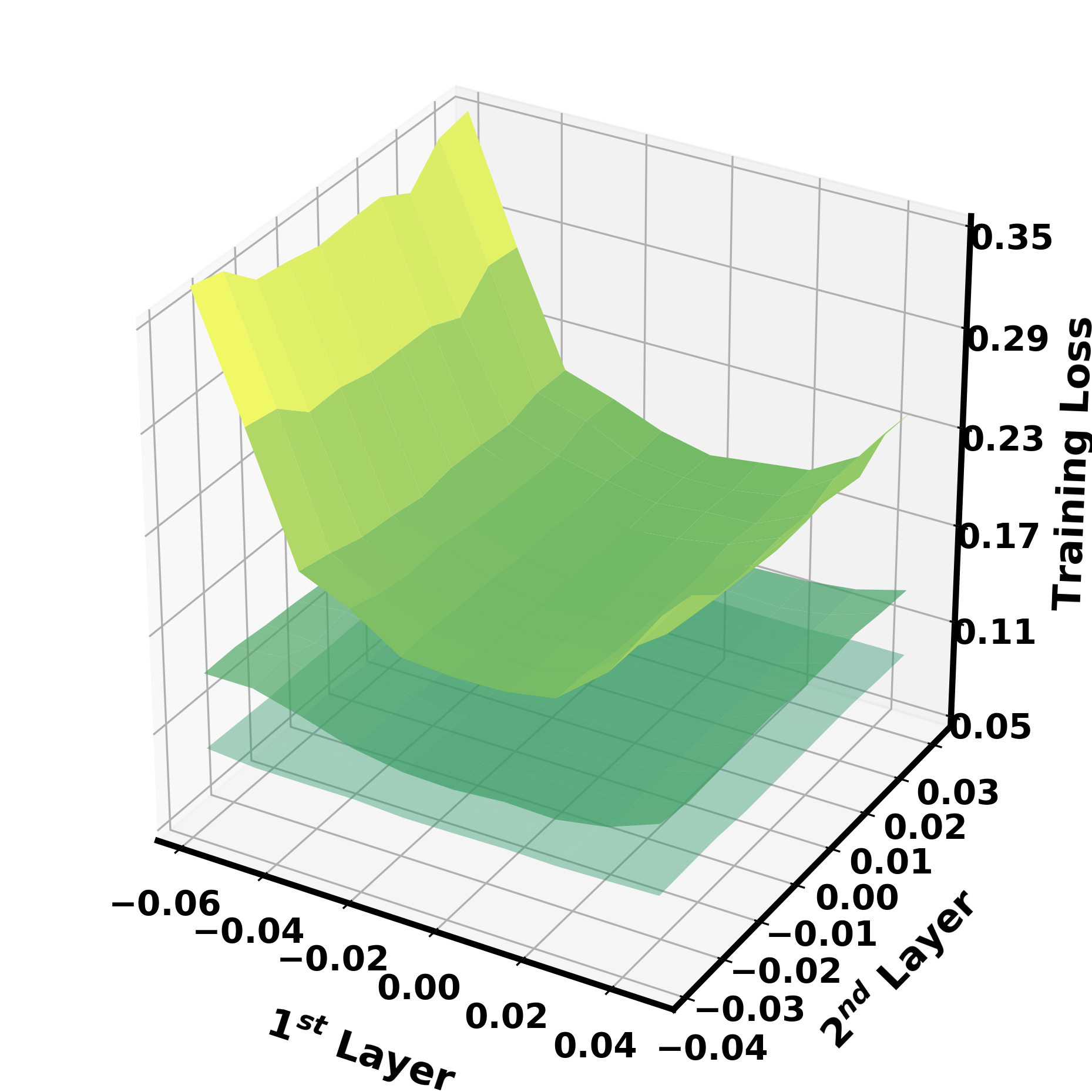}
	}
	\vspace{-0.15in}
	\caption{Loss landscape visualizations w.r.t MHA-K parameters of the $1^{\textrm{st}}$ and $2^{\textrm{nd}}$ Transformer layers on MRPC.}
	\label{fig:key_loss_landscape}
\end{figure*}

\begin{figure*}[h]
\vspace{-0.2in}
    \subfigure[Full-precision Model.]{
	    \includegraphics[width=0.22\textwidth]{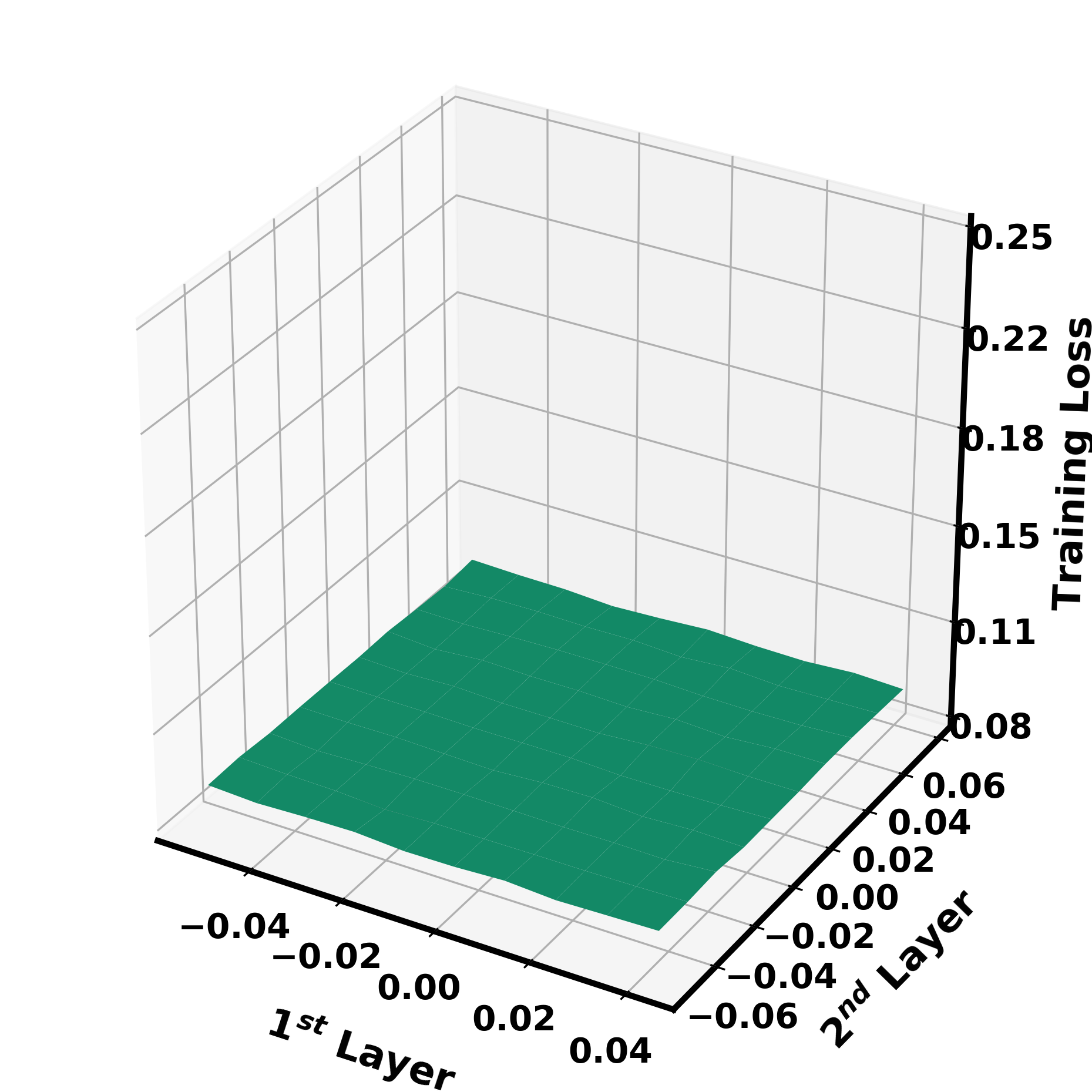}
	}
    \subfigure[Ternary Model.]{
	    \includegraphics[width=0.22\textwidth]{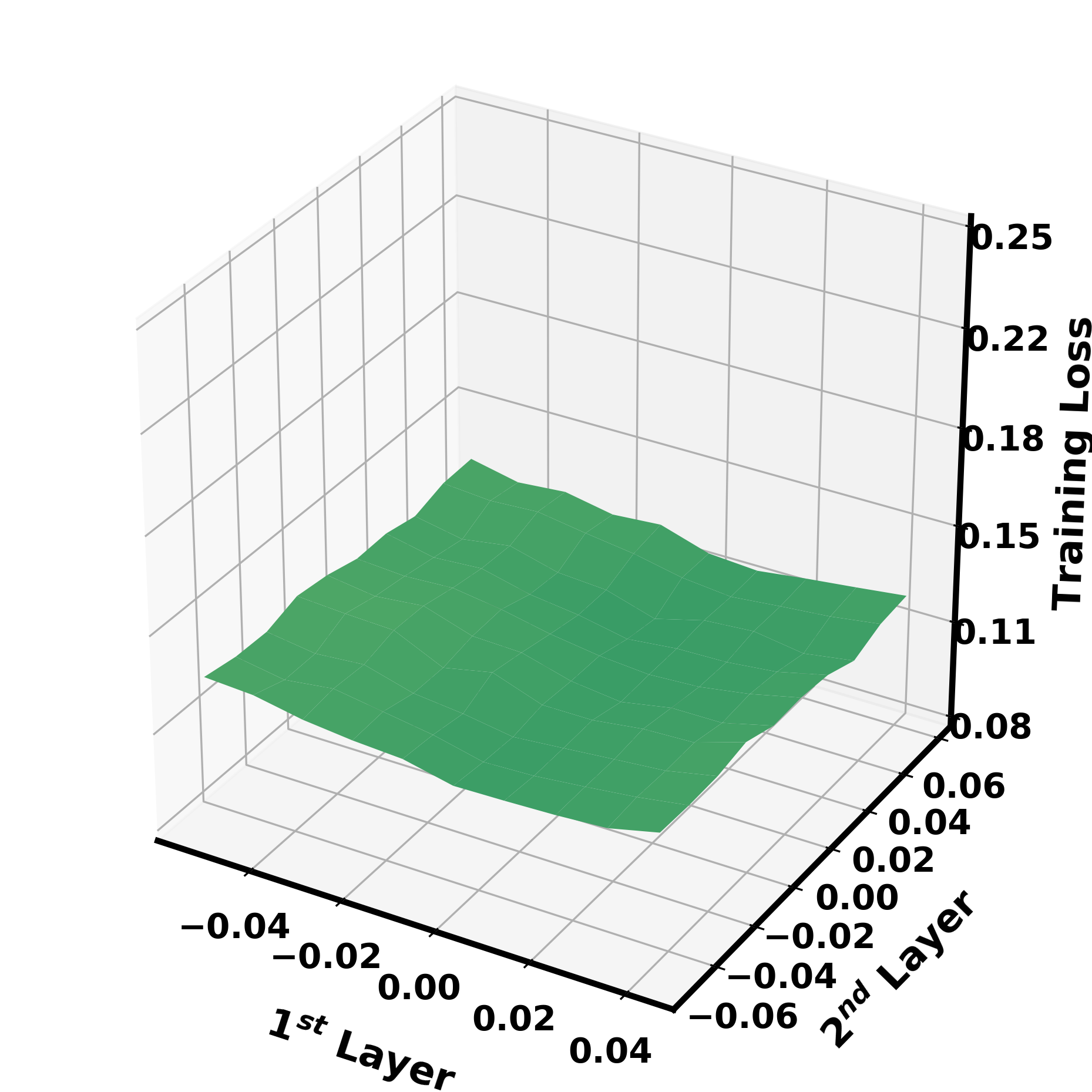}
	}
    \subfigure[Binary Model.]{
	    \includegraphics[width=0.22\textwidth]{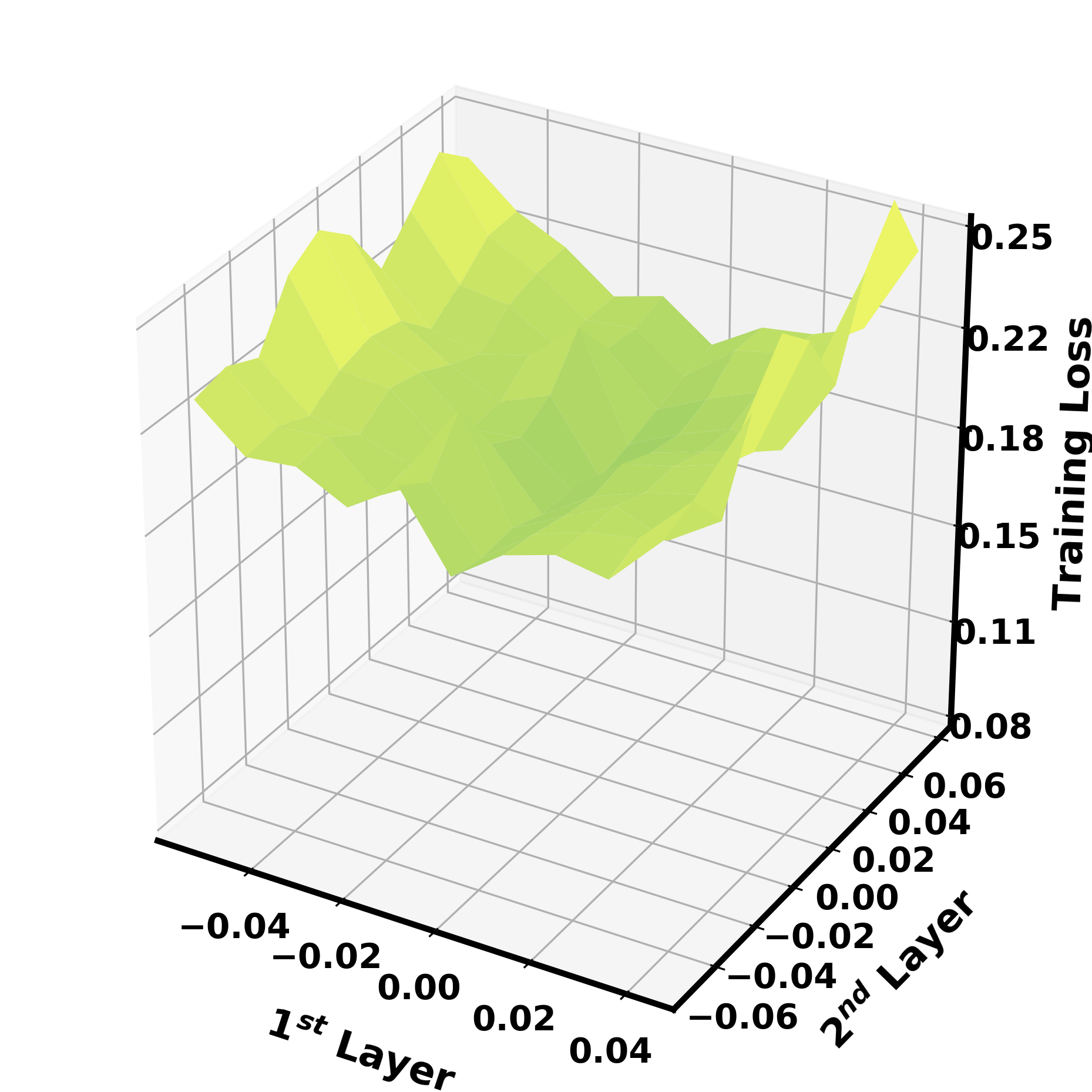}
	}
	\subfigure[All Together.] { 
		\includegraphics[width=0.22\textwidth]{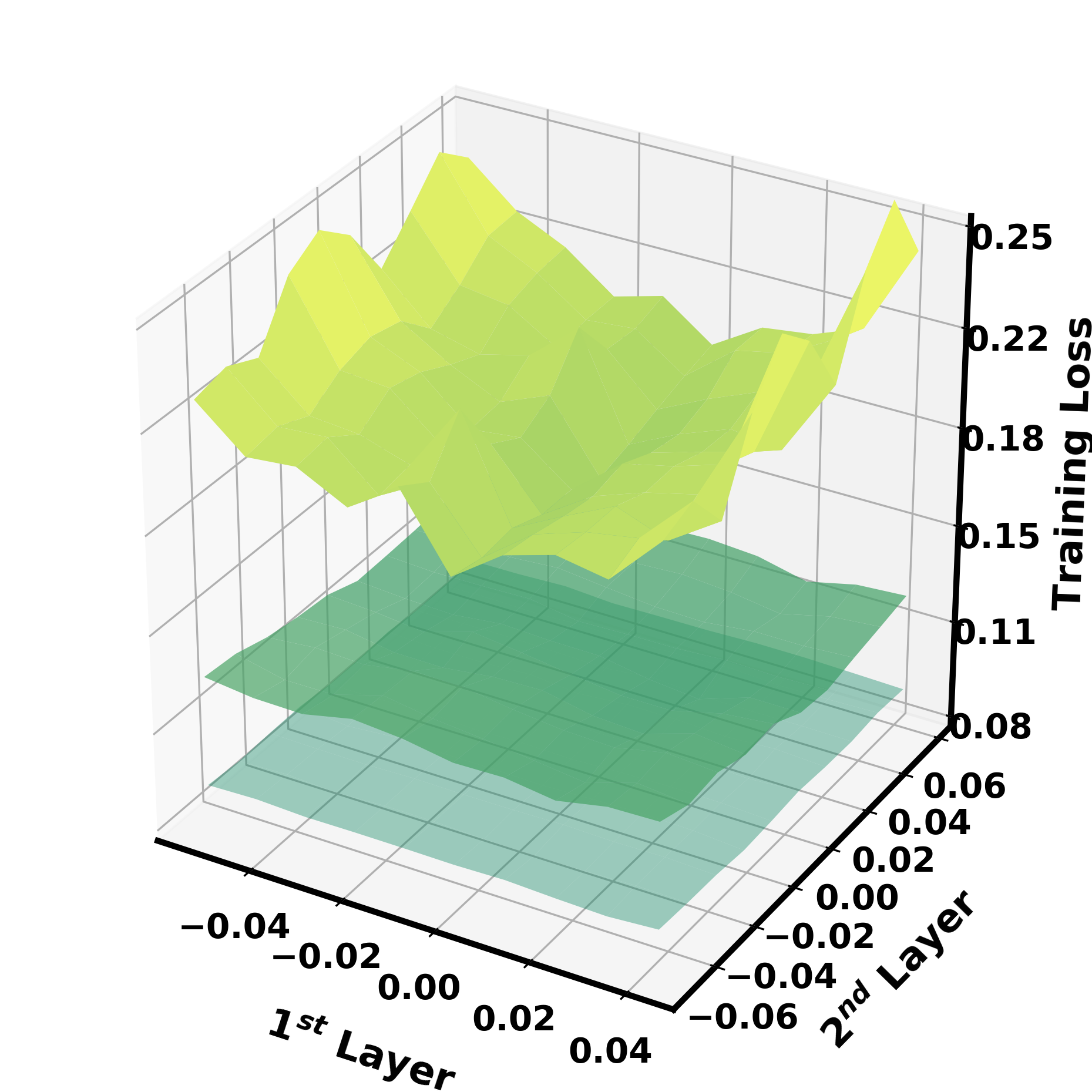}
	}
	\vspace{-0.15in}
	\caption{Loss landscape visualizations w.r.t MHA-Out parameters of the $1^{\textrm{st}}$ and $2^{\textrm{nd}}$ Transformer layers on MRPC.}
	\label{fig:selfout_loss_landscape}
\end{figure*}

\begin{figure*}[h]
\vspace{-0.2in}
    \subfigure[Full-precision Model.]{
	    \includegraphics[width=0.22\textwidth]{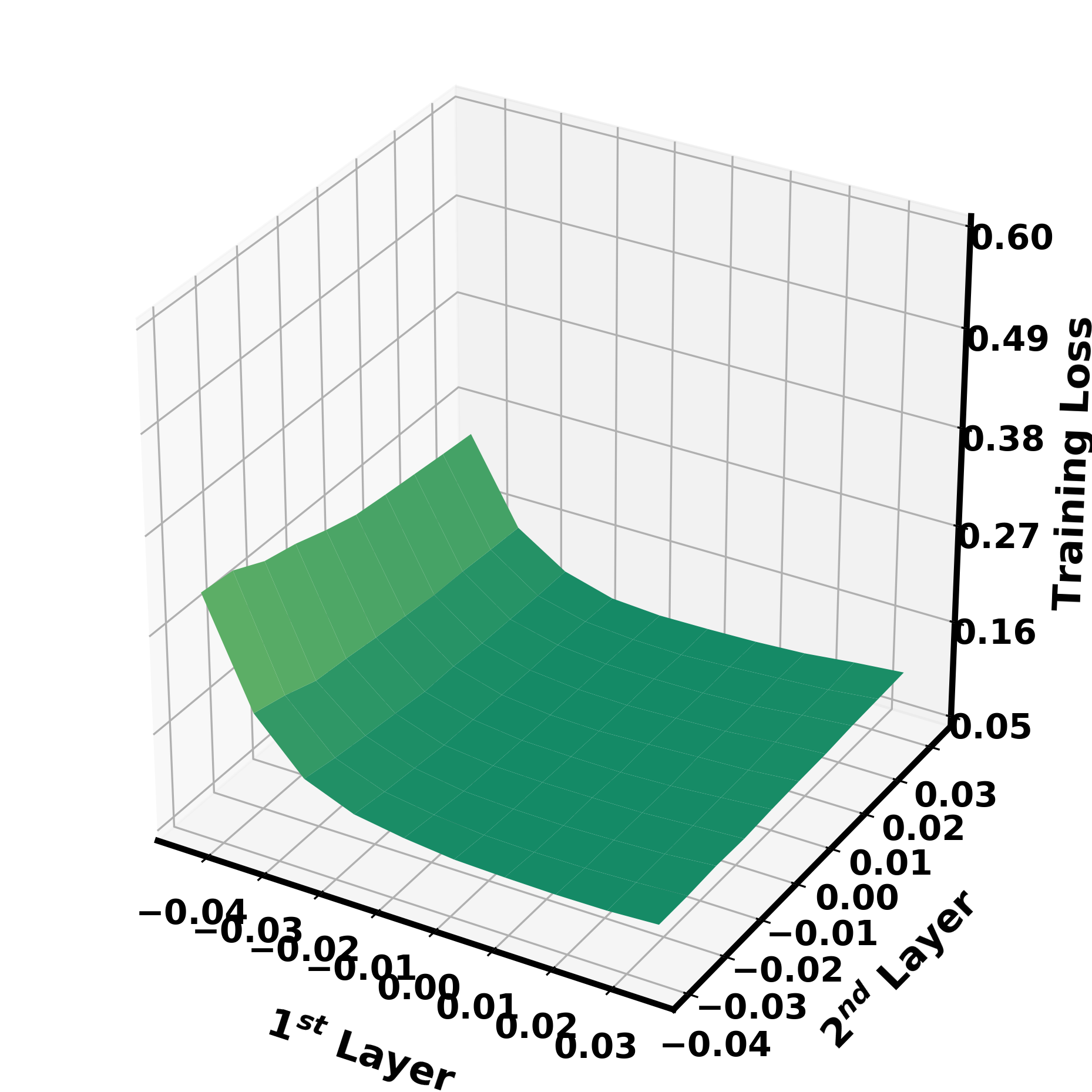}
	}
    \subfigure[Ternary Model.]{
	    \includegraphics[width=0.22\textwidth]{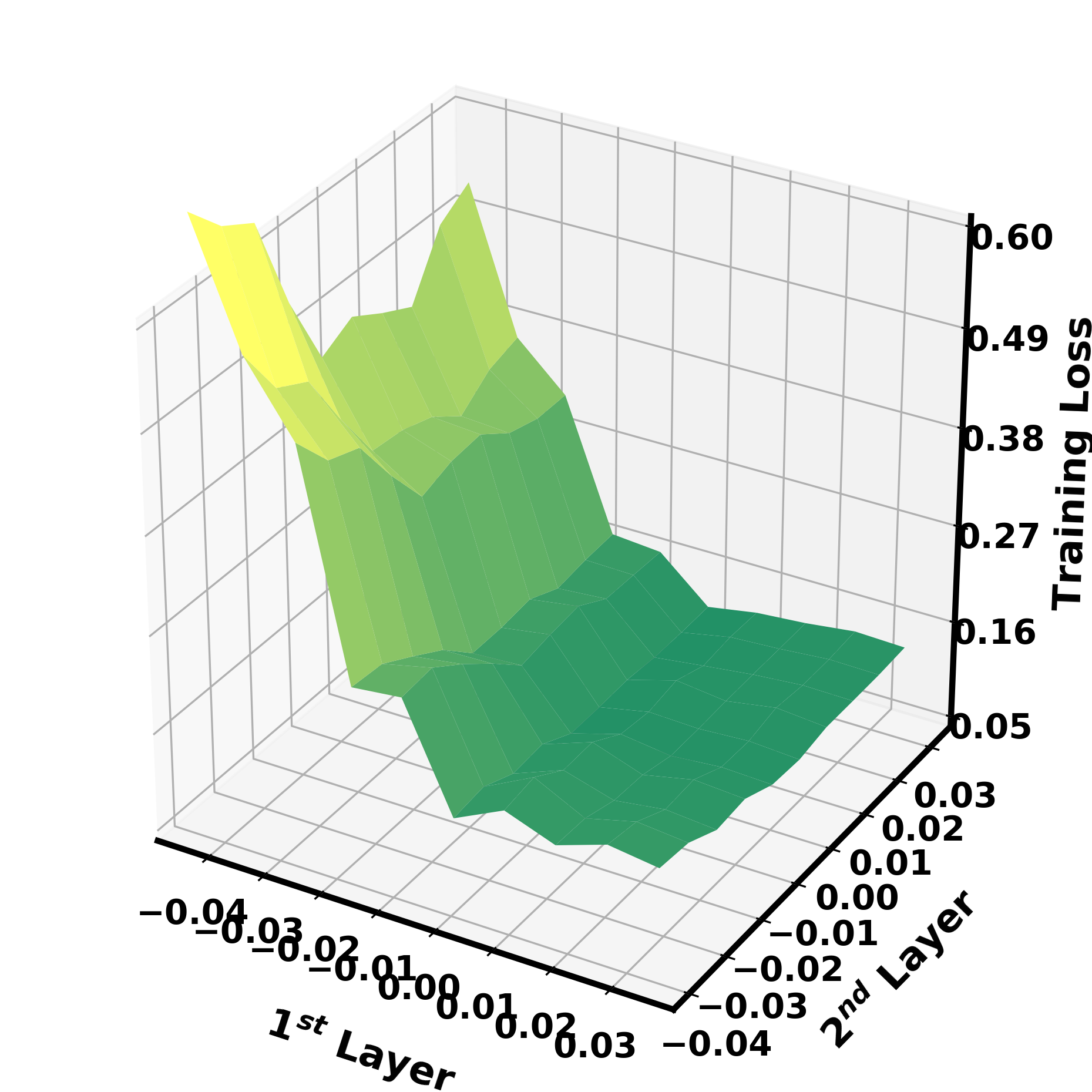}
	}
    \subfigure[Binary Model.]{
	    \includegraphics[width=0.22\textwidth]{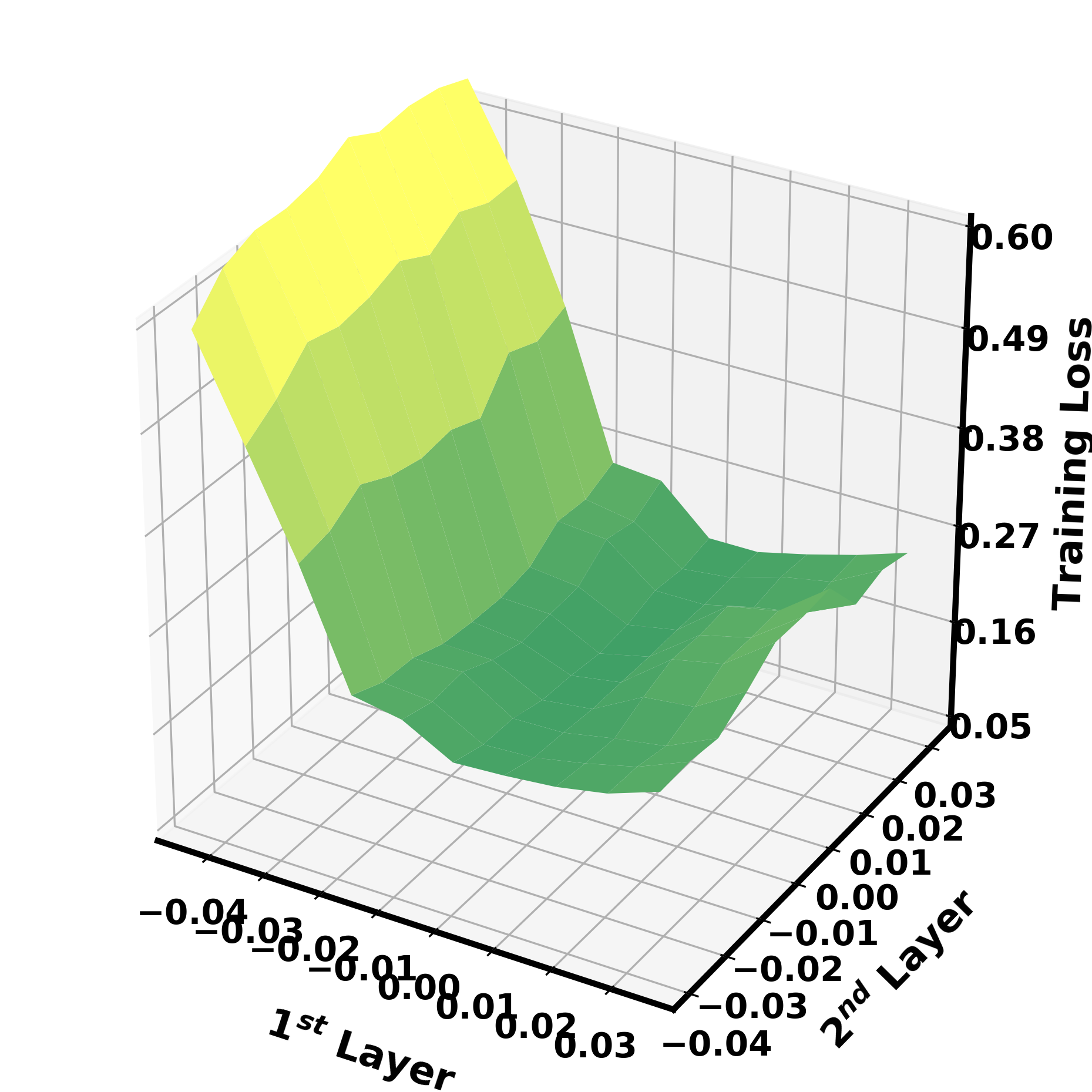}
	}
	\subfigure[All Together.] { 
		\includegraphics[width=0.22\textwidth]{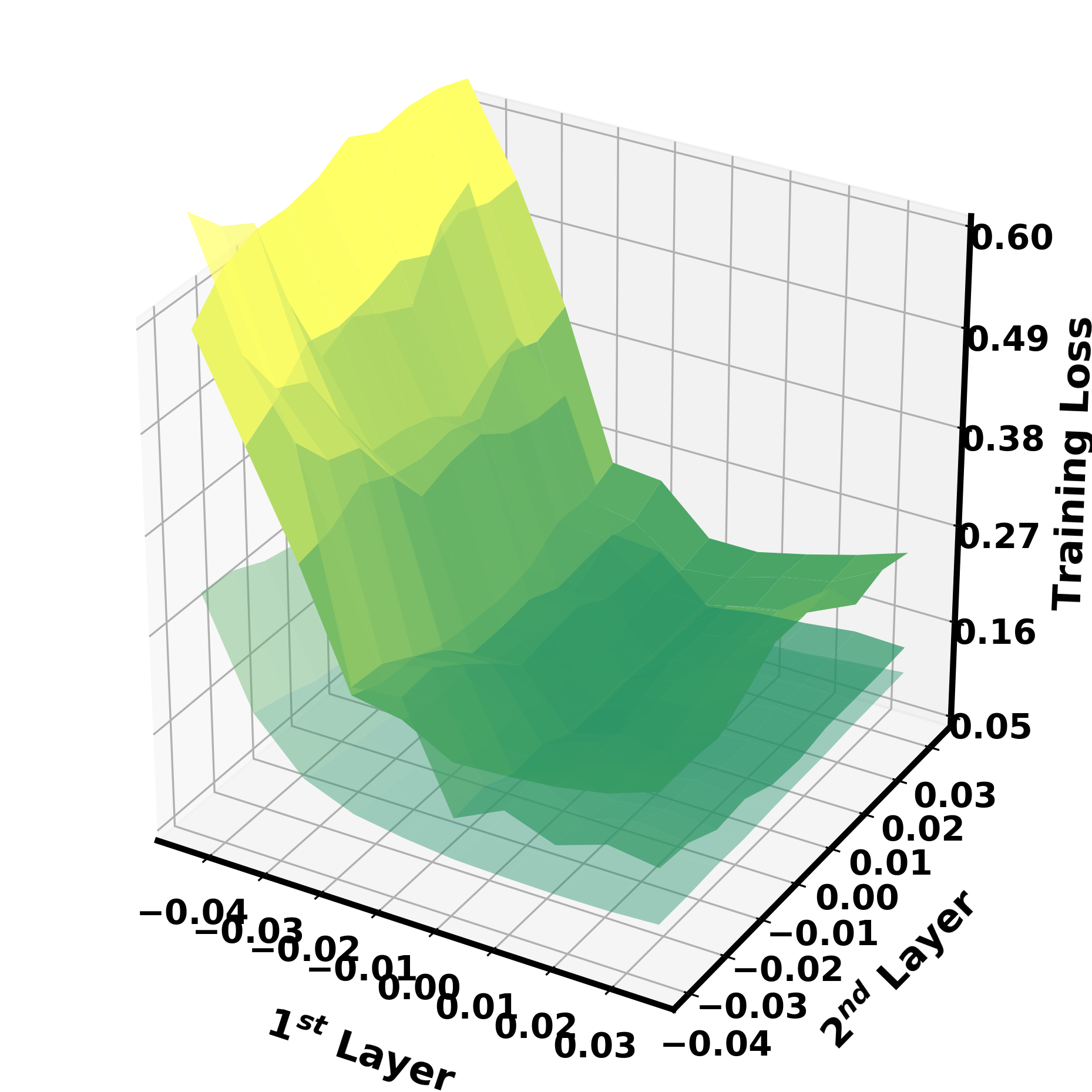}
	}
	\vspace{-0.15in}
	\caption{Loss landscape visualizations w.r.t FFN-Mid parameters of the $1^{\textrm{st}}$ and $2^{\textrm{nd}}$ Transformer layers on MRPC.}
	\label{fig:ffn_mid_loss_landscape}
\end{figure*}

\begin{figure*}[h]
\vspace{-0.2in}
    \subfigure[Full-precision Model.]{
	    \includegraphics[width=0.22\textwidth]{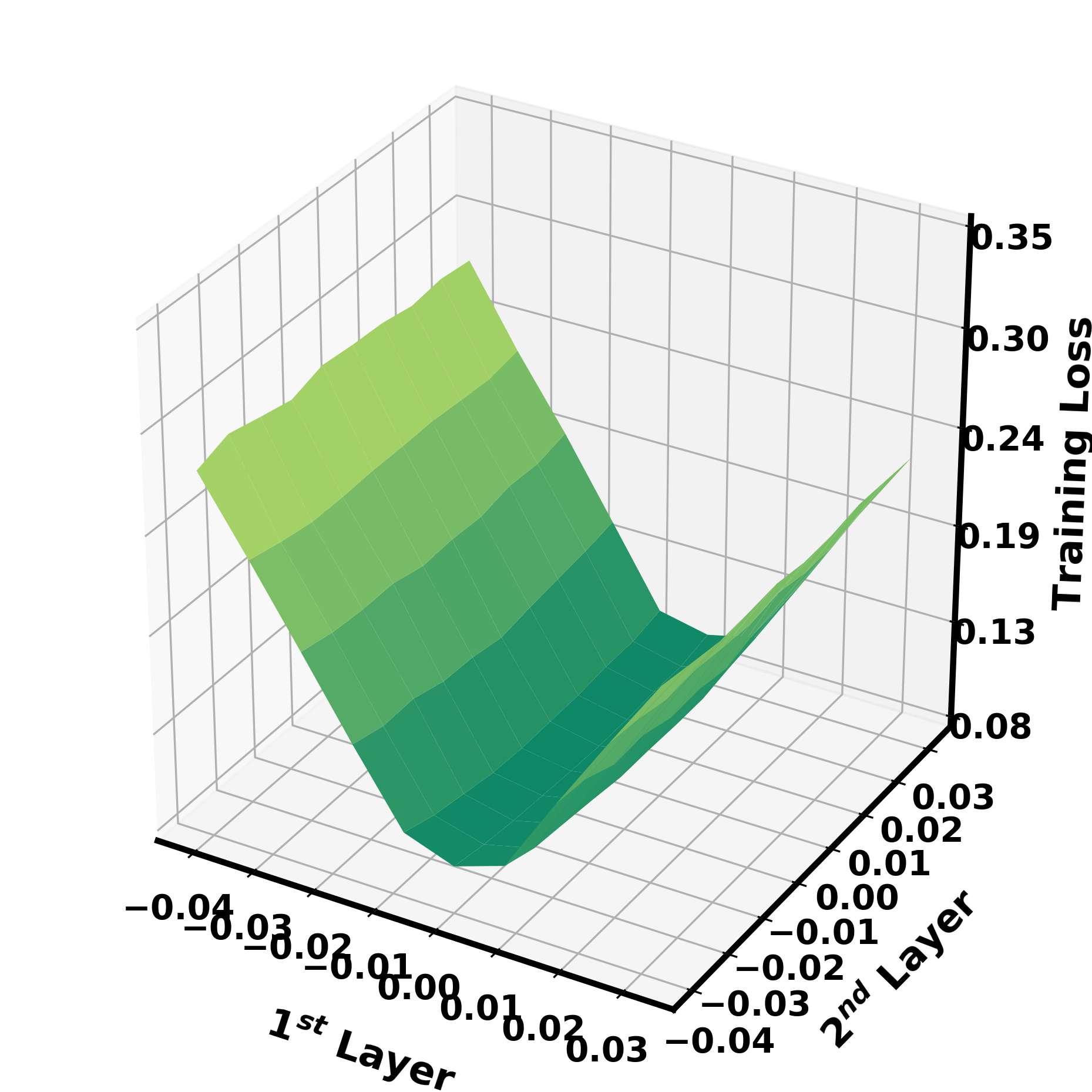}
	}
    \subfigure[Ternary Model.]{
	    \includegraphics[width=0.22\textwidth]{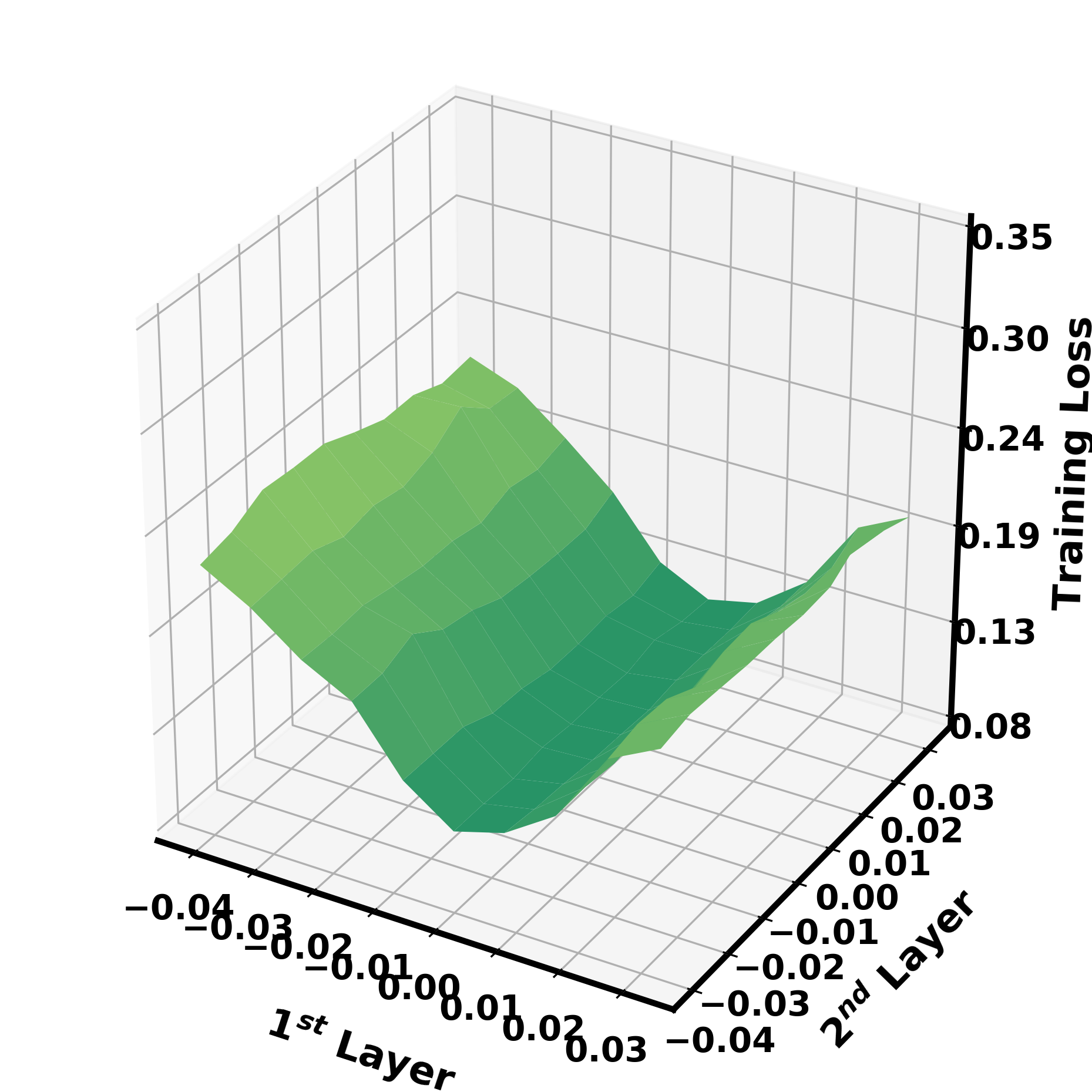}
	}
    \subfigure[Binary Model.]{
	    \includegraphics[width=0.22\textwidth]{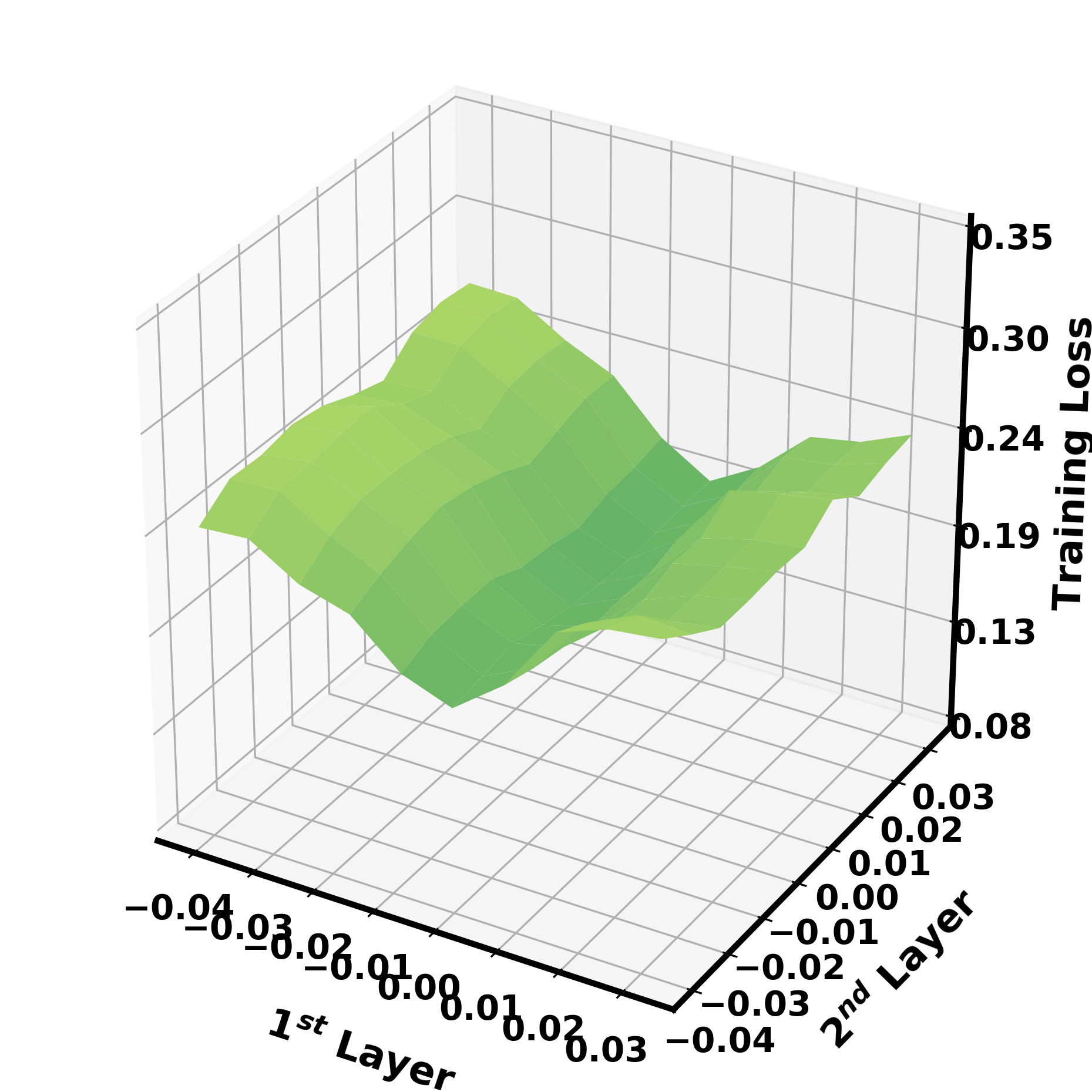}
	}
	\subfigure[All Together.] { 
		\includegraphics[width=0.22\textwidth]{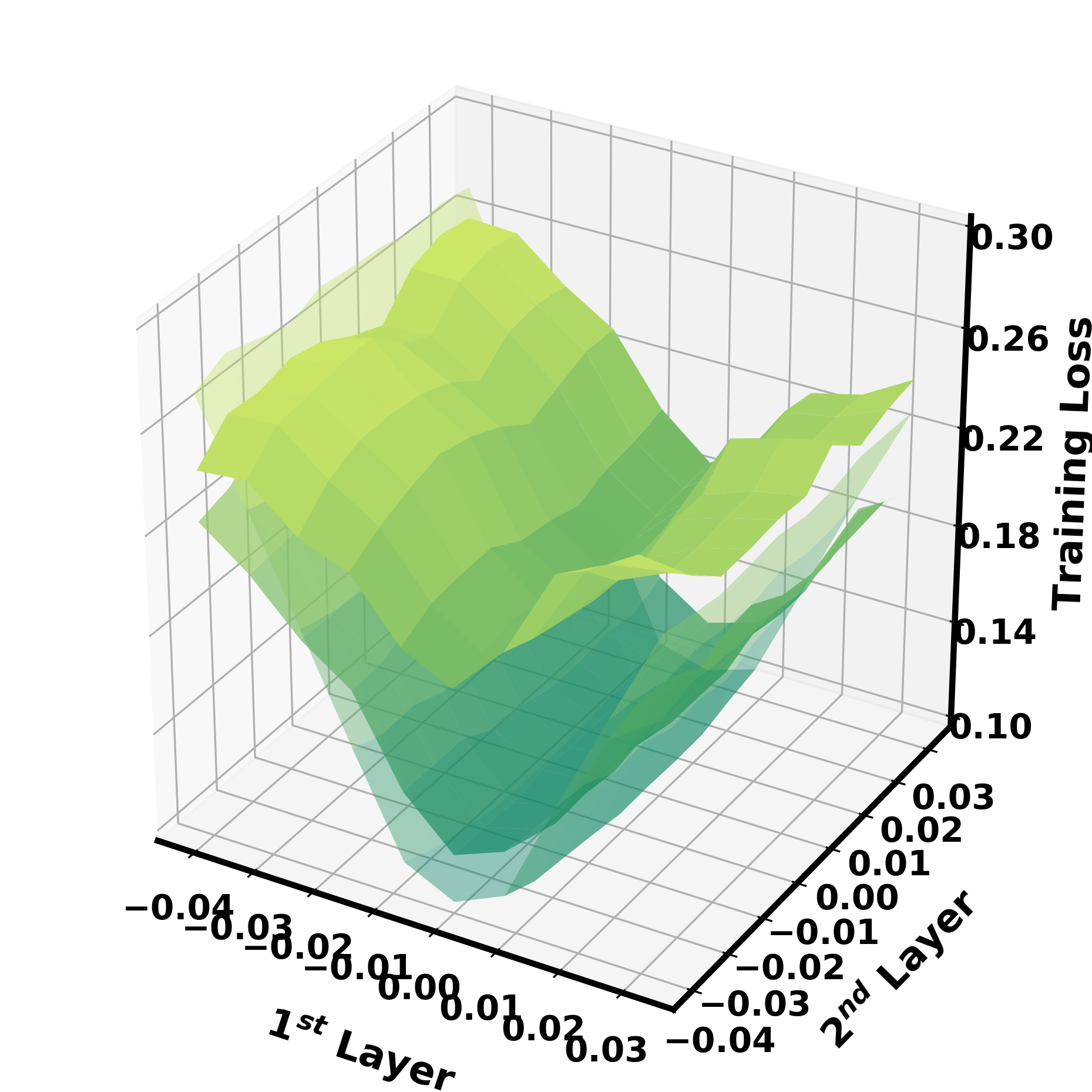}
	}
	\vspace{-0.15in}
	\caption{Loss landscape visualizations w.r.t FFN-Out parameters of the $1^{\textrm{st}}$ and $2^{\textrm{nd}}$ Transformer layers on MRPC.}
	\label{fig:ffn_out_loss_landscape}
	\vspace{-0.1in}
\end{figure*}

To comprehensively compare the loss curvature among the full-precision, ternary and binary models, we provide more  landscape visualizations aside from the value layer in Figure~\ref{fig:loss_landscape}. We extract parameters from MHA-K, MHA-O, FFN-Mid and FFN-out in the first two Transformer layers, and the corresponding landscape are shown in Figure~\ref{fig:key_loss_landscape}, Figure~\ref{fig:selfout_loss_landscape}, Figure~\ref{fig:ffn_mid_loss_landscape}, Figure~\ref{fig:ffn_out_loss_landscape} respectively. We omit MHA-Q due to page limitation, and also it is symmetric to MHA-K with similar landscape observation.
It can be found that binary model have steep and irregular loss landscape in general w.r.t different parameters of the model, and is thus hard to optimize directly.

\begin{table}[t]
    \centering
	\resizebox{0.5\textwidth}{!}{
	\begin{tabular}{c|c|cccccc|c}
		 \textbf{\tabincell{c}{Size\\(MB)}} &
		\textbf{Strategy} & 
		\textbf{QNLI} & \textbf{SST-2} & \textbf{CoLA} & \textbf{STS-B} & \textbf{MRPC} & \textbf{RTE} & \textbf{Avg.} \\\hline
		\multirow{3}{*}{10.6} & Min. & {91.1} & {93.1} & 52.8 & {88.2} & 85.3 & 69.3 & 80.0  \\
         & Rand. & 90.8 & 92.7 & 53.3 & {88.2} & 85.5 & 70.0 & 80.1  \\
         & Max. & 91.0 & 92.7 & {53.7} & 88.0 &  {86.5} & {71.1} & {80.5} \\\hdashline
		\multirow{3}{*}{11.4} & Min. & {91.0} & {93.0} & 53.8 & 88.3 & 85.5 & {71.5} & 80.5 \\
         & Rand. & {91.0} & 92.9 & {54.7} & {88.4} & {86.5} & 70.8 & {80.7} \\
         & Max. & {91.0} & {93.0} & 54.6 & {88.4} & 86.3 & 71.1 & {80.7} \\\hdashline
 	    \multirow{3}{*}{12.2} & Min. & {91.1} & 92.7 & 53.5 & 88.5 & 85.3 & 71.5 & 80.4 \\
         & Rand. & {91.1} & {92.9} & {54.1} & 88.5 & 86.0 & {71.8} & 80.4  \\
         & Max. & 91.0 & {92.9} & 53.8 & {88.6} & {86.8} & 71.1 &  {80.7}\\\hdashline
		\multirow{3}{*}{13.0} & Min. & {91.2} & 92.8 & 54.8 & 88.5 & 85.1 & 72.2 & 80.8  \\
         & Rand. & {91.2} & 92.9 & 54.1 & 88.4 & 86.0 & {71.8} & 80.8 \\
         & Max. & 91.1 & {93.1} & {56.1} & {88.6} & {86.1} & 70.8 & {81.0} \\\hdashline
		\multirow{3}{*}{13.8} & Min. & 91.1 & {93.0} & 55.4 & 88.5 & 85.8 & 71.5 & 80.9 \\
         & Rand. & {91.5} & 92.9 & 54.7 & 88.5 & 85.0 & 72.2 & 80.8 \\
         & Max. & 91.4 & 92.9 & {55.5} & {88.7} & {86.3} & {72.6} & {81.2} \\
	\end{tabular}}
	\caption{Results on GLUE development set for adaptive splitting with 8-bit activation quantization.}
	\label{table:adaptive_8bit}
\end{table}

\begin{table}[t]
	\centering
	\resizebox{0.5\textwidth}{!}{
	\begin{tabular}{c|c|cccccc|c}
		 \textbf{\tabincell{c}{Size\\(MB)}} &
		 \textbf{Strategy} & 
		 \textbf{QNLI} & \textbf{SST-2} & \textbf{CoLA} & \textbf{STS-B} & \textbf{MRPC} & \textbf{RTE} & \textbf{Avg.} \\\hline
		\multirow{3}{*}{10.6} & Min. & 90.6 & 92.6 & 51.7 & 87.4 & {85.3} & {70.8} & 79.7  \\
         & Rand. & {91.1} & {92.7} & 51.3 & {87.6} & 84.8 & 68.2 & 79.3   \\
         & Max. & 90.9 & {92.7} & {53.5} & 87.5 & 84.6 & 70.0 & {79.9} \\\hdashline
		\multirow{3}{*}{11.4} & Min. & 90.9 & {92.8} & 50.9 & 87.6 & {85.3} & 69.4 & 79.5 \\
         & Rand. & 90.8 & {92.8} & 51.7 & 87.5 & 84.6 & {70.4} & 79.6  \\
         & Max. & {91.1} & 92.6 & {52.1} & {87.7} & {85.3} & 70.0 & {79.8}  \\\hdashline
 	    \multirow{3}{*}{12.2} & Min. & 90.9 & 92.7 & 50.8 & {87.6} & 84.8 & {70.4} & 79.5 \\
         & Rand. & {91.2} & {93.0} & 52.0 & {87.6} & {85.1} & 70.0  & 79.8 \\
         & Max. & 90.9 & 92.9 & {52.2} & {87.6} & {85.1} & {70.4} & {79.9} \\\hdashline
         \multirow{3}{*}{13.0} & Min. & 91.1 & 92.8 & 52.6 & 87.7 & {86.3} & {69.7} & 80.0 \\
         & Rand. & {91.3} & {93.0} & 52.9 & {87.8} & 85.8 & {69.7} & {80.1} \\
         & Max. & {91.3} & 92.9 & {53.4} & {87.8} & 85.3 & {69.7} & {80.1}  \\\hdashline
        \multirow{3}{*}{13.8} & Min. & 91.1 & {93.1} & 51.5 & 87.9 & 84.8 & 70.0 & 79.7 \\
         & Rand. & {91.3} & 92.9 & 52.3 & 87.7 & 85.1 & {71.1} & 80.1 \\
        & Max. & {91.3} & 92.8 & {53.6} & {88.0} & {85.8} & 70.8 & {80.4}  \\
	\end{tabular}}
    \caption{Results on GLUE development set for adaptive splitting with 4-bit activation quantization.}
	\label{table:adaptive_4bit}
\end{table} 

\subsection{Ablation of Knowledge Distillation}

While knowledge distillation on BERT has been thoroughly investigated in ~\cite{jiao2020tinybert,hou2020dynabert,zhang2020ternarybert}, here we further conduct ablation study of knowledge distillation on the proposed ternary weight splitting. We compare with no distillation (``N/A"), prediction distillation (``Pred") and our default setting (``Int.+Pred"). For ``N/A" or ``Pred", fine-tuning after splitting follows the same setting to their ternary training. ``Int.+Pred" follows our default setting in Table~\label{tab:hyperparam}. We do not adopt data-augmentation, and results are shown in Table~\ref{tab:ablation_kd}. It can be found that ``Int.+Pred." outperforms both ``N/A" and ``Pred." with a clear margin, which is consistent to the findings in~\cite{zhang2020ternarybert} that knowledge distillation helps BERT quantization.

\begin{table}[t]
	\centering
	\vspace{-0.1in}
	\resizebox{0.49\textwidth}{!}{
	\begin{tabular}{c|c|cccc}
		\textbf{\tabincell{c}{KD}} &
		\textbf{\tabincell{c}{\#Bits\\(W-E-A)}} &
		\textbf{\tabincell{c}{MNLI\\(-m)}} &
        \textbf{\tabincell{c}{SST-2}} &
        \textbf{\tabincell{c}{CoLA}} & \textbf{\tabincell{c}{MRPC}} \\\hline 
		N/A & 1-1-8 & 83.2 & 92.1 & 49.2 & 82.8 \\
		Pred. & 1-1-8 & 84.0 & 91.7 & 48.6 & 84.1 \\
		Int.+Pred. & 1-1-8 & \textbf{84.2} & \textbf{92.6} & \textbf{53.4} & \textbf{85.5} \\\hline
		N/A & 1-1-4 & 82.6 & 90.9 & 39.2 & 76.5 \\
		Pred. & 1-1-4 & 83.4 & \textbf{92.3} & 38.9 & 76.2  \\
		Int.+Pred. & 1-1-4 & \textbf{83.9} & \textbf{92.3} & \textbf{44.4} & \textbf{83.3}
	\end{tabular}}
	\vspace{-0.1in}
	\caption{Ablation study on knowledge distillation.}
	\vspace{-0.1in}
	\label{tab:ablation_kd}
\end{table}


\subsection{Detailed Results of Adaptive Splitting}
\label{sec:detailed_adaptive}
The detailed comparison of our adaptive splitting strategy against the random strategy~(Rand.) and minimal gain strategy~(Min.) 
under different model size are shown
in Table~\ref{table:adaptive_8bit} and Table~\ref{table:adaptive_4bit}. 
It can be found that for both 8-bit and 4-bit activation quantization, our strategy that splits the most sensitive modules mostly performs the best on average under various model sizes.

\subsection{Architecture Visualization}
\begin{figure*}
    \centering
    \includegraphics[width=0.95\textwidth]{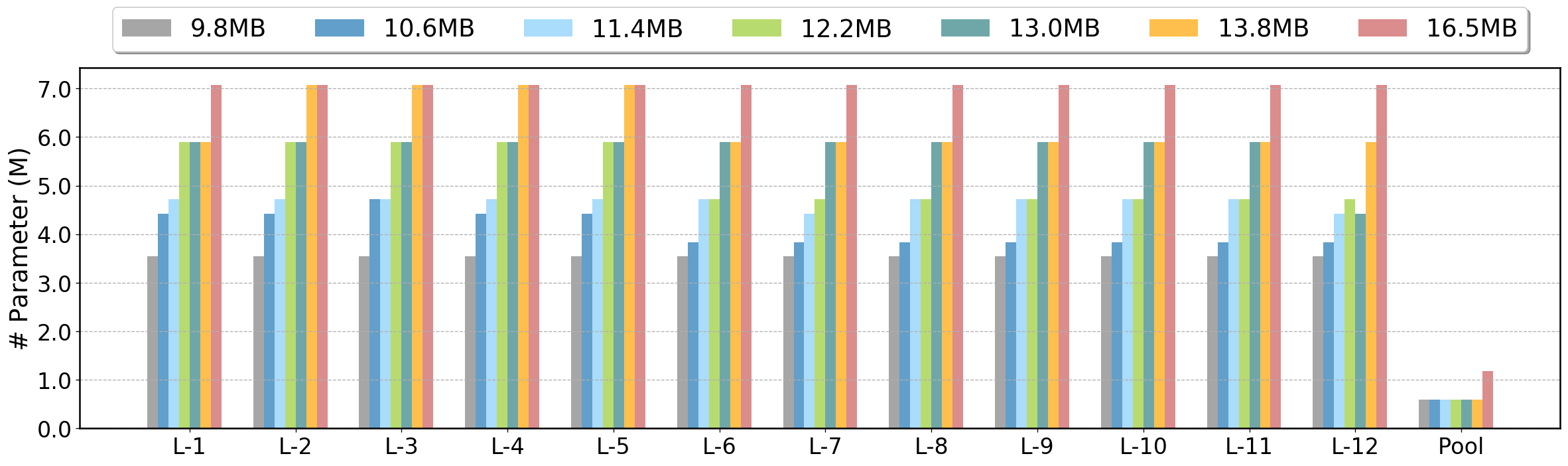}
    \caption{The architecture visualization for adaptive splitting on MRPC. The y-axis records the number of parameters split in each layer instead of the storage.}
    \label{fig:arch_visualization}
\end{figure*}

We further visualize the architectures after adaptive splitting on MRPC in Figure~\ref{fig:arch_visualization}. For clear presentation, we merge all splittable parameters in each Transformer layer.
As the baseline, 9.8MB refers to no splitting, while 16.5MB refers to splitting all splittable parameters in the model.
According to Figure~\ref{fig:arch_visualization}, with the increasing model size, shallower layers are more preferred for splitting than deeper layers, which is consistent to the findings in Figure~\ref{fig:sensitivity}.

\clearpage

\end{document}